\PassOptionsToPackage{capitalise,noabbrev}{cleveref}
\PassOptionsToPackage{hypertexnames=false}{hyperref}
\documentclass[12pt,cleveref]{jmlr}

\usepackage{times}
\usepackage{mathtools}
\usepackage{microtype}
\usepackage{tikz}
\usetikzlibrary{arrows.meta,calc,decorations.pathreplacing}
\usepackage{wrapfig}
\usepackage{xurl}
\hypersetup{linkcolor=blue!55!black,citecolor=blue!55!black,urlcolor=blue!55!black}

\renewenvironment{proof}[1][\proofname]
{\par\noindent{\bfseries\upshape #1\ }}
{\jmlrQED}
\usepackage{thm-restate}
\usepackage{autonum} %

\makeatletter
\let\ps@jmlrtps\ps@plain
\patchcmd{\@jmlrmaketitle}
{\jmlrpreauthor \ignorespaces\@author \jmlrpostauthor}
{\ifx\@author\@empty\vskip\aftermaketitskip\else\jmlrpreauthor \ignorespaces\@author \jmlrpostauthor\fi}
{}{\errmessage{Title layout patch failed}}
\ifdefined\MakeLinkTarget
\patchcmd{\@begintheorem}{\hskip\labelsep}{\MakeLinkTarget{\@currentcounter}\hskip\labelsep}{}{\errmessage{Theorem hyperlink patch failed}}
\patchcmd{\@opargbegintheorem}{\hskip\labelsep}{\MakeLinkTarget{\@currentcounter}\hskip\labelsep}{}{\errmessage{Named theorem hyperlink patch failed}}
\fi
\makeatother

\crefname{lemma}{Lemma}{Lemmas}
\crefname{proposition}{Proposition}{Propositions}
\crefname{equation}{Eq.}{Eqs.}
\Crefname{equation}{Eq.}{Eqs.}
\crefname{equality}{Eq.}{Eqs.}
\Crefname{equality}{Eq.}{Eqs.}
\crefname{inequality}{Ineq.}{Ineqs.}
\Crefname{inequality}{Ineq.}{Ineqs.}
\creflabelformat{equality}{\textup{(#2#1#3)}}
\creflabelformat{inequality}{\textup{(#2#1#3)}}
\crefrangelabelformat{equality}{\textup{(#3#1#4)--(#5#2#6)}}
\crefrangelabelformat{inequality}{\textup{(#3#1#4)--(#5#2#6)}}
\makeatletter
\newcommand{\eqlabel}[1]{\global\@namedef{advcvx@labeltype@#1}{equality}\label{#1}}
\newcommand{\ineqlabel}[1]{\global\@namedef{advcvx@labeltype@#1}{inequality}\label{#1}}
\AtBeginDocument{%
    \let\advcvx@label@noarg\label@noarg
    \def\label@noarg#1{%
        \@ifundefined{advcvx@labeltype@#1}%
        {\advcvx@label@noarg{#1}}%
    {\label@optarg[\csname advcvx@labeltype@#1\endcsname]{#1}}}%
}
\makeatother

\newcommand{\proofstepinline}[1]{%
    \par\addvspace{0.6\baselineskip}
    \noindent
    {\normalfont\normalsize\bfseries\color{blue!55!black}Step #1.}\enspace\ignorespaces
}
\newcounter{popstep}
\crefname{popstep}{Step}{Steps}
\newcommand{\popstep}[1]{\refstepcounter{popstep}\paragraph{Step~\thepopstep: #1}}
\newcommand{\reason}[1]{\texttt{(#1)}}
\newcommand{\explan}[2]{\stackrel{\reason{#2}}{#1}}
\newcommand{\R}{\mathbb R}
\newcommand{\E}{\mathbb E}
\newcommand{\Pp}{\mathbb P}
\newcommand{\K}{\mathcal K}

\newcommand{\F}{\mathcal F}
\newcommand{\N}{\mathcal N}

\newcommand{\tr}{\operatorname{tr}}
\newcommand{\Lip}{\operatorname{Lip}}
\newcommand{\op}{\mathrm{op}}
\newcommand{\Reg}{\operatorname{Reg}}
\newcommand{\eReg}{\operatorname{eReg}}

\newcommand{\qRegbar}{\operatorname{qR\overline{eg}}}
\newcommand{\argmin}{\operatornamewithlimits{arg\,min}}
\newcommand{\1}{\mathbf 1}
\newcommand{\eps}{\varepsilon}
\renewcommand{\epsilon}{\varepsilon}
\newcommand{\ip}[2]{\left\langle #1,#2\right\rangle}
\newcommand{\norm}[1]{\left\lVert #1\right\rVert}
\newcommand{\normm}[2]{\left\lVert #1\right\rVert_{#2}}

\title[Adversarial Bandit Convex Optimization]
{Adversarial Bandit Convex Optimization with \texorpdfstring{$d^{3/2}\sqrt{n}$}{d\^{}(3/2) sqrt(n)} Regret}
\author{%
    \Name{Alireza Bakhtiari}\thanks{Corresponding author.}\thanks{Authors
    are listed in alphabetical order.}
    \Email{alirezaa@cs.washington.edu}\\
    \addr Paul G.\ Allen School of Computer Science \& Engineering,
    University of Washington
    \AND
    \Name{Kevin Jamieson} \Email{jamieson@cs.washington.edu}\\
    \addr Paul G.\ Allen School of Computer Science \& Engineering,
    University of Washington
    \AND
    \Name{Tor Lattimore} \Email{lattimore@google.com}\\
    \addr Google DeepMind
    \AND
    \Name{Lillian Ratliff} \Email{ratliffl@uw.edu}\\
    \addr Department of Electrical \& Computer Engineering,
    University of Washington
}

\begin{document}
\maketitle

\begin{abstract}
    We study adversarial bandit convex optimization, where in each of $n$ rounds a learner chooses a point $A_t$ in a convex body $\K\subset\R^d$ and observes only a noisy value of a convex loss $f_t:\K\to[0,1]$ at that point. The losses are chosen by an adversary, who may change them from round to round and adapt them to the learner's past actions.
    We improve the best known regret bound for this problem from $\widetilde O(d^{5/2}\sqrt n)$ \citep{lattimore2020improved,lattimore2026bandit} to $\widetilde O(d^{3/2}\sqrt n)$, which matches the rate conjectured by \citet{bubeck2017kernel}.
    Together with the recent lower bound of $\Omega(d^{4/3}\sqrt n)$ due to
    \citet{rajaraman2026price}, it determines the minimax regret up to a
    factor of $d^{1/6}$ and logarithmic factors.
    The algorithm is an online Newton step with Gaussian exploration.
    Unlike earlier methods of this kind, it uses unbiased estimates of the
    gradient and Hessian of the Gaussian smoothing of the loss, rather than
    of an optimistic surrogate.
\end{abstract}

\section{Introduction}
\label{sec:introduction}
In adversarial bandit convex optimization, a learner plays against an adversary for $n$ rounds. In round $t$, the adversary chooses a convex loss $f_t:\K\to[0,1]$, which may depend on everything that happened in earlier rounds. The learner then chooses an action $A_t\in\K$ and observes $f_t(A_t)$ plus noise. The goal is small regret: the learner's cumulative loss should be close to that of the best fixed action in hindsight for the realized losses $f_1,\ldots,f_n$, as defined in \cref{eq:regret}. Apart from convexity and boundedness, we make no assumptions on the losses. In particular, they need not be smooth or strongly convex.

We present an algorithm whose regret is $\widetilde O(d^{3/2}\sqrt n)$ with high probability. In each round, the algorithm estimates a quadratic approximation of a Gaussian smoothing of the loss and runs an online Newton step on these quadratics, while keeping its mean inside a focus region where the approximation can be trusted. A restart rule fires only when the regret accumulated since the last restart is already negative, so restarts do not add to the regret.

\section{Related Work}
\label{sec:related-work}
Early algorithms estimate gradients using function values \citep{kleinberg2004nearly,flaxman2005online}. These methods are simple, but their regret bounds grow faster than $\sqrt n$.
\citet{hu2016biased} show that proving $\sqrt n$ regret from these gradient
estimates requires more than the usual bounds on their bias and variance,
even for smooth convex losses.
For $d=1$, \citet{bubeck2015bandit} showed that the minimax regret is $\widetilde O(\sqrt n)$, and \citet{bubeck2016multi} extended this to every dimension with polynomial dependence on $d$. Neither result gives an algorithm. \citet{hazan2016optimal} gave the first algorithm with $\widetilde O(\sqrt n)$ regret,
with exponential dependence on $d$, and \citet{bubeck2017kernel} gave the first polynomial-time algorithm with $\widetilde O(\operatorname{poly}(d)\sqrt n)$ regret. Since then the question has been the dependence on $d$. \citet{lattimore2020improved} showed that the minimax regret is $\widetilde O(d^{5/2}\sqrt n)$, without giving an algorithm.
\citet{lattimore2023second} introduced an online Newton method achieving
$\widetilde O(d^{3/2}\sqrt n)$ regret in the stochastic, Lipschitz,
unconstrained setting.
In the adversarial setting, the online Newton method of \citet{fokkema2024online} has regret $\widetilde O(d^{7/2}\sqrt n)$, and \citet[Chapter~11]{lattimore2026bandit} reduces this to $\widetilde O(d^{5/2}\sqrt n)$. These adversarial methods build quadratic approximations of an optimistic surrogate of the loss, and between restarts they decay the inverse covariance. Our algorithm uses the ordinary Gaussian smoothing and needs no such modification. \Cref{thm:main} improves the best bound by a factor of $d$ and reaches the rate $d^{3/2}\sqrt n$ conjectured by \citet[Conjecture~1]{bubeck2017kernel}.
The same rate was known for stochastic losses \citep{fokkema2024online}. On the lower-bound side, \citet{rajaraman2026price} recently proved $\Omega(d^{4/3}\sqrt n)$, improving on the $\Omega(d\sqrt n)$ of \citet{dani2008stochastic}, which leaves a gap of $d^{1/6}$. We refer to \citet{lattimore2026bandit} for an overview of the literature on bandit convex optimization.

\section{Notation and Problem Formulation}
\label{sec:setting}

For $r>0$, let $\mathbb{B}_r=\{x\in\R^d:\norm{x}\leq r\}$ be the Euclidean ball
of radius $r$. Let $\K\subset\R^d$ be a convex body that is known to the
learner.
In each round $t=1,\ldots,n$, an adversary chooses a convex loss
$f_t:\K\to[0,1]$, the learner chooses an action $A_t\in\K$, and it observes
$f_t(A_t)+\xi_t$, where $\xi_t$ is the noise in round $t$.
The adversary is adaptive: $f_t$ may depend on everything that happened
before round $t$, including the earlier actions of the learner, but not on
the randomness that the learner uses in round $t$.
To make this precise, let $\zeta_t$ be the randomness of the algorithm in
round $t$ which it uses to make decisions, let
$\F_t=\sigma(f_1,\zeta_1,\xi_1,\ldots,f_t,\zeta_t,\xi_t,f_{t+1})$, and let
$\E_t[\cdot]=\E[\cdot\mid\F_{t-1}]$ and $\Pp_t(\cdot)=\Pp(\cdot\mid\F_{t-1})$.
Thus $\F_{t-1}$ contains the history up to the end of round $t-1$ together
with the loss $f_t$, so that under $\E_t$ the loss of round $t$ is fixed and
only the randomness of the learner and the noise in that round remain.
The noise is conditionally subgaussian, i.e., for all $t$,
\begin{align}
    \E_t[\xi_t\mid\zeta_t]=0,
    \qquad
    \E_t[\exp(\xi_t^2)\mid\zeta_t]\leq2.
    \eqlabel{eq:noise}
\end{align}
This includes noiseless feedback.
The quantity of interest is the realized regret against the best fixed
action in hindsight,
\begin{align}
    \Reg_n=\sum_{t=1}^nf_t(A_t)-\inf_{x\in\K}\sum_{t=1}^nf_t(x).
    \eqlabel{eq:regret}
\end{align}
Since the losses may react to the actions of the learner, the best action in
hindsight is itself random and depends on the whole game.
In coordinates with $0\in\operatorname{int}(\K)$, let
$\pi(x)=\inf\{r>0:x\in r\K\}$ be the Minkowski functional of $\K$.
The Minkowski projection onto $\K$ is the map from $\R^d$ to $\K$ given by
$x\longmapsto\frac{x}{\max\{1,\pi(x)\}}$.
It leaves points in $\K$ unchanged and maps each $x\notin\K$ to the point
where the segment from $0$ to $x$ meets the boundary of $\K$.
Let $U$ be uniform on the Euclidean unit sphere.
The only geometric quantity in the result is
\begin{align}
    M=\max\left\{\frac1{\sqrt d},\E[\pi(U)]\right\}.
    \eqlabel{eq:M}
\end{align}
If $\mathbb{B}_1\subseteq\K$, then $\pi$ is $1$-Lipschitz and $M\leq1$.
For a symmetric matrix $B\in\R^{d\times d}$, which need not be positive
semidefinite, we write $\normm{x}{B}^2=x^\top Bx$, and $\norm{B}_{\op}$ for
its operator norm.
Write $\gamma_{\mu,\Sigma}$ for the density of the multivariate normal
distribution $\N(\mu,\Sigma)$, for $\mu\in\R^d$ and $\Sigma\in\R^{d\times d}$
symmetric positive definite.
For a convex function $h:\R^d\to\R$ and points $x,y\in\R^d$ such that $h$ is
differentiable at $y$, the Bregman divergence of $h$ is
$D_h(x\,\|\,y)=h(x)-h(y)-\ip{\nabla h(y)}{x-y}$, which is nonnegative by convexity.
A convex function is differentiable almost everywhere, so $D_h(x\,\|\,X)$ is
defined almost surely when $X$ is Gaussian with a positive definite
covariance.

\paragraph{The Convex Extension.}
Each loss $f_t$ is defined only on $\K$, but the algorithm samples points from
Gaussian distributions, and these can fall outside $\K$.
Following \citet{fokkema2024online} and \citet[Eq.~(10.6)]{lattimore2026bandit},
we work with a convex extension of $f_t$ to $\R^d$ whose value at
any point can be computed from a single value of $f_t$.
Fix $\eps\in(0,1/2]$, let $\K_\eps=(1-\eps)\K$, and define
\begin{align}
    \pi_\eps(x)
    &
    =
    \begin{cases}
        1               & \text{if }x\in\K_\eps, \\
        \frac{\pi(x)}{1-\eps} & \text{otherwise}\,,
    \end{cases}
    \quad\text{and}\quad v(x)=\pi_\eps(x)-1\,.
    \eqlabel{eq:extension-notation}
\end{align}

\begin{wrapfigure}{R}{0.46\textwidth}
    \vspace{-6pt}
    \centering
    \includegraphics{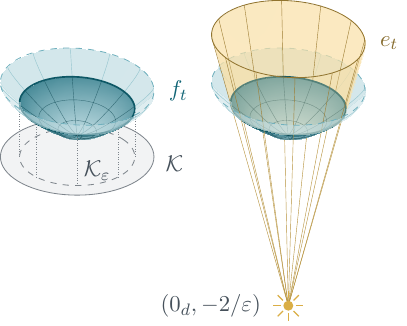}
    \caption{\small Left: a convex loss $f_t:\K\to[0,1]$, with $\K$ the unit disc.
    Right: its extension $e_t:\R^d\to\R$.}
    \label{fig:extension}
    \vspace{-10pt}
\end{wrapfigure}
Define for all $x\in\R^d$
\begin{align}
    e_t(x)
    =\pi_\eps(x)f_t\!\left(\frac{x}{\pi_\eps(x)}\right)
    +\frac{2}{\eps}v(x)\,.
    \eqlabel{eq:extension}
\end{align}

If $x\in\K_\eps$, then $\pi_\eps(x)=1$ and $v(x)=0$, so $e_t(x)=f_t(x)$.
If $x\notin\K_\eps$, then $x/\pi_\eps(x)$ lies on the boundary of
$\K_\eps$. Thus evaluating $e_t$ at any point of $\R^d$ requires only one
evaluation of $f_t$, at the point $x/\pi_\eps(x)\in\K_\eps$.
For $y\in\partial\K_\eps$ and $r\geq1$, the definition gives
$e_t(ry)=rf_t(y)+2(r-1)/\eps$. Hence the extension is affine along each
ray outside $\K_\eps$, as illustrated in \cref{fig:extension}.

The following lemma collects the properties used in the analysis.
\begin{lemma}\label{lem:extension}
    Suppose $\mathbb B_1\subseteq\K$. For $0<\eps\leq1/2$, the functions $e_t$ and $e_t-v/\eps$ are convex on
    $\R^d$, and $\Lip(e_t)\leq5/\eps$. Moreover, $e_t=f_t$ on $\K_\eps$ and
    \begin{align}
        f_t\bigl(x/\pi_\eps(x)\bigr)\leq e_t(x)
        \qquad\text{for every }x\in\R^d.
    \end{align}
\end{lemma}
The convexity statements follow from
\citet[Proposition~3.23]{lattimore2026bandit}. Appendix~\ref{app:extension}
gives the short argument and verifies the constant in the Lipschitz bound.

\WFclear
\section{The Algorithm}
\label{sec:algorithm}

There are three main ingredients in \textit{Smooth ONS} (\cref{alg:main}).
The first is \textit{Gaussian smoothing}:
the algorithm estimates the gradient and Hessian of a
smoothed version of the loss $e_t$, and forms a
quadratic surrogate of the loss at the current mean $\mu_t$.
The second is \textit{online Newton step}:
the mean and covariance of the sampling distribution are updated
by an online Newton step on the quadratic surrogate.
Third is \textit{focus region and restart}: the algorithm maintains an intersection
of ellipsoids called the \textit{focus region}.
The algorithm restarts with fresh parameters
when it certifies that it has done better than every action in
the \textit{focus region}.
The analysis shows that the regret accumulated before a restart is negative, so
the total regret is bounded by the regret accumulated after the last restart.
\begingroup
\RestyleAlgo{ruled}
\DontPrintSemicolon
\SetAlgoNoLine
\SetAlgoNoEnd
\LinesNumbered
\SetAlFnt{\small}
\newcommand{\algmark}[1]{\tikz[remember picture,overlay]\coordinate (#1);}
\newcommand{\algbrace}[3]{%
    \begin{tikzpicture}[remember picture,overlay]
        \draw[decorate,decoration={brace,amplitude=4pt},black!60]
        ([xshift=-64pt,yshift=7pt]ons-right |- #1) --
        ([xshift=-64pt,yshift=-5pt]ons-right |- #2)
        node[midway,right=7pt,text width=55pt,align=left,
        font=\scriptsize\itshape,text=black!65]{#3};
    \end{tikzpicture}%
}
\SetKwInput{KwIn}{Input}
\begin{algorithm2e}[!t]
    \caption{Smooth ONS}
    \label{alg:main}
    \KwIn{$\sigma$, $\eta$, $\lambda$, $R$, $\rho$ and $\eps$\hfill
    \algmark{ons-right}}
    \setlength{\rightskip}{70pt}
    \While{fewer than $n$ rounds have been played}{
        Start an epoch: set $\mu_1=0$, $\Sigma_1=\sigma^2I_d$, $K_0=\K_\eps$ and $Q_0=0$\;
        \For{$t=1,2,\ldots$ as long as rounds remain}{
            {\setlength{\rightskip}{0pt}%
            Set $E_t=\bigl\{x\in\R^d:\normm{x-\mu_t}{\Sigma_t^{-1}}^2\leq R^2\bigr\}$ and $K_t=K_{t-1}\cap E_t$\;}
            \algmark{ons-sample-top}Sample $X_t\sim\N(\mu_t,\Sigma_t)$ and play $A_t=X_t/\pi_\eps(X_t)$ and observe $f_t(A_t)+\xi_t$\;
            \algmark{ons-sample-bottom}Compute $g_t$ and $H_t$ by \cref{eq:gaussian-derivatives-estimates}\;
            \If{$\Sigma_t^{-1}+\eta\lambda H_t/2\not\succ0$}{play $0$ in all remaining rounds and stop\;}
            \algmark{ons-newton-top}Set $\Sigma_{t+1}^{-1}=\Sigma_t^{-1}+\eta\lambda H_t/2$\;
            \algmark{ons-newton-bottom}Set $\mu_{t+1}=\argmin_{u\in K_t} \normm{u-\bigl(\mu_t-\eta \Sigma_{t+1}g_t\bigr)}{\Sigma_{t+1}^{-1}}^2$\;
            \algmark{ons-restart-top}Set $Q_t=Q_{t-1}+q_t$ and $J_t(x)=\eta Q_t(x)+\normm{x}{\Sigma_1^{-1}}^2$\;
            Find $y_t\in K_t$ such that $J_t(y_t)\leq\min_{x\in K_t}J_t(x)+\rho/8$\;
            \If{$\eta Q_t(y_t)\geq2\rho$}{%
                \algmark{ons-restart-bottom}end the epoch%
                \algbrace{ons-sample-top}{ons-sample-bottom}{Gaussian\\sampling and\\derivative\\estimates}%
                \algbrace{ons-newton-top}{ons-newton-bottom}{Online\\Newton step}%
            \algbrace{ons-restart-top}{ons-restart-bottom}{Restart\\mechanism}\;}
        }
    }
\end{algorithm2e}
\endgroup

\paragraph{The Gaussian Smoothing}
The algorithm maintains a Gaussian distribution $\N(\mu_t,\Sigma_t)$ on $\R^d$.
The algorithm works with the convex extension $e_t$ in place of $f_t$, and
this costs little. Since $A_t=X_t/\pi_\eps(X_t)$, \cref{lem:extension}
gives $f_t(A_t)\leq e_t(X_t)$. For every $x\in\K$ the point $(1-\eps)x$
lies in $\K_\eps$, where $e_t=f_t$, and convexity gives
$f_t((1-\eps)x)\leq f_t(x)+\eps$. Hence, for every $x\in\K$,
\begin{align}
    \sum_{t=1}^n\bigl(f_t(A_t)-f_t(x)\bigr)
    \leq\sum_{t=1}^n\bigl(e_t(X_t)-e_t((1-\eps)x)\bigr)+n\eps,
    \ineqlabel{eq:extension-cost}
\end{align}
so it is enough to compare $e_t(X_t)$ with points of $\K_\eps$. Unlike
$f_t$, the extension is defined on all of $\R^d$, which allows us to define
\begin{align}
    F_{t,\Sigma}(y)=\E_{X\sim\N(y,\Sigma)}[e_t(X)]=\int_{\R^d}e_t(x)\gamma_{y,\Sigma}(x)\,dx
    \qquad\text{for }y\in\R^d.
    \eqlabel{eq:gaussian-smoothing}
\end{align}
This is the \textit{Gaussian smoothing} of $e_t$ with covariance $\Sigma$.
It is worth noting that $F_{t,\Sigma}$ is defined on all of $\R^d$,
is infinitely differentiable, and is a convex function that upper bounds $e_t$ by Jensen's inequality.
Our main interest in $F_{t,\Sigma}$ is that its gradient and Hessian can be expressed as
\begin{align}
    \nabla F_{t,\Sigma}(y)
    &=\E_{X\sim\N(y,\Sigma)}\bigl[e_t(X)\Sigma^{-1}(X-y)\bigr],\\
    \nabla^2 F_{t,\Sigma}(y)
    &=\E_{X\sim\N(y,\Sigma)}\bigl[e_t(X)\Sigma^{-1}\bigl((X-y)(X-y)^\top-\Sigma\bigr)\Sigma^{-1}\bigr].
    \eqlabel{eq:gaussian-derivatives}
\end{align}
These identities follow by differentiating the Gaussian density in \cref{eq:gaussian-smoothing}.
Indeed,
\begin{align}
    \nabla_y\gamma_{y,\Sigma}(x)
    &=\gamma_{y,\Sigma}(x)\Sigma^{-1}(x-y),\\
    \nabla_y^2\gamma_{y,\Sigma}(x)
    &=\gamma_{y,\Sigma}(x)\Sigma^{-1}\bigl((x-y)(x-y)^\top-\Sigma\bigr)\Sigma^{-1}.
\end{align}
Since $e_t$ is Lipschitz, it grows at most linearly, and the Gaussian decay
justifies differentiating under the integral.

These identities in \cref{eq:gaussian-derivatives} suggest a natural way to estimate the gradient and Hessian of $F_{t,\Sigma_t}$ at $\mu_t$.
Precisely, if the algorithm could build an unbiased estimate of $e_t(X_t)$ for a sample $X_t\sim\N(\mu_t,\Sigma_t)$, then it could use the same sample to form unbiased estimates of both the gradient and Hessian of $F_{t,\Sigma_t}$ at $\mu_t$.
To this end, after drawing one sample $X_t\sim\N(\mu_t,\Sigma_t)$, the algorithm plays a
point $A_t=X_t/\pi_\eps(X_t)\in\K$, observes $f_t(A_t)+\xi_t$ and computes
\begin{align}
    Y_t
    =\pi_\eps(X_t)\bigl(f_t(A_t)+\xi_t\bigr)
    +\frac2\eps v(X_t)
    \quad\text{so that}\quad
    \E_t[Y_t\mid X_t]=e_t(X_t).
    \eqlabel{eq:Y}
\end{align}
The point $A_t$ is the Minkowski projection of $X_t$ onto
$\K_\eps$.
Since $\pi_\eps(X_t)\geq1$ and $f_t\geq0$, we have $f_t(A_t)\leq e_t(X_t)$,
so $F_{t,\Sigma_t}(\mu_t)=\E_t[e_t(X_t)]$ is an upper bound on the actual expected
loss of the learner in round $t$.
As a result the algorithm only stores the observation pair $(X_t,Y_t)$,
which by the tower property of conditional expectation and identities in \cref{eq:gaussian-smoothing,eq:gaussian-derivatives} gives
\begin{align}
    \bar{g}_t \triangleq
    \nabla F_{t,\Sigma_t}(\mu_t)
    &=\E_t\left[\underbrace{Y_t\Sigma_t^{-1}(X_t-\mu_t)}_{\triangleq g_t}\right]\,,\\
    \bar{H}_t \triangleq
    \nabla^2 F_{t,\Sigma_t}(\mu_t)
    &=\E_t\left[\underbrace{Y_t\Sigma_t^{-1}\bigl((X_t-\mu_t)(X_t-\mu_t)^\top-\Sigma_t\bigr)\Sigma_t^{-1}}_{\triangleq H_t}\right].
    \eqlabel{eq:gaussian-derivatives-estimates}
\end{align}
Together, the pair $(g_t,H_t)$ is an unbiased estimate of $(\bar{g}_t,\bar{H}_t)$, which are the gradient and Hessian of the smoothed loss $F_{t,\Sigma_t}(\mu_t)$.
The algorithm uses these two estimates to form a \textit{shifted and damped second-order approximation} of $F_{t,\Sigma_t}$ at $\mu_t$.
Precisely, define
\begin{align}
    q_t(y)
    &\triangleq\ip{g_t}{y-\mu_t}
    +\frac\lambda2\normm{y-\mu_t}{H_t}^2\,,
    \quad
    \text{and}\\
    \bar{q}_t(y) &\triangleq
    \ip{\bar{g}_t}{y-\mu_t} + \frac\lambda2\normm{y-\mu_t}{\bar{H}_t}^2
    =
    \E_t[q_t(y)]
    ,
    \eqlabel{eq:quadratic-model}
\end{align}
for a parameter $\lambda\in(0,1)$ that damps the curvature of the surrogate.
Note that these quadratics are shifted in the sense that we dropped the constant term $F_{t,\Sigma_t}(\mu_t)=\E_t[Y_t]$ from both $q_t$ and $\bar{q}_t$.
The result is that $q_t(\mu_t)=\bar{q}_t(\mu_t)=0$.
Note that the algorithm has access to $q_t$ which we will refer to as the \textit{quadratic surrogate} of the loss in round $t$.
Note that $q_t$ differs from the second-order
Taylor polynomial of $F_{t,\Sigma_t}$ at $\mu_t$ in three ways: it is noisy through both observation noise $\xi_t$
and sampling noise $X_t\sim\N(\mu_t,\Sigma_t)$,
it is shifted by a constant so that it vanishes at $\mu_t$, and its curvature
is damped by the factor $\lambda$.
Its conditional mean $\bar{q}_t$ is
free of noise, but it is shifted and damped in the same way.
It is also important to note that $\bar{q}_t$ is convex as $F_{t,\Sigma_t}$ is convex (so $\bar{H}_t\succeq0$) while $q_t$ is not necessarily convex.

The upshot of the whole process is that the algorithm turns the single observation pair $(X_t,Y_t)$
into an entire quadratic function $q_t$ which preserves useful information about the loss $f_t$.
In summary, the algorithm
(i) extends the loss $f_t$ to $e_t$,
(ii) smooths $e_t$ to obtain $F_{t,\Sigma_t}$,
(iii) forms the shifted and damped second-order approximation $\bar{q}_t$ of $F_{t,\Sigma_t}$ at $\mu_t$, and
(iv) uses the single pair $(X_t,Y_t)$ to form an unbiased estimate $q_t$ of $\bar{q}_t$.

\paragraph{The Online Newton Step (ONS).}
With $q_t$ in hand, we update the mean and covariance by a projected online
Newton step \citep[Section~10.2]{lattimore2026bandit}:
\begin{align}
    \Sigma_{t+1}^{-1}&=\Sigma_t^{-1}+\frac{\eta\lambda}{2}H_t,\quad\text{and}\quad
    \mu_{t+1}=\argmin_{u\in K_t}
    \normm{u-\bigl(\mu_t-\eta\Sigma_{t+1}g_t\bigr)}{\Sigma_{t+1}^{-1}}^2.
    \eqlabel{eq:updates}
\end{align}
The projection keeps the next mean inside the focus region $K_t$.
Without the projection, these updates are continuous exponential weights with initial distribution $\N(0,\sigma^2I_d)$ and learning rate $\eta$, run on the quadratics $q_t$ with half their curvature \citep{vanderhoeven2018many}.
Note that the inverse covariance update uses only half of the quadratic
surrogate's curvature. The reason is rather technical as this leaves an extra negative
term in the analysis, which the restart mechanism relies on.
Also, $H_t$ may be indefinite, so before applying these updates the
algorithm checks that the proposed inverse covariance $\Sigma_{t+1}^{-1}$ is positive definite.
If the check fails (line~7 of Algorithm~\ref{alg:main}),
the algorithm plays $0$ for all remaining rounds. We show that this happens
with very small probability, so it does not affect the regret bound.

\paragraph{The focus region and the restart test.}
The quadratic surrogate is a faithful proxy for $e_t$ only near the current mean, so the algorithm keeps its mean in the \textit{focus region} $K_t=K_{t-1}\cap E_t$, where $K_0=\K_\eps$ and
\begin{align}
    E_t=\bigl\{x\in\R^d:\normm{x-\mu_t}{\Sigma_t^{-1}}^2\leq R^2\bigr\}.
\end{align}
The algorithm runs in epochs. Each epoch starts from $\mu_1=0$, $\Sigma_1=\sigma^2I_d$ and $K_0=\K_\eps$, with the round index starting again from one, and ends when the following test fires.
Let $Q_t=\sum_{s=1}^tq_s$. Since $q_s(\mu_s)=0$, the regret of the means $\mu_1,\ldots,\mu_t$ on $q_1,\ldots,q_t$ against a point $x$ is $-Q_t(x)$.
The test checks whether this regret is very negative against every point of $K_t$, that is, whether $\min_{x\in K_t}Q_t(x)$ is large.
Since $Q_t$ need not be convex, the algorithm works instead with
\begin{align}
    J_t(x)=\eta Q_t(x)+\normm{x}{\Sigma_1^{-1}}^2,
\end{align}
whose Hessian is $2\Sigma_{t+1}^{-1}\succ0$ by \cref{eq:updates}. It finds a point $y_t$ with
\begin{align}
    y_t\in K_t
    \qquad\text{and}\qquad
    J_t(y_t)\leq\min_{x\in K_t}J_t(x)+\frac\rho8,
    \eqlabel{eq:restart}
\end{align}
and restarts if $\eta Q_t(y_t)\geq2\rho$.
In \cref{pop:section} we show why an epoch that ends in a restart
has nonpositive regret against every point of $\K_\eps$.

\paragraph{Computation.}
We assume that the gauge $\pi$ of $\K$ can be evaluated, which is needed to
compute $A_t$ and $Y_t$. Besides this, each round solves two convex problems
over $K_t$: the projection in \cref{eq:updates} and the minimization in
\cref{eq:restart}. Both have strongly convex quadratic objectives, and $K_t$
is the intersection of $\K_\eps$ with at most $t$ ellipsoids, so both can be
solved in time polynomial in $d$ and $t$.

\begin{theorem}
    \label{thm:main}
    Suppose that $\mathbb B_1\subseteq\K\subseteq\mathbb B_{2d}$.
    Let $\delta\in(0,1/2)$, let $A$ be a sufficiently large universal
    constant, and let $L=100\log(100nd/\delta)$. Set
    \begin{align}
        \sigma=\frac{1}{M\sqrt d\,L^2}
        \qquad\text{and}\qquad
        \eta=\frac{1}{A\sigma L^3}\sqrt{\frac dn},
        \eqlabel{eq:parameters-free}
    \end{align}
    and from these define
    \begin{align}
        \rho=A^3\Bigl(\frac d\sigma\Bigr)^2,
        \qquad
        \lambda=\frac{1}{L\sqrt{A\rho}},
        \qquad
        R^2=\frac{1}{32\lambda^2L},
        \qquad
        \eps=\frac{\rho}{\eta n}.
        \eqlabel{eq:parameters-derived}
    \end{align}
    If $\eps\leq1/2$, run \cref{alg:main} with these parameters.
    Otherwise play a fixed action in $\K$.
    With probability at least $1-\delta$,
    \begin{align}
        \Reg_n\leq\frac{3\rho}{\eta}=3A^4d^2ML^5\sqrt n .
        \ineqlabel{eq:main-regret}
    \end{align}
\end{theorem}
We prove \cref{thm:main} in \cref{transfer:section}, building on the analysis of a population version of the algorithm in \cref{pop:section}.

The bound in \cref{thm:main} depends on the position of $\K$ through
$M$. An invertible affine change of coordinates preserves regret: the
learner transforms its actions, and the losses remain convex and bounded.
\Cref{prop:positioning} in \cref{app:positioning} shows that, for every
convex body, randomized affine preprocessing can produce coordinates
satisfying the ball inclusions in \cref{thm:main} and
$M\leq C\sqrt{\log(ed)/d}$ for a universal constant $C\geq1$.
Given a separation oracle and known inner and outer ball bounds as in
\cref{prop:positioning}, this preprocessing takes polynomial time.
Using failure probability $\delta/2$ for both preprocessing and the
algorithm, a union bound and \cref{thm:main} give, with probability at
least $1-\delta$,
\begin{align}
    \Reg_n=O\!\left(d^{3/2}\sqrt{\log(ed)}\,L^5\sqrt n\right)
    =\widetilde O(d^{3/2}\sqrt n)
\end{align}
for every convex body $\K$.
\FloatBarrier

\section{The Population Algorithm}
\label{pop:section}
\suppressfloats[t]
Throughout the regret analysis, we use the geometric assumptions of
\cref{thm:main}.
In this section we analyze a hypothetical \textit{population algorithm} that has access to the population gradient $\bar g_t = \E_t[g_t]$ and Hessian $\bar H_t = \E_t[H_t]$ in each round $t$.
Of course, such an algorithm is not implementable, but it serves as a useful theoretical tool for understanding the behavior of the actual sampled algorithm.
Our goal is to show that the population algorithm satisfies the regret bound of \cref{thm:main} in expectation, instead of with high probability.
Because the population algorithm uses the exact gradient and Hessian, there is no estimation error to control. This makes the analysis simpler, and it shows the main reason the sampled algorithm works without the extra work of concentration inequalities.
The population analysis does not see the size of the estimates $g_t$ and
$H_t$, which is much larger than that of their means. It therefore does not
tell the whole story, and in the sampled algorithm this size could cost
extra factors of $d$. \Cref{transfer:section} shows that, with our
parameters, it does not, and the sampled algorithm has the same rate.

Precisely, the population algorithm is \cref{alg:main} with $g_t$ and $H_t$ replaced by $\bar g_t$ and $\bar H_t$, and with this change carried through every quantity built from them.
For instance, the restart test \cref{eq:restart} uses $\bar Q_t=\sum_{s=1}^t\bar q_s$ in place of $Q_t$.
Throughout this section the index $t$ starts at one in each epoch, except in statements explicitly summed over all $n$ rounds.

\begin{proposition}[Population regret]\label{pop:oracle}
    Suppose the parameters are as in \cref{thm:main} and $\eps\leq 1/2$. Then, for the population algorithm and for any sequence of losses $(f_t)_{t=1}^n$, it holds that
    \begin{align}
        \eReg_n \triangleq
        \sup_{y\in\K_\eps}\sum_{t=1}^n\bigl(\E_t[e_t(X_t)]-e_t(y)\bigr)\leq \frac{\rho}{\eta}\,.
    \end{align}
\end{proposition}
We call $\eReg_n$ the \textit{extension regret}. It differs from the regret $\Reg_n$ in two ways: it evaluates the extensions $e_t$ at $X_t$ instead of $f_t$ at $A_t$, and it compares against points of $\K_\eps$ instead of $\K$.
Still, a bound on $\eReg_n$ gives a bound on $\E[\Reg_n]$. Taking expectations in \cref{eq:extension-cost}, and using that $e_t(X_t)-\E_t[e_t(X_t)]$ has mean zero, gives $\E[\Reg_n]\leq\E[\eReg_n]+n\eps$.
Since each $e_t$ is Lipschitz and $\K_\eps$ is compact, the supremum in $\eReg_n$ is attained. We fix
\begin{align}
    x_\star\in\argmin_{u\in\K_\eps}\sum_{t=1}^n e_t(u),
\end{align}
so that $\eReg_n=\sum_{t=1}^n\bigl(\E_t[e_t(X_t)]-e_t(x_\star)\bigr)$. This is a single comparator for all $n$ rounds, used across restarts.

\popstep{Relating extension regret to quadratic regret.}\label{pop:step-relate}
Looking closely at the population algorithm, the only information about the losses that it keeps is $\bar g_t$ and $\bar H_t$, and together these define the quadratic $\bar q_t$. Its updates and its restart test depend on the losses only through $\bar q_1,\ldots,\bar q_t$.
So the losses pass through three stages: the adversary chooses $f_t$, the algorithm extends it to $e_t$, and $e_t$ is then replaced by its quadratic approximation $\bar q_t$ at $\mu_t$.
The algorithm knows $\bar q_t$ completely, and on these quadratics it runs a variant of Online Newton step, just as in online learning with full information.
A natural first step is therefore to relate the extension regret $\eReg_n$, which we saw controls the actual regret, to the (full-information) regret on the quadratics.
Within an epoch, the \textit{quadratic regret} of the means against a point
$y\in\R^d$ is
\begin{align}
    \qRegbar_t(y)
    \triangleq\sum_{s=1}^t\bigl(\bar q_s(\mu_s)-\bar q_s(y)\bigr)
    =-\bar Q_t(y),
    \eqlabel{pop:qreg}
\end{align}
where the equality uses $\bar q_s(\mu_s)=0$.
The link between the two regrets is the following lemma, proved in \cref{sw:section}.
\begin{restatable}[Regret switching]{lemma}{regretswitching}
    \label{sw:tangent}
    Suppose that $\eps\leq1/2$, $0\prec\Sigma_t\preceq4\sigma^2 I_d$ and $\mu_t\in\K_\eps$. With the parameters of \cref{thm:main}, for every $x_0\in E_t$ and every $x_\star\in\K_\eps$,
    \begin{align}
        F_{t,\Sigma_t}(\mu_t)-e_t(x_\star)
        \leq \bar q_t(\mu_t)-\bar q_t(x_\star)
        +\frac{\lambda}{2}\normm{x_0-x_\star}{\bar H_t}^2
        +2\tr(\bar H_t\Sigma_t)+n^{-3}.
    \end{align}
\end{restatable}
Let us pause to discuss this lemma. First, its hypotheses hold at every round of the population algorithm. We assume $\eps\leq1/2$. Each $\bar H_t$ is positive semidefinite, so the covariance only shrinks within an epoch and $\Sigma_t\preceq\Sigma_1=\sigma^2I_d$. Finally, $\mu_t\in\K_\eps$ because the mean update projects onto the focus region, which lies in $\K_\eps$.

The lemma bounds the extension regret of round $t$ against $x_\star$ by the quadratic regret against $x_\star$, plus three extra terms. The term $n^{-3}$ is negligible. After summing over rounds, the usual log-determinant argument controls the trace term. The remaining term, $\frac\lambda2\normm{x_0-x_\star}{\bar H_t}^2$, needs more care. The comparator $x_\star$ can be any point of $\K_\eps$, but $x_0$ has to lie in the current focus ellipsoid $E_t$, and we pay for the distance between $x_0$ and $x_\star$ in the norm given by $\bar H_t$. Since $x_0$ is free, the question is how to choose it, and we turn to this next.

\popstep{Choosing $x_0$ in the focus region.}\label{pop:step-choose}
Fix an epoch and a round $t$ in it, so that the restart test has not been triggered in rounds $1,\ldots,t-1$. Since $K_t\subseteq E_s$ for every $s\leq t$, a single point $x_0\in K_t$ can be used in \cref{sw:tangent} in every round $s\leq t$. Write
\begin{align}
    c_s=2\tr(\bar H_s\Sigma_s)+n^{-3}
\end{align}
for the trace term and the negligible term together. We show later the bound on $\sum_{s=1}^t c_s$.
Summing \cref{sw:tangent} over $s=1,\ldots,t$ gives
\begin{align}
    \sum_{s=1}^t\bigl(F_{s,\Sigma_s}(\mu_s)-e_s(x_\star)\bigr)
    \leq \qRegbar_t(x_\star)
    +\frac{\lambda}{2}\sum_{s=1}^t\normm{x_0-x_\star}{\bar H_s}^2
    +\sum_{s=1}^t c_s.
\end{align}
Since $\bar Q_t$ is a quadratic with Hessian $\lambda\sum_{s=1}^t\bar H_s$, the middle term equals
\begin{align}
    \frac{\lambda}{2}\sum_{s=1}^t\normm{x_0-x_\star}{\bar H_s}^2
    =\bar Q_t(x_\star)-\bar Q_t(x_0)-\ip{\nabla\bar Q_t(x_0)}{x_\star-x_0}.
\end{align}
By \cref{pop:qreg}, $\qRegbar_t(x_\star)=-\bar Q_t(x_\star)$, so the first two terms on the right combine to give
\begin{align}
    \sum_{s=1}^t\bigl(F_{s,\Sigma_s}(\mu_s)-e_s(x_\star)\bigr)
    \leq \qRegbar_t(x_0)
    -\ip{\nabla\bar Q_t(x_0)}{x_\star-x_0}
    +\sum_{s=1}^t c_s.
    \ineqlabel{pop:switch-sum}
\end{align}
The quadratic regret against $x_\star$ has become the quadratic regret against $x_0$, a point in the focus region.
The algorithm controls this regret: it runs a variant of exponential weights on the quadratics $\bar q_s$, which it sees in full, and we bound this regret in \cref{pop:step-ons}. That leaves the inner product. Because $\bar Q_t$ is convex, whenever $x_0$ minimizes $\bar Q_t$ over $\K_\eps$ the inner product is nonnegative, so the term $-\ip{\nabla\bar Q_t(x_0)}{x_\star-x_0}$ can be dropped.

This suggests choosing $x_0$ as a minimizer of the cumulative quadratic. Since $x_0$ must lie in $K_t$, we take
\begin{align}
    x^q_t\in\argmin_{u\in K_t}\bar Q_t(u).
\end{align}
To see why this is a good choice, suppose for the moment that $x^q_t$ also minimizes $\bar Q_t$ over all of $\K_\eps$, then the inner product term can be dropped, and the extension regret over rounds $1,\ldots,t$ is at most $\qRegbar_t(x^q_t)+\sum_{s=1}^tc_s$, and both terms are under control.
Later, we will show that it does minimize $\bar Q_t$ over $\K_\eps$ as long as the epoch continues.

\popstep{Bounding the quadratic regret of the online Newton step.}\label{pop:step-ons}
We now bound the quadratic regret $\qRegbar_t(x_0)$ against points $x_0\in K_t$.
The Hessian of $\bar q_t$ is $\lambda\bar H_t$, but the update of the inverse covariance adds only $\eta\lambda\bar H_t/2$. As a result, the usual bound for the online Newton step comes with an extra negative term which will be useful later.
\begin{lemma}\label{pop:ons-lemma}
    For every $x_0\in K_t$, the population algorithm satisfies
    \begin{align}
        \qRegbar_t(x_0)
        \leq \frac{\normm{\mu_1 - x_0}{\Sigma_1^{-1}}^2
        -\normm{\mu_{t+1}-x_0}{\Sigma_{t+1}^{-1}}^2}{2\eta}
        +\frac{\eta}{2}\sum_{s=1}^t\normm{\bar g_s}{\Sigma_{s+1}}^2
        -\frac{\lambda}{4}\sum_{s=1}^t\normm{x_0-\mu_s}{\bar H_s}^2.
        \ineqlabel{pop:ons-bound}
    \end{align}
\end{lemma}
\begin{proof}
    Fix $s\leq t$. Since $K_t\subseteq K_s$, the point $x_0$ lies in $K_s$, and $\mu_{s+1}$ is the projection of $\mu_s-\eta\Sigma_{s+1}\bar g_s$ onto $K_s$ in the norm $\normm{\cdot}{\Sigma_{s+1}^{-1}}$. Hence
    \begin{align}
        \frac12\normm{\mu_{s+1}-x_0}{\Sigma_{s+1}^{-1}}^2
        &\explan{\leq}{a}\frac12\normm{\mu_s-\eta\Sigma_{s+1}\bar g_s-x_0}{\Sigma_{s+1}^{-1}}^2\nonumber\\
        &=\frac12\normm{\mu_s-x_0}{\Sigma_{s+1}^{-1}}^2
        -\eta\ip{\bar g_s}{\mu_s-x_0}
        +\frac{\eta^2}{2}\normm{\bar g_s}{\Sigma_{s+1}}^2\nonumber\\
        &\explan{=}{b}\frac12\normm{\mu_s-x_0}{\Sigma_s^{-1}}^2
        +\frac{\eta\lambda}{4}\normm{x_0-\mu_s}{\bar H_s}^2
        -\eta\ip{\bar g_s}{\mu_s-x_0}
        +\frac{\eta^2}{2}\normm{\bar g_s}{\Sigma_{s+1}}^2\nonumber\\
        &\explan{=}{c}\frac12\normm{\mu_s-x_0}{\Sigma_s^{-1}}^2
        -\eta\bigl(\bar q_s(\mu_s)-\bar q_s(x_0)\bigr)
        -\frac{\eta\lambda}{4}\normm{x_0-\mu_s}{\bar H_s}^2
        +\frac{\eta^2}{2}\normm{\bar g_s}{\Sigma_{s+1}}^2.
    \end{align}
    Here \reason{a} holds because projecting onto the convex set $K_s$ does not increase the distance to $x_0\in K_s$, \reason{b} uses $\Sigma_{s+1}^{-1}-\Sigma_s^{-1}=\eta\lambda\bar H_s/2$, and \reason{c} uses \cref{eq:quadratic-model} and $\bar q_s(\mu_s)=0$. Summing over $s=1,\ldots,t$ telescopes the distance terms, and rearranging with $\mu_1=0$ gives the claim.
\end{proof}

We now combine the two lemmas. Taking $x_0=x^q_t\in K_t$ in \cref{pop:switch-sum} and bounding $\qRegbar_t(x^q_t)$ by \cref{pop:ons-lemma} gives
\begin{align}
    \sum_{s=1}^t\bigl(F_{s,\Sigma_s}(\mu_s)-e_s(x_\star)\bigr)
    &\leq \frac{\normm{x^q_t}{\Sigma_1^{-1}}^2
    -\normm{\mu_{t+1}-x^q_t}{\Sigma_{t+1}^{-1}}^2}{2\eta}
    +\frac{\eta}{2}\sum_{s=1}^t\normm{\bar g_s}{\Sigma_{s+1}}^2
    +\sum_{s=1}^tc_s\nonumber\\
    &\quad-\frac{\lambda}{4}\sum_{s=1}^t\normm{x^q_t-\mu_s}{\bar H_s}^2
    -\ip{\nabla\bar Q_t(x^q_t)}{x_\star-x^q_t}.
    \ineqlabel{pop:combined}
\end{align}
The first line contains the terms of the usual bound for the online Newton step, together with $\sum_{s=1}^tc_s$. The two terms on the second line are new.
The first is nonpositive term we got from \cref{pop:ons-lemma} and the second is nonpositive whenever $x^q_t$ minimizes $\bar Q_t$ over $\K_\eps$ as we discussed in \cref{pop:step-choose}.

\popstep{$x^q_t$ is a global minimizer.}\label{pop:step-global}
Suppose first that $x^q_t$ minimizes $\bar Q_t$ over all of $\K_\eps$. Then the inner product in \cref{pop:combined} is nonnegative, so the last term there is nonpositive. We drop it, along with the other two nonpositive terms.
By $c_s=2\tr(\bar H_s\Sigma_s)+n^{-3}$ we have
\begin{align}
    \sum_{s=1}^t\bigl(F_{s,\Sigma_s}(\mu_s)-e_s(x_\star)\bigr)
    \leq \frac{\normm{x^q_t}{\Sigma_1^{-1}}^2}{2\eta}
    +\frac{\eta}{2}\sum_{s=1}^t\normm{\bar g_s}{\Sigma_{s+1}}^2
    +2\sum_{s=1}^t\tr(\bar H_s\Sigma_s)
    +tn^{-3}.
    \ineqlabel{pop:easy}
\end{align}
The first term is at most $2d^2/(\eta\sigma^2)$, since $\Sigma_1=\sigma^2I_d$ and $\|x^q_t\|\leq2d$ because $\K_\eps\subseteq\K\subseteq\mathbb B_{2d}$. The other two sums are controlled by the following lemma, which we prove in \cref{pop:appendix}.
\begin{lemma}\label{pop:sums}
    For every epoch of the population algorithm and every round $t$ in it,
    \begin{align}
        \sum_{s=1}^t\tr(\bar H_s\Sigma_s)\leq\frac{16dL}{\lambda\eta}
        \qquad\text{and}\qquad
        \sum_{s=1}^t\normm{\bar g_s}{\Sigma_{s+1}}^2\leq5n.
    \end{align}
\end{lemma}
Substituting these bounds into \cref{pop:easy} and using $t\leq n$, we get, whenever $x^q_t$ minimizes $\bar Q_t$ over $\K_\eps$,
\begin{align}
    \sum_{s=1}^t\bigl(F_{s,\Sigma_s}(\mu_s)-e_s(x_\star)\bigr)
    \leq \frac{2d^2}{\eta\sigma^2}+\frac{5\eta n}{2}+\frac{32dL}{\lambda\eta}+n^{-2}
    \leq\frac{\rho}{\eta},
\end{align}
where the last inequality uses \cref{eq:parameters-free,eq:parameters-derived}.

This is the bound of \cref{pop:oracle}, but only under the assumption that $x^q_t$ minimizes $\bar Q_t$ over $\K_\eps$.
If $x^q_t$ does not minimize $\bar Q_t$ over $\K_\eps$, then it lies on the boundary of some focus ellipsoid $E_r$ with $r\leq t$.
Indeed, otherwise $x^q_t$ lies in the interior of each of $E_1,\ldots,E_t$, so $K_t$ and $\K_\eps$ coincide near $x^q_t$. Then $x^q_t$ is a local minimizer of $\bar Q_t$ over $\K_\eps$, and since $\bar Q_t$ is convex, it minimizes $\bar Q_t$ over $\K_\eps$.
We turn to this case next.

\popstep{Touching a focus ellipsoid forces a restart.}\label{pop:step-touch}
We now prove that if $x^q_t$ lies on the boundary of an earlier focus ellipsoid $E_r$, then the restart test fires at round $t$.
In \cref{pop:step-restart} we show that an epoch that ends in a restart has nonpositive extension regret.
Together with \cref{pop:step-global}, this covers every epoch.
It is worth mentioning that we do not need to show that every restart is triggered by $x^q_t$ touching a focus ellipsoid.
The algorithm may restart for other reasons, and the analysis still holds.

Concretely, suppose that $x^q_t$ lies on the boundary of $E_r$ for some $r\leq t$, then
we prove that $\qRegbar_t(x^q_t) = -\bar Q_t(x^q_t)$ is too negative, so $\bar Q_t(y_t)\geq\bar Q_t(x^q_t)$ is too large, and the test fires.
First, the quadratic regret against $x^q_t \in \K_t$ is bounded by \cref{pop:ons-lemma} at $x_0=x^q_t$:
\begin{align}
    -\bar Q_t(x^q_t)=\qRegbar_t(x^q_t)
    \leq \frac{\normm{x^q_t-\mu_1}{\Sigma_1^{-1}}^2}{2\eta}
    +\frac{\eta}{2}\sum_{s=1}^t\normm{\bar g_s}{\Sigma_{s+1}}^2
    -\frac{\lambda}{4}\sum_{s<r}\normm{x^q_t-\mu_s}{\bar H_s}^2,
    \label{eq:pop-quad-reg}
\end{align}
where we deliberately dropped the nonnegative terms in the last sum over $s \in [r,t]$.
On the other hand, from $x^q_t \in \partial E_r$ we get
\begin{align}
    R^2
    = \normm{x^q_t - \mu_r}{\Sigma_r^{-1}}^2
    = \normm{x^q_t - \mu_r}{\Sigma_1^{-1}}^2
    + \frac{\eta\lambda}{2}\sum_{s<r}\normm{x^q_t - \mu_r}{\bar H_s}^2.
    \label{eq:pop-ellipsoid-boundary}
\end{align}
By adding and subtracting \cref{eq:pop-ellipsoid-boundary} divided by $2\eta$ to the RHS of \cref{eq:pop-quad-reg}, we get
\begin{align}
    \qRegbar_t(x^q_t)
    &\leq
    -\frac{R^2}{2\eta}
    + \frac{
        \normm{x^q_t-\mu_1}{\Sigma_1^{-1}}^2 +
        \normm{x^q_t-\mu_r}{\Sigma_1^{-1}}^2
    }{2\eta}
    +\frac{\eta}{2}\sum_{s=1}^t\normm{\bar g_s}{\Sigma_{s+1}}^2
    \\
    &\quad+\frac{\lambda}{4}\sum_{s<r}\left( \normm{x^q_t-\mu_r}{\bar H_s}^2 - \normm{x^q_t-\mu_s}{\bar H_s}^2 \right) \\
    &\leq
    -\frac{R^2}{2\eta}
    +
    \frac{(4d)^2}{\eta\sigma^2}
    + \frac{5\eta n}{2}
    +\frac{\lambda}{4}\sum_{s<r}
    \Bigl(\frac12\normm{x^q_t-\mu_r}{\bar H_s}^2+\normm{\mu_r-\mu_s}{\bar H_s}^2\Bigr)\\
    &\leq-\frac{R^2}{4\eta}
    +\frac{(4d)^2}{\eta\sigma^2}
    +\frac{5\eta n}{2}
    +\frac{\lambda}{4}\sum_{s<r}\normm{\mu_s-\mu_r}{\bar H_s}^2,
    \ineqlabel{pop:boundary-movement}
\end{align}
where the second inequality uses \cref{pop:sums} together with $x^q_t, \mu_1, \mu_r \in \K_\eps \subseteq \K \subseteq \mathbb B_{2d}$,
and the inequality $\normm{x^q_t-\mu_s}{\bar H_s}^2 + \normm{\mu_r-\mu_s}{\bar H_s}^2 \geq \frac12\normm{x^q_t-\mu_r}{\bar H_s}^2$;
the last inequality uses \cref{eq:pop-ellipsoid-boundary} which shows that $\frac{\lambda}{8}\sum_{s<r}\normm{x^q_t-\mu_r}{\bar H_s}^2\leq\frac{R^2}{4\eta}$.
It remains to bound the last sum, which essentially measures how far the means $\mu_s$ have moved before round $r$.
The means cannot have moved far before round $r$, because the restart test did not fire at round $r-1$. A short calculation in \cref{pop:appendix}, which applies \cref{pop:ons-lemma} at round $r-1$, makes this precise:
\begin{align}
    \frac{\lambda}{4}\sum_{s<r}\normm{\mu_s-\mu_r}{\bar H_s}^2
    \leq\frac{1}{\eta}\Bigl(\frac{4d^2}{\sigma^2}+4\rho\Bigr)+5\eta n.
    \ineqlabel{pop:movement}
\end{align}
Substituting this into \cref{pop:boundary-movement} and multiplying by $\eta$,
\begin{align}
    \eta\,\qRegbar_t(x^q_t)
    \leq-\frac{R^2}{4}+\frac{20d^2}{\sigma^2}+4\rho+\frac{15\eta^2n}{2}
    \leq-2\rho,
\end{align}
where the last inequality uses $d^2/\sigma^2=\rho/A^3$, $\eta^2n=\rho/(A^5dL^6)$ and $R^2\geq256\rho$. Finally, $y_t\in K_t$ and $x^q_t$ minimizes $\bar Q_t$ over $K_t$, so
\begin{align}
    \eta\bar Q_t(y_t)\geq\eta\bar Q_t(x^q_t)=-\eta\,\qRegbar_t(x^q_t)\geq2\rho,
\end{align}
and the restart test fires at round $t$.
Consequently, if the restart test does not fire in rounds $1,\ldots,t$, then $x^q_t$ lies on none of the boundaries of $E_1,\ldots,E_t$, and so, by the end of \cref{pop:step-global}, it minimizes $\bar Q_t$ over $\K_\eps$.

\popstep{$x^q_t$ stays in the focus region.}\label{pop:step-stays}
We now prove that if round $t$ does not restart, then $x^q_t\in K_{t+1}$. We use this fact in the next step.
Since $x^q_t\in K_t$, it suffices to show $x^q_t\in E_{t+1}$. The test did not fire and $y_t\in K_t$, so $\eta\bar Q_t(x^q_t)\leq\eta\bar Q_t(y_t)<2\rho$. Applying \cref{pop:ons-lemma} with $x_0=x^q_t$, using $\qRegbar_t(x^q_t)=-\bar Q_t(x^q_t)$, and dropping the curvature term gives
\begin{align}
    \frac12\normm{\mu_{t+1}-x^q_t}{\Sigma_{t+1}^{-1}}^2
    \leq\frac12\normm{x^q_t}{\Sigma_1^{-1}}^2
    +\frac{\eta^2}{2}\sum_{s=1}^t\normm{\bar g_s}{\Sigma_{s+1}}^2
    +\eta\bar Q_t(x^q_t)
    \leq\frac{2d^2}{\sigma^2}+\frac{5\eta^2n}{2}+2\rho
    \leq\frac{5\rho}{2},
\end{align}
where we used $\|x^q_t\|\leq2d$, \cref{pop:sums}, and $2d^2/\sigma^2+5\eta^2n/2\leq\rho/2$, which follows from \cref{eq:parameters-free,eq:parameters-derived}. Hence $\normm{\mu_{t+1}-x^q_t}{\Sigma_{t+1}^{-1}}^2\leq5\rho<R^2$, that is, $x^q_t\in E_{t+1}$.

\popstep{An epoch that ends in a restart has nonpositive regret.}\label{pop:step-restart}
Suppose round $t$ ends the epoch in a restart, and let $z=x^q_{t-1}$, with $z=0$ if $t=1$. Rounds $1,\ldots,t-1$ did not restart, so by \cref{pop:step-touch} the point $z$ minimizes $\bar Q_{t-1}$ over $\K_\eps$, and by \cref{pop:step-stays} $z \in K_t$.

For the first $t-1$ rounds, apply \cref{pop:switch-sum} at round $t-1$ with $x_0=z$. The inner product there is nonnegative because $z$ minimizes $\bar Q_{t-1}$ over $\K_\eps$, so
\begin{align}
    \sum_{s=1}^{t-1}\bigl(F_{s,\Sigma_s}(\mu_s)-e_s(x_\star)\bigr)
    \leq \qRegbar_{t-1}(z)+\sum_{s=1}^{t-1}c_s.
    \ineqlabel{pop:first-rounds}
\end{align}
For the last round, apply \cref{sw:tangent} with both points equal to $z$, which removes the quadratic term. Since $e_t$ takes values in $[0,1]$ on $\K_\eps$, we have
\begin{align}
    F_{t,\Sigma_t}(\mu_t)-e_t(x_\star)
    =F_{t,\Sigma_t}(\mu_t)-e_t(z)+e_t(z)-e_t(x_\star)
    \leq \bar q_t(\mu_t)-\bar q_t(z)+c_t+1.
    \ineqlabel{pop:last-round}
\end{align}
Adding the two bounds, and using $\qRegbar_t(z)=-\bar Q_t(z)$, gives
\begin{align}
    \sum_{s=1}^{t}\bigl(F_{s,\Sigma_s}(\mu_s)-e_s(x_\star)\bigr)
    \leq -\bar Q_t(z)+\sum_{s=1}^{t}c_s+1.
    \ineqlabel{pop:restart-sum}
\end{align}
It remains to show that $\bar Q_t(z)$ is large, and the restart test gives exactly this. Since $z\in K_t$, the definition of $y_t$ in \cref{eq:restart} gives $J_t(y_t)\leq J_t(z)+\rho/8$, that is,
\begin{align}
    \eta\bar Q_t(z)
    \geq\eta\bar Q_t(y_t)+\normm{y_t}{\Sigma_1^{-1}}^2-\normm{z}{\Sigma_1^{-1}}^2-\frac{\rho}{8}
    \geq2\rho-\frac{4d^2}{\sigma^2}-\frac{\rho}{8}
    \geq\frac{7\rho}{4},
\end{align}
where the second inequality uses the restart condition $\eta\bar Q_t(y_t)\geq2\rho$ and $\|z\|\leq2d$, and the last uses $4d^2/\sigma^2\leq\rho/8$. By \cref{pop:sums} and \cref{eq:parameters-free,eq:parameters-derived}, $\eta\bigl(1+\sum_{s=1}^tc_s\bigr)\leq2\eta+32dL/\lambda\leq\rho/4$. Substituting both bounds into \cref{pop:restart-sum} gives
\begin{align}
    \eta\sum_{s=1}^{t}\bigl(F_{s,\Sigma_s}(\mu_s)-e_s(x_\star)\bigr)
    \leq-\frac{7\rho}{4}+\frac{\rho}{4}
    =-\frac{3\rho}{2}<0.
\end{align}
So every epoch that ends in a restart has negative extension regret against $x_\star$.

\popstep{Summing over epochs.}\label{pop:step-sum}
We now return to calendar time and prove \cref{pop:oracle}. The $n$ rounds split into epochs. By \cref{pop:step-restart}, every epoch that ends in a restart has negative extension regret against $x_\star$. If the last epoch does not end in a restart, let $t$ be its number of rounds. The restart test did not fire in any of these rounds, so by \cref{pop:step-touch} the point $x^q_t$ minimizes $\bar Q_t$ over $\K_\eps$, and \cref{pop:step-global} bounds the extension regret of this epoch by $\rho/\eta$. If the last round ends in a restart, there is no such epoch. The comparator $x_\star$ is the same in every epoch, and $\E_t[e_t(X_t)]=F_{t,\Sigma_t}(\mu_t)$, so adding these bounds gives
\begin{align}
    \eReg_n=\sum_{t=1}^n\bigl(F_{t,\Sigma_t}(\mu_t)-e_t(x_\star)\bigr)\leq\frac{\rho}{\eta},
\end{align}
which proves \cref{pop:oracle}.

\section{Discussion}
\label{sec:discussion}
Several questions remain open. The bound matches the rate conjectured by \citet{bubeck2017kernel}. Up to logarithmic factors, our bound exceeds the $\Omega(d^{4/3}\sqrt n)$ lower bound of \citet{rajaraman2026price} by a factor of $d^{1/6}$. Our bound also carries the factor $L^5$, and we do not know how far the logarithmic factors can be reduced. Finally, within an epoch the focus region is an intersection of up to $t$ ellipsoids, so the computation per round grows polynomially with $n$. \citet[Conjecture~1]{bubeck2017kernel} asks for regret $\widetilde O(d^{3/2}\sqrt n)$ with computation per round polynomial in $d$ and $\log n$. It would be interesting to know whether the algorithm can keep only $O(\log n)$ ellipsoids without losing the rate.

\paragraph{AI use and concurrent work.}
We date our solution to November~1, 2026, when GPT-6 Astra first judged our proposed proof of the $\widetilde O(d^{3/2}\sqrt n)$ regret bound to be correct. The proof grew out of a series of discussions with the model and drew on ideas and results from our earlier work on the problem.
The \href{https://github.com/mt0-svg/adversarial-bco/tree/2abc77c3989eb894409e892da475b6d7316882ed}{first version} of a concurrent manuscript by \citet{mt0svg2026optimal} reported the same rate in expectation for losses fixed in advance.
According to GitHub's activity records, we first pushed a draft containing our bound to \href{https://github.com/sarb9/cvxbyopt/activity/}{our repository} on November~9 at 10:50~p.m.\ PST. The first push to \href{https://github.com/mt0-svg/adversarial-bco/activity/}{the other manuscript's repository} was on November~10 at 3:36~p.m.\ PST.
We have not independently assessed the methods or verified the results of that manuscript.

\newpage
\begingroup
\raggedright
\providecommand{\doi}[1]{\href{https://doi.org/#1}{doi:\nolinkurl{#1}}}
\bibliography{references}
\endgroup

\newpage
\appendix
\crefalias{section}{appendix}

\section{Omitted proofs from the population argument}\label{pop:appendix}

\begin{proof}[Proof of \cref{pop:sums}]
    The hypothesis \cref{fb:comparison} of \cref{fb:trace,fb:budget} holds with equality for the population algorithm. The first bound is therefore \cref{fb:trace-sum}.

    For the second bound, fix $s\leq t$ and let $Z_s=\Sigma_s^{-1/2}(X_s-\mu_s)$, which is standard Gaussian given the past. Since $\E_s[Y_s\mid X_s]=e_s(X_s)$, \cref{eq:gaussian-derivatives-estimates} gives $\Sigma_s^{1/2}\bar g_s=\E_s[e_s(X_s)Z_s]$. The coordinates of $Z_s$ are orthonormal in $L^2$, and the $i$th coordinate of $\E_s[e_s(X_s)Z_s]$ is the inner product of $e_s(X_s)$ with the $i$th coordinate of $Z_s$. So Bessel's inequality gives
    \begin{align}
        \normm{\bar g_s}{\Sigma_s}^2=\bigl\|\E_s[e_s(X_s)Z_s]\bigr\|^2\leq\E_s\bigl[e_s(X_s)^2\bigr].
    \end{align}
    Since $\bar H_s\succeq0$, we have $\Sigma_{s+1}\preceq\Sigma_s$, and hence $\normm{\bar g_s}{\Sigma_{s+1}}\leq\normm{\bar g_s}{\Sigma_s}$. Moreover, $\E_s[e_s(X_s)^2]\leq\E_s[Y_s^2]$, because $Y_s=e_s(X_s)+\pi_\eps(X_s)\xi_s$ and the noise has conditional mean zero. Summing over $s$ and applying \cref{fb:budget} proves the second bound.
\end{proof}

\begin{proof}[Proof of \cref{pop:movement}]
    Recall the setting of \cref{pop:step-touch}: the restart test has not fired in rounds $1,\ldots,t-1$, and $x^q_t$ lies on the boundary of $E_r$ for some $r\leq t$. If $r=1$, the sum is empty. Otherwise let $z=x^q_{r-1}$, and split each term at $z$:
    \begin{align}
        \frac{\lambda}{4}\sum_{s<r}\normm{\mu_s-\mu_r}{\bar H_s}^2
        &\leq\frac{\lambda}{2}\sum_{s<r}\normm{\mu_r-z}{\bar H_s}^2
        +\frac{\lambda}{2}\sum_{s<r}\normm{z-\mu_s}{\bar H_s}^2\\
        &\leq\frac{1}{\eta}\normm{\mu_r-z}{\Sigma_r^{-1}}^2
        +\frac{\lambda}{2}\sum_{s<r}\normm{z-\mu_s}{\bar H_s}^2,
    \end{align}
    where the second inequality uses $\frac{\eta\lambda}{2}\sum_{s<r}\bar H_s=\Sigma_r^{-1}-\Sigma_1^{-1}\preceq\Sigma_r^{-1}$. The right side is what the online Newton step controls at round $r-1$. Indeed, $z\in K_{r-1}$, and \cref{pop:ons-lemma} at round $r-1$ with $x_0=z$, multiplied by two and rearranged, gives
    \begin{align}
        \frac{1}{\eta}\normm{\mu_r-z}{\Sigma_r^{-1}}^2
        +\frac{\lambda}{2}\sum_{s<r}\normm{z-\mu_s}{\bar H_s}^2
        \leq\frac{\normm{z-\mu_1}{\Sigma_1^{-1}}^2}{\eta}
        +\eta\sum_{s<r}\normm{\bar g_s}{\Sigma_{s+1}}^2
        +2\bar Q_{r-1}(z).
    \end{align}
    The restart test did not fire at round $r-1$, and $z$ minimizes $\bar Q_{r-1}$ over $K_{r-1}$, which contains $y_{r-1}$. So $\eta\bar Q_{r-1}(z)\leq\eta\bar Q_{r-1}(y_{r-1})<2\rho$. Since $\mu_1=0$, $\|z\|\leq2d$ and $\Sigma_1=\sigma^2I_d$, the first term on the right is at most $4d^2/(\eta\sigma^2)$, and by \cref{pop:sums} the second is at most $5\eta n$. This proves the claim.
\end{proof}

\section{Properties of the convex extension}
\label{app:extension}

\begin{proof}[Proof of \cref{lem:extension}]
    Fix a loss $f$ and omit the round index. By
    \citet[Proposition~3.23]{lattimore2026bandit}, the function
    \begin{align}
        x\longmapsto
        \pi_\eps(x)f(x/\pi_\eps(x))+\frac{1-\eps}{\eps}v(x)
    \end{align}
    is convex. Since $v$ is convex, adding $v$ or $(1+\eps)v/\eps$ shows
    that $e-v/\eps$ and $e$ are convex, respectively.
    The equality on $\K_\eps$ and the inequality
    $f(x/\pi_\eps(x))\leq e(x)$ follow from the definition, since
    $\pi_\eps\geq1$, $v\geq0$ and $f\geq0$.

    To bound the Lipschitz constant, note that
    $0\leq e(x)\leq(1+2/\eps)\max\{1,\|x\|/(1-\eps)\}$ since $f\leq1$
    and $\pi(x)\leq\|x\|$. For $g\in\partial e(x)$ and a unit vector $u$,
    the subgradient inequality at $x+ru$ gives
    \begin{align}
        \ip{g}{u}\leq\limsup_{r\to\infty}\frac{e(x+ru)-e(x)}r
        \leq\frac{2+\eps}{\eps(1-\eps)}\leq\frac5\eps.
    \end{align}
    Taking a supremum over $u$ proves the claim.
\end{proof}

\section{Regret switching}
\label{sw:section}

\let\savedlabel\label\regretswitching*\global\let\label\savedlabel

The restriction on $x_0$ controls how far the slope of the quadratic
can depart from $\bar g_t$. The comparator $x_\star$ may be any point of
$\K_\eps$, including points outside the focus region.

\begin{proof}[Proof of \cref{sw:tangent}]
    \begingroup\allowdisplaybreaks[1]
    Fix the information available before the sample and suppress the round
    index. Write $e=e_t$, $X\sim\N(\mu,\Sigma)$ with $\mu=\mu_t$ and
    $\Sigma=\Sigma_t$, and
    \begin{align}
        F=F_{t,\Sigma_t}(\mu_t)=\E e(X),\qquad
        \bar g=\bar g_t,\qquad \bar H=\bar H_t,\qquad \bar q=\bar q_t.
    \end{align}
    All expectations in this proof are conditional on that information.
    In particular, $\bar q$ is the population quadratic, with Hessian
    $\lambda\bar H$, and $\bar q(\mu)=0$.
    We first give the calculation for a smooth convex $e$, so that
    $\bar g=\E\nabla e(X)$ and $\bar H=\E\nabla^2e(X)$. At the end we
    justify the same argument for the possibly nonsmooth extension.

    The exact quadratic expansion
    \begin{align}
        \bar q(x_\star)
        =\bar q(x_0)+\ip{\nabla\bar q(x_0)}{x_\star-x_0}
        +\frac\lambda2\normm{x_\star-x_0}{\bar H}^2
        \eqlabel{sw:derivation-tangent}
    \end{align}
    shows that our goal is to prove
    \begin{align}
        F-e(x_\star)
        &\leq \bar q(\mu)
        -\bigl(\bar q(x_0)+\ip{\nabla\bar q(x_0)}{x_\star-x_0}\bigr)
        +2\tr(\bar H\Sigma)+n^{-3}.
        \ineqlabel{sw:derivation-goal}
    \end{align}
    We will use two consequences of the parameters:
    \begin{align}
        \lambda&=\frac{1}{A^2d^{3/2}ML^3}\leq\frac{1}{8(d+L)},\qquad
        \lambda\normm{x_0-\mu}{\Sigma^{-1}}
        \leq\lambda R=\frac{1}{\sqrt{32L}}\leq\frac{1}{\sqrt{8L}}.
        \ineqlabel{sw:derivation-conditions}
    \end{align}
    The first follows from $M\geq d^{-1/2}$, $L\geq100$ and sufficiently
    large $A$; the second follows from $x_0\in E_t$ and the definition of $R$.

    For every sample $X$, the definition of the Bregman divergence gives
    \begin{align}
        e(X)-e(x_\star)
        =\ip{\nabla e(X)}{X-\mu}+\ip{\nabla e(X)}{\mu-x_\star}-D_e(x_\star\,\|\,X).
        \eqlabel{sw:derivation-starting-point}
    \end{align}
    Taking expectations,
    \begin{align}
        \E\bigl[e(X)-e(x_\star)\bigr]
        &=\E\bigl[\ip{\nabla e(X)}{X-x_\star}-D_e(x_\star\,\|\,X)\bigr]\\
        &=\E\ip{\nabla e(X)}{X-\mu}+\E\ip{\nabla e(X)}{\mu-x_\star}
        -\E\bigl[D_e(x_\star\,\|\,X)\bigr]\\
        &=\tr(\bar H\Sigma)+\ip{\nabla \bar q(\mu)}{\mu-x_\star}
        -\E\bigl[D_e(x_\star\,\|\,X)\bigr].
    \end{align}

    The goal is to make $\nabla \bar q(x_0)$ appear in place of $\nabla \bar q(\mu)$.
    Note that $\nabla \bar q(\mu)=\bar g$ came from the term $\E\nabla e(X)$.
    The key step is that there is a function
    \begin{align}
        w(x) = 1+\lambda\ip{x_0-\mu}{\Sigma^{-1}(x-\mu)}
    \end{align}
    such that
    \begin{align}
        \E[w(X) \nabla e(X)] = \nabla \bar q(x_0).
        \eqlabel{sw:derivation-key}
    \end{align}
    Indeed, Gaussian integration by parts gives
    $\E[\nabla e(X)(X-\mu)^\top]=\E[\nabla^2e(X)]\Sigma=\bar H\Sigma$, and therefore
    \begin{align}
        \E[w(X)\nabla e(X)]
        &=\E\nabla e(X)+\lambda\E\bigl[\nabla e(X)(X-\mu)^\top\bigr]\Sigma^{-1}(x_0-\mu)
        \nonumber\\
        &=\bar g+\lambda \bar H(x_0-\mu)
        =\nabla \bar q(x_0).
    \end{align}
    Moreover, $\E w(X)=1$. The weight can be negative, so it is not a
    probability density. Its connection with a change of Gaussian measure
    explains the choice of $w$.
    \paragraph{Where the weight comes from.}
    The weight is the first-order Taylor expansion of the Gaussian density
    ratio. More precisely, consider $\N(\mu+\lambda(x_0-\mu),\Sigma)$, the
    Gaussian moved a fraction $\lambda$ of the way from $\mu$ toward $x_0$.
    Its density with respect to $\N(\mu,\Sigma)$ is
    \begin{align}
        \frac{d\N(\mu+\lambda(x_0-\mu),\Sigma)}{d\N(\mu,\Sigma)}(x)
        &=\exp\Bigl(\lambda\ip{x_0-\mu}{\Sigma^{-1}(x-\mu)}
        -\frac{\lambda^2}{2}\normm{x_0-\mu}{\Sigma^{-1}}^2\Bigr)\nonumber\\
        &=1+\lambda\ip{x_0-\mu}{\Sigma^{-1}(x-\mu)}+O(\lambda^2)\nonumber\\
        &=w(x)+O(\lambda^2).
    \end{align}
    Here the expansion is for fixed $x$ as $\lambda\to0$. Thus $w$ is the
    linear approximation to the density ratio for this shift. The proof
    below uses the affine weight exactly, without a Taylor remainder.

    Now that we know how to get $\nabla \bar q(x_0)$, we can multiply \cref{sw:derivation-starting-point} by $w(X)$ and take expectations.
    The left-hand side becomes
    \begin{align}
        \E\bigl[w(X)\bigl(e(X)-e(x_\star)\bigr)\bigr]
        &\explan{=}{a}\E\bigl[w(X)e(X)\bigr]-e(x_\star)\nonumber\\
        &\explan{=}{b}F+\lambda\ip{x_0-\mu}{\Sigma^{-1}\E\bigl[(X-\mu)e(X)\bigr]}-e(x_\star)\nonumber\\
        &\explan{=}{c}F-e(x_\star)+\lambda\ip{x_0-\mu}{\bar g}.
        \eqlabel{sw:derivation-lhs}
    \end{align}
    Here \reason{a} uses $\E w(X)=1$; \reason{b} expands $w$ and uses $F=\E e(X)$; and
    \reason{c} uses Gaussian integration by parts,
    $\E[(X-\mu)e(X)]=\Sigma\,\E\nabla e(X)=\Sigma \bar g$.

    The right-hand side becomes
    \begin{align}
        &\E\Bigl[w(X)\Bigl(\ip{\nabla e(X)}{X-\mu}+\ip{\nabla e(X)}{\mu-x_\star}-D_e(x_\star\,\|\,X)\Bigr)\Bigr]\nonumber\\
        &\explan{=}{a}\E\bigl[w(X)\ip{\nabla e(X)}{X-\mu}\bigr]+\ip{\nabla \bar q(x_0)}{\mu-x_\star}
        -\E\bigl[w(X)D_e(x_\star\,\|\,X)\bigr]\nonumber\\
        &\explan{=}{b}\tr(\bar H\Sigma)+\lambda\ip{x_0-\mu}{\Sigma^{-1}\E\bigl[(X-\mu)\ip{\nabla e(X)}{X-\mu}\bigr]}\nonumber\\
        &\qquad+\ip{\nabla \bar q(x_0)}{\mu-x_\star}-\E\bigl[w(X)D_e(x_\star\,\|\,X)\bigr]\nonumber\\
        &\explan{=}{c}\tr(\bar H\Sigma)+\lambda\ip{x_0-\mu}{\E\bigl[\nabla^2e(X)(X-\mu)+\nabla e(X)\bigr]}\nonumber\\
        &\qquad+\ip{\nabla \bar q(x_0)}{\mu-x_\star}-\E\bigl[w(X)D_e(x_\star\,\|\,X)\bigr]\nonumber\\
        &\explan{=}{d}\tr(\bar H\Sigma)+\lambda\E\bigl[(x_0-\mu)^\top\nabla^2e(X)(X-\mu)\bigr]+\lambda\ip{x_0-\mu}{\bar g}\nonumber\\
        &\qquad+\ip{\nabla \bar q(x_0)}{\mu-x_\star}-\E\bigl[w(X)D_e(x_\star\,\|\,X)\bigr].
        \eqlabel{sw:derivation-rhs}
    \end{align}
    Here \reason{a} uses \cref{sw:derivation-key}; \reason{b} expands
    $w$ and uses $\E\ip{\nabla e(X)}{X-\mu}=\tr(\bar H\Sigma)$, which is Gaussian
    integration by parts; \reason{c} applies Gaussian integration by parts to
    $\varphi(X)=\ip{\nabla e(X)}{X-\mu}$, whose gradient is
    $\nabla^2e(X)(X-\mu)+\nabla e(X)$; and \reason{d} uses $\E\nabla e(X)=\bar g$.

    Now, note that the term $\lambda\ip{x_0-\mu}{\bar g}$ is present in both \cref{sw:derivation-lhs} and \cref{sw:derivation-rhs}, so we can cancel it out,
    and the weighted version of \cref{sw:derivation-starting-point} becomes
    \begin{align}
        F-e(x_\star)
        &=
        \tr(\bar H\Sigma)+\lambda\E\bigl[(x_0-\mu)^\top\nabla^2e(X)(X-\mu)\bigr]\nonumber\\
        &\qquad+\ip{\nabla \bar q(x_0)}{\mu-x_\star}-\E\bigl[w(X)D_e(x_\star\,\|\,X)\bigr].
        \eqlabel{sw:derivation-weighted}
    \end{align}
    Next we make the quadratic at $x_0$ appear. Since $\bar q$ is a quadratic with
    Hessian $\lambda \bar H$, its expansion around $x_0$ is exact,
    $\bar q(\mu)=\bar q(x_0)+\ip{\nabla \bar q(x_0)}{\mu-x_0}+\frac\lambda2\normm{\mu-x_0}{\bar H}^2$.
    Therefore
    \begin{align}
        F-e(x_\star)
        &\explan{=}{a}\tr(\bar H\Sigma)+\lambda\E\bigl[(x_0-\mu)^\top\nabla^2e(X)(X-\mu)\bigr]\nonumber\\
        &\qquad+\ip{\nabla \bar q(x_0)}{\mu-x_0}-\ip{\nabla \bar q(x_0)}{x_\star-x_0}-\E\bigl[w(X)D_e(x_\star\,\|\,X)\bigr]\nonumber\\
        &\explan{=}{b}\bar q(\mu)-\bigl(\bar q(x_0)+\ip{\nabla \bar q(x_0)}{x_\star-x_0}\bigr)+\tr(\bar H\Sigma)\nonumber\\
        &\qquad
        +\lambda\E\bigl[(x_0-\mu)^\top\nabla^2e(X)(X-\mu)\bigr]-\frac\lambda2\normm{x_0-\mu}{\bar H}^2
        \nonumber\\
        &\qquad-\E\bigl[w(X)D_e(x_\star\,\|\,X)\bigr].
        \eqlabel{sw:derivation-quadratic}
    \end{align}
    Here \reason{a} uses \cref{sw:derivation-weighted} and splits
    $\mu-x_\star=(\mu-x_0)-(x_\star-x_0)$; and \reason{b} uses the expansion of $\bar q$
    around $x_0$.
    The rest of the proof is to upper bound the last three terms in \cref{sw:derivation-quadratic}.

    \proofstepinline{1} We bound the curvature term.
    First we address the following term in \cref{sw:derivation-quadratic}:
    \begin{align}
        \lambda\E\bigl[(x_0-\mu)^\top\nabla^2e(X)(X-\mu)\bigr]-\frac\lambda2\normm{x_0-\mu}{\bar H}^2.
        \label{sw:derivation-step1}
    \end{align}
    First, we have
    \begin{align}
        &\E\bigl[(x_0-\mu)^\top\nabla^2e(X)(X-\mu)\bigr]\nonumber\\
        &\quad\explan{\le}{a}\frac12\E\bigl[(x_0-\mu)^\top\nabla^2e(X)(x_0-\mu)\bigr]
        +\frac12\E\bigl[(X-\mu)^\top\nabla^2e(X)(X-\mu)\bigr]\nonumber\\
        &\quad\explan{=}{b}\frac12\normm{x_0-\mu}{\bar H}^2+\frac12\E\normm{X-\mu}{\nabla^2e(X)}^2.
    \end{align}
    Here \reason{a} applies $2a^\top Ac\le a^\top Aa+c^\top Ac$, which holds for every
    $A\succeq0$ because $(a-c)^\top A(a-c)\ge0$, with $a=x_0-\mu$, $c=X-\mu$ and
    $A=\nabla^2e(X)\succeq0$ by convexity of $e$; and \reason{b} uses $\bar H=\E\nabla^2e(X)$.
    Consequently, considering the term in \cref{sw:derivation-step1}, we have
    \begin{align}
        \lambda\E\bigl[(x_0-\mu)^\top\nabla^2e(X)(X-\mu)\bigr]-\frac\lambda2\normm{x_0-\mu}{\bar H}^2
        \le\frac\lambda2\E\normm{X-\mu}{\nabla^2e(X)}^2.
        \ineqlabel{sw:derivation-step1-continued}
    \end{align}

    Now we bound the right-hand side of \cref{sw:derivation-step1-continued}.
    First, we standardize the sample. Let $Z=\Sigma^{-1/2}(X-\mu)\sim\mathcal N(0,I_d)$
    and $B=\Sigma^{1/2}\nabla^2e(X)\Sigma^{1/2}$, which is positive semidefinite by
    convexity of $e$. Then
    \begin{align}
        \normm{X-\mu}{\nabla^2e(X)}^2
        &\explan{=}{a}Z^\top BZ
        \explan{\le}{b}\|Z\|^2\tr(B)
        \explan{=}{c}\|Z\|^2\tr\bigl(\nabla^2e(X)\Sigma\bigr).
    \end{align}
    Here \reason{a} substitutes $X-\mu=\Sigma^{1/2}Z$; \reason{b} uses
    $z^\top Bz\le\|z\|^2\lambda_{\max}(B)\le\|z\|^2\tr(B)$, which holds for every
    $B\succeq0$; and \reason{c} uses the cyclic property of the trace.
    Taking expectations,
    \begin{align}
        \frac\lambda2\E\normm{X-\mu}{\nabla^2e(X)}^2
        \le\frac\lambda2\E\bigl[\|Z\|^2\tr\bigl(\nabla^2e(X)\Sigma\bigr)\bigr].
        \ineqlabel{sw:derivation-step1-1}
    \end{align}
    Compared with $\tr(\bar H\Sigma)=\E\tr(\nabla^2e(X)\Sigma)$, the only difference is
    the factor $\|Z\|^2$, which is about $d$ for a typical sample.

    Next we handle the typical samples. Let $r^2=16(d+L)$ and split the
    expectation in \cref{sw:derivation-step1-1} according to
    whether $\|Z\|^2\le r^2$. The part on this event satisfies
    \begin{align}
        \frac\lambda2\E\bigl[\|Z\|^2\tr\bigl(\nabla^2e(X)\Sigma\bigr)\1\{\|Z\|^2\le r^2\}\bigr]
        &\explan{\le}{a}\frac{\lambda r^2}2\E\bigl[\tr\bigl(\nabla^2e(X)\Sigma\bigr)\1\{\|Z\|^2\le r^2\}\bigr]\nonumber\\
        &\explan{\le}{b}\frac{\lambda r^2}2\E\tr\bigl(\nabla^2e(X)\Sigma\bigr)\nonumber\\
        &\explan{=}{c}8\lambda(d+L)\tr(\bar H\Sigma)\nonumber\\
        &\explan{\le}{d}\tr(\bar H\Sigma).
        \ineqlabel{sw:derivation-step1-2}
    \end{align}
    Here \reason{a} uses $\|Z\|^2\le r^2$ on the event; \reason{b} drops the indicator, which is
    allowed because $\tr(\nabla^2e(X)\Sigma)\ge0$; \reason{c} uses $\bar H=\E\nabla^2e(X)$ and
    $r^2=16(d+L)$; and \reason{d} uses $\lambda\le1/(8(d+L))$ from
    \cref{sw:derivation-conditions}.

    It remains to handle the atypical samples, $\|Z\|^2>r^2$. There the curvature
    of $e$ is not controlled by $\tr(\bar H\Sigma)$, so we use instead that these samples
    are exponentially rare and that $e$ is Lipschitz.
    Let $h(z)=e(\mu+\Sigma^{1/2}z)$, let $L_e$ be the Lipschitz constant of $e$, and
    let $s=\|\Sigma\|_{\op}^{1/2}$. Then $h$ is convex and $L_es$-Lipschitz, and
    $\Delta h(Z)=\tr(\nabla^2e(X)\Sigma)$, where $\Delta h\ge0$ by convexity. We write $\Delta h(dz)=\Delta h(z)\,dz$ in the integrals below.
    Write $\gamma_{0,I_d}$ and $\gamma_{0,2I_d}$ for the Gaussian densities with
    covariances $I_d$ and $2I_d$, and let $Z_2\sim\N(0,2I_d)$. Then
    \begin{align}
        &\frac\lambda2\E\bigl[\|Z\|^2\tr\bigl(\nabla^2e(X)\Sigma\bigr)\1\{\|Z\|^2>r^2\}\bigr]\nonumber\\
        &\quad\explan{=}{a}\frac\lambda2\int_{\|z\|^2>r^2}\|z\|^2\gamma_{0,I_d}(z)\,\Delta h(dz)\nonumber\\
        &\quad\explan{\le}{b}\frac\lambda2\,2^{d/2}e^{-r^2/4}\int\|z\|^2\gamma_{0,2I_d}(z)\,\Delta h(dz)\nonumber\\
        &\quad\explan{=}{c}-\frac\lambda2\,2^{d/2}e^{-r^2/4}\int\ip{\nabla h(z)}{\bigl(2-\|z\|^2/2\bigr)z}\gamma_{0,2I_d}(z)\,dz\nonumber\\
        &\quad\explan{\le}{d}\frac\lambda2\,2^{d/2}e^{-r^2/4}L_es\,\E\bigl[\bigl(2+\|Z_2\|^2/2\bigr)\|Z_2\|\bigr]\nonumber\\
        &\quad\explan{\le}{e}\frac\lambda2L_es\cdot5(d+2)^{3/2}2^{d/2}e^{-4(d+L)}\nonumber\\
        &\quad\explan{\le}{f}\frac\lambda2L_ese^{-4L}.
        \ineqlabel{sw:derivation-step1-3}
    \end{align}
    Here \reason{a} writes the expectation as an integral against the measure $\Delta h$;
    \reason{b} uses $\gamma_{0,I_d}(z)=2^{d/2}e^{-\|z\|^2/4}\gamma_{0,2I_d}(z)$ and
    $\|z\|^2>r^2$, and then enlarges the domain of a nonnegative integral;
    \reason{c} integrates by parts, using
    $\nabla\bigl(\|z\|^2\gamma_{0,2I_d}(z)\bigr)=(2-\|z\|^2/2)z\,\gamma_{0,2I_d}(z)$;
    \reason{d} uses the Cauchy--Schwarz inequality, $\|\nabla h\|\le L_es$ and
    $|2-\|z\|^2/2|\le2+\|z\|^2/2$;
    \reason{e} uses $\E\|Z_2\|\le\sqrt{2d}$ and
    $\E\|Z_2\|^3\le(\E\|Z_2\|^4)^{3/4}=(4d(d+2))^{3/4}$, which give
    $2\E\|Z_2\|+\frac12\E\|Z_2\|^3\le3\sqrt2(d+2)^{3/2}\le5(d+2)^{3/2}$,
    together with $r^2=16(d+L)$; and \reason{f} uses
    $5(d+2)^{3/2}2^{d/2}e^{-4d}\le1$ for every $d\ge1$, since the left-hand side
    is decreasing in $d$ and is less than one at $d=1$.

    Adding \cref{sw:derivation-step1-2,sw:derivation-step1-3}
    and using \cref{sw:derivation-step1-1}, we obtain
    \begin{align}
        \frac\lambda2\E\normm{X-\mu}{\nabla^2e(X)}^2\le\tr(\bar H\Sigma)+\frac\lambda2L_ese^{-4L}.
        \ineqlabel{sw:derivation-step1-final}
    \end{align}

    \proofstepinline{2} We control the negative part of the weight.
    Next we address the last term in \cref{sw:derivation-quadratic}:
    \begin{align}
        -\E\bigl[w(X)D_e(x_\star\,\|\,X)\bigr].
        \label{sw:derivation-step2}
    \end{align}
    Without the weight this term would be nonpositive, since
    $D_e(x_\star\,\|\,X)\ge0$ by convexity of $e$. With the weight it is still
    nonpositive on the event $w(X)\ge0$, but it can be positive on the event
    $w(X)<0$. We drop the first part and keep only the second:
    \begin{align}
        -\E\bigl[w(X)D_e(x_\star\,\|\,X)\bigr]
        &\explan{\le}{a}-\E\bigl[w(X)D_e(x_\star\,\|\,X)\1\{w(X)<0\}\bigr]
        \nonumber\\
        &\explan{=}{b}\E\bigl[(-w(X))_+D_e(x_\star\,\|\,X)\bigr].
        \ineqlabel{sw:derivation-step2-sign}
    \end{align}
    Here \reason{a} drops $-\E[w(X)D_e(x_\star\,\|\,X)\1\{w(X)\ge0\}]$, which is
    nonpositive; and \reason{b} uses $-w\1\{w<0\}=(-w)_+$.

    Convexity gives no upper bound on the Bregman divergence, so we use the
    Lipschitz constant of $e$ instead:
    \begin{align}
        D_e(x_\star\,\|\,X)
        &\explan{=}{a}e(x_\star)-e(X)-\ip{\nabla e(X)}{x_\star-X}\nonumber\\
        &\explan{\le}{b}2L_e\|x_\star-X\|\nonumber\\
        &\explan{\le}{c}2L_e\bigl(\|x_\star-\mu\|+\|\Sigma^{1/2}Z\|\bigr)\nonumber\\
        &\explan{\le}{d}2L_e\bigl(\|x_\star-\mu\|+s\|Z\|\bigr).
        \ineqlabel{sw:derivation-step2-bregman}
    \end{align}
    Here \reason{a} is the definition of the Bregman divergence; \reason{b} uses
    $|e(x_\star)-e(X)|\le L_e\|x_\star-X\|$, the Cauchy--Schwarz inequality and
    $\|\nabla e(X)\|\le L_e$; \reason{c} uses the triangle inequality and
    $X-\mu=\Sigma^{1/2}Z$; and \reason{d} uses $\|\Sigma^{1/2}\|_{\op}=s$.

    Next we write the weight in terms of $Z$. Let
    $a=\lambda\normm{x_0-\mu}{\Sigma^{-1}}$. If $a=0$, then $w=1$ and the
    right-hand side of \cref{sw:derivation-step2-sign} is
    zero, so assume $a>0$. Let
    $u=\Sigma^{-1/2}(x_0-\mu)/\normm{x_0-\mu}{\Sigma^{-1}}$, which is a unit
    vector, and write $Z=Gu+Z_\perp$ with $G=\ip uZ$. Then
    $G\sim\mathcal N(0,1)$ is independent of $Z_\perp$,
    $\|Z\|\le|G|+\|Z_\perp\|$, and
    $\E\|Z_\perp\|\le(\E\|Z_\perp\|^2)^{1/2}=\sqrt{d-1}\le\sqrt d$. Moreover,
    \begin{align}
        w(X)=1+\lambda\ip{\Sigma^{-1/2}(x_0-\mu)}{Z}=1+aG,
    \end{align}
    so $w(X)<0$ exactly when $G<-1/a$, and $(-w(X))_+=a(-G-1/a)_+$.
    We need two moments of $G$ on this event. Let $\phi$ and $\Phi$ be the
    standard Gaussian density and distribution function, and let $b>0$. Since
    $t\phi(t)=-\phi'(t)$, integration by parts gives
    \begin{align}
        \E\bigl[(-G-b)_+|G|\bigr]=\int_b^\infty(t-b)\,t\phi(t)\,dt=\Phi(-b),
    \end{align}
    and since $|G|\ge b$ wherever $(-G-b)_+>0$, also
    $\E(-G-b)_+\le\Phi(-b)/b$. Therefore
    \begin{align}
        &\E\bigl[(-w(X))_+D_e(x_\star\,\|\,X)\bigr]\nonumber\\
        &\quad\explan{\le}{a}2L_e\E\bigl[(-w(X))_+\bigl(\|x_\star-\mu\|+s\|Z\|\bigr)\bigr]\nonumber\\
        &\quad\explan{\le}{b}2L_ea\E\bigl[(-G-1/a)_+\bigl(\|x_\star-\mu\|+s\|Z_\perp\|+s|G|\bigr)\bigr]\nonumber\\
        &\quad\explan{=}{c}2L_ea\Bigl[\bigl(\|x_\star-\mu\|+s\E\|Z_\perp\|\bigr)\E(-G-1/a)_+
        +s\E\bigl[(-G-1/a)_+|G|\bigr]\Bigr]\nonumber\\
        &\quad\explan{\le}{d}2L_ea\Bigl[\bigl(\|x_\star-\mu\|+s\sqrt d\bigr)a\Phi(-1/a)+s\Phi(-1/a)\Bigr]\nonumber\\
        &\quad\explan{\le}{e}L_ea\bigl(a\|x_\star-\mu\|+s(1+a\sqrt d)\bigr)e^{-1/(2a^2)}\nonumber\\
        &\quad\explan{\le}{f}L_e\bigl(\|x_\star-\mu\|+2s\sqrt d\bigr)e^{-4L}.
        \ineqlabel{sw:derivation-step2-tail}
    \end{align}
    Here \reason{a} uses \cref{sw:derivation-step2-bregman};
    \reason{b} uses $(-w(X))_+=a(-G-1/a)_+$ and $\|Z\|\le|G|+\|Z_\perp\|$; \reason{c} uses
    the independence of $G$ and $Z_\perp$; \reason{d} uses $\E\|Z_\perp\|\le\sqrt d$
    and the two moments of $G$ with $b=1/a$; \reason{e} collects terms and uses
    $\Phi(-b)\le\frac12e^{-b^2/2}$; and \reason{f} uses $a\le1/\sqrt{8L}$ from
    \cref{sw:derivation-conditions}, which gives
    $e^{-1/(2a^2)}\le e^{-4L}$ and $a\le1$, together with $1\le\sqrt d$.

    Combining \cref{sw:derivation-step2-sign,sw:derivation-step2-tail}, we obtain
    \begin{align}
        -\E\bigl[w(X)D_e(x_\star\,\|\,X)\bigr]\le L_e\bigl(\|x_\star-\mu\|+2s\sqrt d\bigr)e^{-4L}.
        \ineqlabel{sw:derivation-step2-final}
    \end{align}

    \proofstepinline{3} We combine the bounds.
    Finally we put the two steps together. Bounding the last three terms of
    \cref{sw:derivation-quadratic} by
    \cref{sw:derivation-step1-continued,sw:derivation-step1-final,sw:derivation-step2-final}
    gives
    \begin{align}
        F-e(x_\star)
        &\le \bar q(\mu)-\bigl(\bar q(x_0)+\ip{\nabla \bar q(x_0)}{x_\star-x_0}\bigr)+2\tr(\bar H\Sigma)\nonumber\\
        &\qquad+\frac\lambda2L_ese^{-4L}+L_e\bigl(\|x_\star-\mu\|+2s\sqrt d\bigr)e^{-4L}.
        \ineqlabel{sw:derivation-step3-combined}
    \end{align}
    It remains to show that the last two terms are at most $n^{-3}$. We use four
    facts. First, $\lambda\le1$ by \cref{sw:derivation-conditions}.
    Second, $s\le1$: the covariance assumption in the lemma gives
    $\Sigma\preceq4\sigma^2I_d$, so
    $ s\le2\sigma=2/(M\sqrt d\,L^2)\le2/L^2\le1$, using
    $M\ge d^{-1/2}$ and $L\ge100$.
    Third, $\|x_\star-\mu\|\le4d$, since $x_\star\in\K_\eps$, the mean $\mu$ lies
    in $\K_\eps$ by assumption, and $\K_\eps\subseteq\K\subseteq\mathbb B_{2d}$.
    Fourth, $L_e\le5/\eps\le5\sqrt n$ by \cref{lem:extension}, since
    $\eps=A^4d^2ML^5/\sqrt n\ge1/\sqrt n$. Therefore
    \begin{align}
        \frac\lambda2L_ese^{-4L}+L_e\bigl(\|x_\star-\mu\|+2s\sqrt d\bigr)e^{-4L}
        &\explan{\le}{a}\Bigl(\frac12+4d+2\sqrt d\Bigr)L_ee^{-4L}\nonumber\\
        &\explan{\le}{b}7dL_ee^{-4L}\nonumber\\
        &\explan{\le}{c}35d\sqrt n\,e^{-4L}\nonumber\\
        &\explan{=}{d}35d\sqrt n\Bigl(\frac\delta{100nd}\Bigr)^{400}\nonumber\\
        &\explan{\le}{e}n^{-3}.
        \ineqlabel{sw:derivation-step3-error}
    \end{align}
    Here \reason{a} uses $\lambda\le1$, $s\le1$ and $\|x_\star-\mu\|\le4d$; \reason{b} uses
    $d\ge1$; \reason{c} uses $L_e\le5\sqrt n$; \reason{d} substitutes
    $L=100\log(100nd/\delta)$; and \reason{e} uses $\delta<1$ and $n,d\ge1$.
    Substituting \cref{sw:derivation-step3-error} into
    \cref{sw:derivation-step3-combined}, we obtain
    \begin{align}
        F-e(x_\star)
        \le \bar q(\mu)-\bigl(\bar q(x_0)+\ip{\nabla \bar q(x_0)}{x_\star-x_0}\bigr)+2\tr(\bar H\Sigma)+n^{-3}.
        \ineqlabel{sw:derivation-step3-final}
    \end{align}

    By \cref{sw:derivation-tangent}, this is precisely the stated inequality
    when $e$ is smooth.

    To remove the smoothness assumption, let
    $e_\tau(x)=\E_Z[e(x+\tau Z)]$ with $Z\sim\N(0,I_d)$ and $\tau>0$.
    Then $e_\tau$ is smooth and convex, has Lipschitz constant at most $L_e$,
    and satisfies
    \begin{align}
        \sup_{x\in\R^d}|e_\tau(x)-e(x)|\leq\tau L_e\E\|Z\|
        \longrightarrow0.
    \end{align}
    Apply the preceding calculation to $e_\tau$, defining $F_\tau$,
    $\bar g_\tau$, $\bar H_\tau$ and $\bar q_\tau$ by the same Gaussian
    averages and quadratic formula. All bounds use only the common
    Lipschitz bound, the covariance assumption and the displayed parameter
    conditions, so they hold uniformly in $\tau$. The Gaussian score
    identities in \cref{eq:gaussian-derivatives}, together with uniform
    convergence and integrability of the Gaussian scores, imply
    $F_\tau\to F$, $\bar g_\tau\to\bar g$ and
    $\bar H_\tau\to\bar H$. Hence $\bar q_\tau$ and its gradient converge
    at each fixed point. Passing to the limit gives
    \cref{sw:derivation-goal} for $e$, and
    \cref{sw:derivation-tangent} proves the lemma.
    \endgroup
\end{proof}

\section{Trace bound}
\label{app:trace}

The trace term in \cref{sw:tangent} measures the curvature of the
smoothed loss in the directions in which the sampling distribution is
wide. Indeed, the Jensen gap of the population quadratic is
\begin{align}
    \E_s[\bar q_s(X_s)]-\bar q_s(\mu_s)
    &=\frac\lambda2\E_s\bigl[(X_s-\mu_s)^\top\bar H_s(X_s-\mu_s)\bigr]
    =\frac\lambda2\tr(\bar H_s\Sigma_s).
\end{align}
The covariance update responds to this curvature. We will show that the
sum of the traces is controlled by a log-determinant ratio, and then
bound this ratio by controlling the growth of the inverse covariance.

Throughout this and following appendices we will use the parameters of \cref{thm:main} in  expanded form:
\begin{align}
    \sigma^2&=\frac{1}{dM^2L^4},&
    \eta&=\frac{dM}{AL\sqrt n},&
    \rho&=A^3d^3M^2L^4,\notag\\
    \lambda&=\frac{1}{A^2d^{3/2}ML^3},&
    \eps&=\frac{A^4d^2ML^5}{\sqrt n}.&&
    \eqlabel{eq:parameters}
\end{align}

The sampled
Hessians $H_s$ can be indefinite, so we work instead with
\begin{align}
    \bar\Sigma_1=\Sigma_1=\sigma^2I_d,\qquad
    \bar\Sigma_{s+1}^{-1}
    =\bar\Sigma_s^{-1}+\frac{\lambda\eta}{2}\bar H_s,
    \eqlabel{fb:inverse-covariance}
\end{align}
and compare these matrices with $\Sigma_s$. Since $\bar H_s\succeq0$, the
matrices $\bar\Sigma_s^{-1}$ are nondecreasing and $\bar\Sigma_s\preceq\sigma^2I_d$.

\begin{lemma}[Growth of the inverse covariance]
    \label{fb:spectrum}
    Let $0\leq t\leq n$, and suppose that
    \begin{align}
        \tfrac12\bar\Sigma_s^{-1}\preceq\Sigma_s^{-1}
        \preceq2\bar\Sigma_s^{-1}\qquad(1\leq s\leq t).
        \ineqlabel{fb:comparison}
    \end{align}
    Then
    \begin{align}
        \|\bar\Sigma_{t+1}^{-1}\|_{\op}^{1/2}
        \leq\sigma^{-1}+\frac{2t\lambda\eta}{\eps}\leq n.
        \ineqlabel{fb:spectrum-bound}
    \end{align}
    Consequently,
    $\Sigma_1^{-1}\preceq\bar\Sigma_{t+1}^{-1}\preceq n^2I_d$.
\end{lemma}

\begin{proof}[Proof of \cref{fb:spectrum}]
    The lower bound follows from $\bar H_s\succeq0$. Also, inverting the
    first inequality in \cref{fb:comparison} gives
    $\Sigma_s\preceq2\bar\Sigma_s\preceq2\sigma^2I_d$.
    It remains to control the largest eigenvalue of
    $\bar\Sigma_{t+1}^{-1}$. The update gives
    \begin{align}
        \|\bar\Sigma_{s+1}^{-1}\|_{\op}
        &\leq\|\bar\Sigma_s^{-1}\|_{\op}
        +\frac{\lambda\eta}{2}\|\bar H_s\|_{\op}.
        \ineqlabel{fb:spectrum-update}
    \end{align}
    We therefore need an upper bound on $\|\bar H_s\|_{\op}$.

    Fix $u\in\R^d$, write $\partial_u$ for differentiation in
    direction $u$, and let $W_s=X_s-\mu_s$. Gaussian integration by parts gives
    \begin{align}
        u^\top\bar H_su
        &\explan{=}{a}\int_{\R^d}e_s(x)\,
        \partial_u^2\gamma_{\mu_s,\Sigma_s}(x)\,dx\nonumber\\
        &\explan{=}{b}-\int_{\R^d}\partial_ue_s(x)\,
        \partial_u\gamma_{\mu_s,\Sigma_s}(x)\,dx\nonumber\\
        &\explan{=}{c}\E_s\!\left[
        \ip{\nabla e_s(X_s)}u\ip{\Sigma_s^{-1}W_s}u\right]\nonumber\\
        &\explan{\leq}{d}\frac5\eps\|u\|\normm{u}{\Sigma_s^{-1}}.
        \ineqlabel{fb:direction}
    \end{align}
    Here \reason{a} uses the Gaussian derivative formula;
    \reason{b} integrates by parts; and \reason{c} uses
    $\partial_u\gamma_{\mu_s,\Sigma_s}(x)
    =-\ip{\Sigma_s^{-1}(x-\mu_s)}u\gamma_{\mu_s,\Sigma_s}(x)$.
    For \reason{d}, Cauchy--Schwarz and \cref{lem:extension} give
    \begin{align}
        \E_s\!\left[
        \ip{\nabla e_s(X_s)}u\ip{\Sigma_s^{-1}W_s}u\right]
        \leq\frac5\eps\|u\|
        \left(\E_s\ip{\Sigma_s^{-1}W_s}u^2\right)^{1/2}
        =\frac5\eps\|u\|\normm{u}{\Sigma_s^{-1}}.
    \end{align}
    These identities also hold when $e_s$ is nonsmooth: a Lipschitz
    function has bounded weak first derivatives, and the Gaussian
    density and its derivatives decay fast enough to justify integration
    by parts. No pointwise second derivative of $e_s$ is needed.

    Since $\bar H_s\succeq0$, taking the supremum over unit vectors in
    \cref{fb:direction} yields
    \begin{align}
        \|\bar H_s\|_{\op}
        =\sup_{\|u\|=1}u^\top\bar H_su
        \leq\frac5\eps\|\Sigma_s^{-1}\|_{\op}^{1/2}.
        \ineqlabel{fb:hessian-norm}
    \end{align}
    Substitute this into \cref{fb:spectrum-update}, and use the upper
    bound in \cref{fb:comparison}:
    \begin{align}
        \|\bar\Sigma_{s+1}^{-1}\|_{\op}
        &\leq\|\bar\Sigma_s^{-1}\|_{\op}
        +\frac{5\lambda\eta}{2\eps}\|\Sigma_s^{-1}\|_{\op}^{1/2}\nonumber\\
        &\leq\|\bar\Sigma_s^{-1}\|_{\op}
        +\frac{5\sqrt2\lambda\eta}{2\eps}
        \|\bar\Sigma_s^{-1}\|_{\op}^{1/2}\nonumber\\
        &\leq\left(\|\bar\Sigma_s^{-1}\|_{\op}^{1/2}
        +\frac{2\lambda\eta}{\eps}\right)^2.
    \end{align}
    The last inequality follows by expanding the square and using
    $5\sqrt2/2\leq4$. Taking square roots and iterating gives
    \begin{align}
        \|\bar\Sigma_{t+1}^{-1}\|_{\op}^{1/2}
        \leq\|\bar\Sigma_1^{-1}\|_{\op}^{1/2}
        +\frac{2t\lambda\eta}{\eps}
        =\sigma^{-1}+\frac{2t\lambda\eta}{\eps}.
    \end{align}
    Finally, $\eps\leq1/2$ and \cref{eq:parameters} imply
    $n\geq4A^8d^4M^2L^{10}$. Together with $M\geq d^{-1/2}$,
    this gives
    \begin{align}
        \sigma^{-1}=M\sqrt d\,L^2\leq\frac n4,
        \qquad
        \frac{2\lambda\eta}{\eps}
        =\frac{2}{A^7d^{5/2}ML^9}\leq\frac14.
    \end{align}
    Thus $\sigma^{-1}+2t\lambda\eta/\eps\leq n/4+t/4\leq n$,
    proving \cref{fb:spectrum-bound}.
\end{proof}

\begin{lemma}[Trace sum]
    \label{fb:trace}
    Let $0\leq t\leq n$, and suppose that \cref{fb:comparison} holds. Then
    \begin{align}
        \sum_{s=1}^t\tr(\bar H_s\Sigma_s)
        \leq\frac{8}{\lambda\eta}
        \log\frac{\det\Sigma_1}{\det\bar\Sigma_{t+1}}
        \leq\frac{16dL}{\lambda\eta}.
        \ineqlabel{fb:trace-sum}
    \end{align}
\end{lemma}

\begin{proof}[Proof of \cref{fb:trace}]
    The case $t=0$ is immediate. For $s\leq t$, we have
    $\Sigma_s\preceq2\bar\Sigma_s$ by \cref{fb:comparison}.
    Since $\bar H_s\succeq0$, this implies
    \begin{align}
        \tr(\bar H_s\Sigma_s)
        \leq2\tr(\bar H_s\bar\Sigma_s).
        \ineqlabel{fb:trace-comparison}
    \end{align}
    We now normalize the Hessian by $\bar\Sigma_s$. Let
    \begin{align}
        B_s=\frac{\lambda\eta}{2}
        \bar\Sigma_s^{1/2}\bar H_s\bar\Sigma_s^{1/2}.
    \end{align}
    By cyclicity of the trace,
    \begin{align}
        \sum_{s=1}^t\tr(\bar H_s\Sigma_s)
        \leq2\sum_{s=1}^t\tr(\bar H_s\bar\Sigma_s)
        =\frac4{\lambda\eta}\sum_{s=1}^t\tr(B_s).
        \ineqlabel{fb:normalized-trace}
    \end{align}
    To turn the last sum into log determinants, we first show that the
    eigenvalues of $B_s$ lie in $[0,1]$.

    For every unit vector $u$, apply \cref{fb:direction} with
    $\bar\Sigma_s^{1/2}u$ in place of $u$:
    \begin{align}
        u^\top\bar\Sigma_s^{1/2}\bar H_s\bar\Sigma_s^{1/2}u
        &\leq\frac5\eps\|\bar\Sigma_s^{1/2}u\|
        \sqrt{u^\top\bar\Sigma_s^{1/2}
        \Sigma_s^{-1}\bar\Sigma_s^{1/2}u}
        \leq\frac{5\sqrt2}{\eps}\|\bar\Sigma_s^{1/2}u\|
        \leq\frac{5\sqrt2\sigma}{\eps}.
    \end{align}
    The second inequality uses $\Sigma_s^{-1}\preceq2\bar\Sigma_s^{-1}$,
    and the third uses $\bar\Sigma_s\preceq\sigma^2I_d$. Consequently,
    \begin{align}
        0\preceq B_s
        \preceq\frac{5\sqrt2\lambda\eta\sigma}{2\eps}I_d
        =\frac{5\sqrt2}{2A^7d^3M^2L^{11}}I_d
        \preceq I_d,
    \end{align}
    where the last two steps use \cref{eq:parameters},
    $M\geq d^{-1/2}$, $L\geq100$ and sufficiently large $A$.

    Apply $z\leq2\log(1+z)$ for $z\in[0,1]$ to each eigenvalue of
    $B_s$. Returning to \cref{fb:normalized-trace}, we get
    \begin{align}
        \sum_{s=1}^t\tr(\bar H_s\Sigma_s)
        &\leq\frac4{\lambda\eta}\sum_{s=1}^t\tr(B_s)\nonumber\\
        &\leq\frac8{\lambda\eta}\sum_{s=1}^t\log\det(I_d+B_s)\nonumber\\
        &\explan{=}{a}\frac8{\lambda\eta}
        \sum_{s=1}^t\log
        \frac{\det(\bar\Sigma_{s+1}^{-1})}
        {\det(\bar\Sigma_s^{-1})}\nonumber\\
        &\explan{=}{b}\frac8{\lambda\eta}
        \log\frac{\det(\bar\Sigma_{t+1}^{-1})}
        {\det(\bar\Sigma_1^{-1})}\nonumber\\
        &=\frac8{\lambda\eta}
        \log\frac{\det\Sigma_1}{\det\bar\Sigma_{t+1}}.
        \ineqlabel{fb:trace-logdet}
    \end{align}
    For \reason{a}, the update in \cref{fb:inverse-covariance} gives
    \begin{align}
        I_d+B_s
        =\bar\Sigma_s^{1/2}\bar\Sigma_{s+1}^{-1}\bar\Sigma_s^{1/2},
    \end{align}
    so its determinant is the displayed ratio. In \reason{b}, the
    logarithms telescope.

    It remains to bound the final determinant ratio. By
    \cref{fb:spectrum}, $\|\bar\Sigma_{t+1}^{-1}\|_{\op}\leq n^2$.
    Since $\Sigma_1=\sigma^2I_d$ and $\sigma\leq1$,
    \begin{align}
        \frac{\det\Sigma_1}{\det\bar\Sigma_{t+1}}
        &=\sigma^{2d}\det(\bar\Sigma_{t+1}^{-1})
        \leq\sigma^{2d}\|\bar\Sigma_{t+1}^{-1}\|_{\op}^d
        \leq\sigma^{2d}n^{2d}\leq n^{2d}.
    \end{align}
    Substituting this into \cref{fb:trace-logdet} gives
    \begin{align}
        \sum_{s=1}^t\tr(\bar H_s\Sigma_s)
        \leq\frac{16d\log n}{\lambda\eta}
        \leq\frac{16dL}{\lambda\eta},
    \end{align}
    which proves \cref{fb:trace-sum}.
\end{proof}

\section{Feedback bounds}
\label{fb:section}

The concentration argument needs a bound on the sum of the conditional
second moments of the observed responses. The responses $Y_s$ are not
bounded: both the Gaussian sample and the observation noise can be
large. We will nevertheless prove that
$\sum_{s=1}^t\E_sY_s^2\leq5n$ on every epoch prefix satisfying the
covariance comparison from \cref{app:trace}.

Throughout this appendix, the parameters are those of \cref{thm:main},
with $\eps\leq1/2$. All indices belong to one epoch, and $n$ is the
original horizon. Write $W_s=X_s-\mu_s$ and $m_s=\E_sY_s^2$.
The key quantity is the extension penalty $v(X_s)$: it determines how
large $Y_s$ can be outside $\K_\eps$. We first control the contribution
of the event $v(X_s)>1$. We then relate the mean of $v(X_s)$ to $\tr(\bar H_s\Sigma_s)$
and apply the trace bound.

\begin{lemma}[Gaussian gauge tail]
    \label{fb:gauge}
    Suppose $\mu_s\in\K_\eps$ and
    $0\prec\Sigma_s\preceq2\sigma^2I_d$. Then, for every $u\geq0$,
    \begin{align}
        \Pp_s\!\left(\pi(W_s)>2/L^2+u\right)
        &\leq e^{-u^2/(8\sigma^2)},\nonumber\\
        \E_s\!\left[v(X_s)^2\1_{\{v(X_s)>1\}}\right]
        &\leq2e^{-L^4/128}.
        \ineqlabel{fb:gaugetail}
    \end{align}
\end{lemma}

\begin{proof}[Proof of \cref{fb:gauge}]
    \proofstepinline{1}
    First, we bound the mean of the gauge.
    Conditional on the past, let $W'_s$ be independent of $W_s$, centered
    Gaussian, and have covariance $4\sigma^2I_d-\Sigma_s$.
    This covariance is positive semidefinite by assumption. If
    $Z\sim\N(0,I_d)$ and $U$ is uniform on the Euclidean unit sphere,
    then
    \begin{align}
        \E_s\pi(W_s)
        &\explan{\leq}{a}\E_s\pi(W_s+W'_s)
        =2\sigma\E\pi(Z)
        \explan{=}{b}2\sigma\E\|Z\|\E\pi(U)
        \explan{\leq}{c}2\sigma\sqrt d\,M
        =\frac2{L^2}.
        \ineqlabel{fb:gauge-mean}
    \end{align}
    For \reason{a}, apply conditional Jensen to the convex gauge,
    using $\E_s[W_s+W'_s\mid W_s]=W_s$. The next equality uses
    $W_s+W'_s\sim\N(0,4\sigma^2I_d)$ and positive homogeneity of
    $\pi$. In \reason{b}, the radius $\|Z\|$ and direction
    $Z/\|Z\|$ are independent, and the latter is uniform on the sphere.
    For \reason{c}, use $\E\|Z\|\leq\sqrt d$ and
    $\E\pi(U)\leq M$. The last equality substitutes $\sigma$.

    \proofstepinline{2} We show that the gauge concentrates around its mean.
    The map $z\mapsto\pi(\Sigma_s^{1/2}z)$ is $2\sigma$-Lipschitz.
    Indeed, $\pi$ is $1$-Lipschitz because $\mathbb B_1\subseteq\K$,
    and
    \begin{align}
        |\pi(\Sigma_s^{1/2}z)-\pi(\Sigma_s^{1/2}z')|
        \leq\|\Sigma_s^{1/2}(z-z')\|
        \leq\|\Sigma_s\|_{\op}^{1/2}\|z-z'\|
        \leq2\sigma\|z-z'\|.
    \end{align}
    Gaussian concentration for a Lipschitz function
    \citep[Theorem~5.6]{boucheron2013concentration} therefore gives
    \begin{align}
        \Pp_s\!\left(\pi(W_s)>2/L^2+u\right)
        &\leq\Pp_s\!\left(\pi(W_s)>\E_s\pi(W_s)+u\right)
        \leq e^{-u^2/(8\sigma^2)}.
    \end{align}
    The first inequality uses \cref{fb:gauge-mean}. This proves the
    first assertion of the lemma.

    We also use this to prove second claim in the following form:
    for $r\geq1/2$, the parameter bounds $L\geq100$ and
    $\sigma^2\leq L^{-4}$ imply
    \begin{align}
        \Pp_s\!\left(\pi(W_s)>r\right)
        &\leq\exp\!\left(-\frac{(r-2/L^2)^2}{8\sigma^2}\right)
        \leq e^{-r^2L^4/32},
        \ineqlabel{fb:gauge-large-tail}
    \end{align}
    where the last inequality uses $r-2/L^2\geq r/2$.

    \proofstepinline{3} We bound the contribution from large values of the penalty.
    The mean lies in $\K_\eps$, so $\pi(\mu_s)\leq1-\eps$.
    Subadditivity of the gauge and the definition of $v$ give
    \begin{align}
        v(X_s)
        =\max\!\left\{\frac{\pi(\mu_s+W_s)}{1-\eps}-1,0\right\}
        \leq\max\!\left\{\frac{\pi(\mu_s)+\pi(W_s)}{1-\eps}-1,0\right\}
        \leq\frac{\pi(W_s)}{1-\eps}\leq2\pi(W_s).
        \ineqlabel{fb:penalty-gauge}
    \end{align}
    In particular, $v(X_s)>1$ implies $\pi(W_s)>1/2$. Hence
    \begin{align}
        \E_s\!\left[v(X_s)^2\1_{\{v(X_s)>1\}}\right]
        &\leq4\E_s\!\left[\pi(W_s)^2\1_{\{\pi(W_s)>1/2\}}\right]\nonumber\\
        &\explan{=}{a}\Pp_s\!\left(\pi(W_s)>1/2\right)
        +8\int_{1/2}^{\infty}r\Pp_s\!\left(\pi(W_s)>r\right)\,dr\nonumber\\
        &\explan{\leq}{b}e^{-L^4/128}
        +8\int_{1/2}^{\infty}r e^{-r^2L^4/32}\,dr\nonumber\\
        &=\left(1+\frac{128}{L^4}\right)e^{-L^4/128}
        \leq2e^{-L^4/128}.
    \end{align}
    Here \reason{a} is the tail-integral formula for the square of a
    nonnegative random variable above $1/2$, and \reason{b} applies
    \cref{fb:gauge-large-tail}. The final inequality uses $L\geq100$.
\end{proof}

\begin{lemma}[Conditional second moments]
    \label{fb:budget}
    Let $0\leq t\leq n$, and suppose that \cref{fb:comparison} holds. Then
    \begin{align}
        \sum_{s=1}^t\E_sY_s^2\leq5n.
        \ineqlabel{fb:second-moment-sum}
    \end{align}
\end{lemma}

\begin{proof}[Proof of \cref{fb:budget}]
    The case $t=0$ is immediate. The covariance comparison implies
    $\Sigma_s\preceq2\sigma^2I_d$ for all $s\leq t$ by
    \cref{fb:spectrum}, so \cref{fb:gauge} applies in every round.

    \proofstepinline{1} We reduce the response second moment to a bound on the penalty $v(X_s)$.
    Since $X_s$ is determined by the learner's randomness in round $s$,
    conditioning \cref{eq:noise} on $X_s$ gives
    $\E_s[\xi_s\mid X_s]=0$ and
    $\E_s[e^{\xi_s^2}\mid X_s]\leq2$. In particular,
    \begin{align}
        \E_s[\xi_s^2\mid X_s]
        \leq\log\E_s[e^{\xi_s^2}\mid X_s]
        \leq\log2\leq1,
    \end{align}
    where the first inequality is conditional Jensen.
    The extension formula and $0\leq f_s\leq1$ give
    \begin{align}
        0\leq e_s(X_s)
        &\leq1+v(X_s)+\frac2\eps v(X_s)
        \leq1+\frac3\eps v(X_s).
    \end{align}
    Now use $Y_s=e_s(X_s)+(1+v(X_s))\xi_s$. Expanding its square
    conditionally on $X_s$ removes the cross term, and yields
    \begin{align}
        m_s = \E_sY_s^2
        &=\E_s\!\left[e_s(X_s)^2
        +(1+v(X_s))^2\E_s[\xi_s^2\mid X_s]\right]\nonumber\\
        &\leq\E_s\!\left[(1+3v(X_s)/\eps)^2+(1+v(X_s))^2\right]\nonumber\\
        &\leq4+\left(\frac{18}{\eps^2}+2\right)\E_s v(X_s)^2\nonumber\\
        &\leq4+\frac{20}{\eps^2}\E_s v(X_s)^2.
        \ineqlabel{fb:response-penalty}
    \end{align}
    The penultimate inequality applies $(a+b)^2\leq2a^2+2b^2$ to
    each square; the last uses $\eps\leq1$.

    On $0\leq v(X_s)\leq1$, the square is at most $v(X_s)$.
    The remaining part is controlled by \cref{fb:gauge}, so
    \begin{align}
        \E_s v(X_s)^2
        &\leq\E_s\!\left[v(X_s)\1_{\{v(X_s)\leq1\}}\right]
        +\E_s\!\left[v(X_s)^2\1_{\{v(X_s)>1\}}\right]\nonumber\\
        &\leq\E_s v(X_s)+2e^{-L^4/128}.
        \ineqlabel{fb:penalty-second-moment}
    \end{align}
    Thus it remains to bound the mean penalty.

    \proofstepinline{2} We control the mean penalty by curvature.
    Let
    \begin{align}
        V_s=\langle D^2v,\gamma_{\mu_s,\Sigma_s}\rangle.
    \end{align}
    This is the distributional Hessian of $v$ integrated against the
    Gaussian density. The function $v$ is convex and Lipschitz, so
    $D^2v$ is a positive semidefinite matrix-valued measure so this
    pairing is finite. By \cref{lem:extension},
    $e_s-v/\eps$ is also convex. Its Hessian measure is positive
    semidefinite, and integrating it against the same density gives
    \begin{align}
        0\preceq V_s\preceq\eps\bar H_s.
        \ineqlabel{fb:penalty-curvature}
    \end{align}
    The supporting-plane inequality at $X_s$, with comparison point
    $\mu_s$, gives
    $v(X_s)\leq v(\mu_s)+\ip{\nabla v(X_s)}{X_s-\mu_s}$.
    The gradient exists almost surely because $v$ is Lipschitz and
    $X_s$ has a density. Since $v(\mu_s)=0$,
    \begin{align}
        \E_s v(X_s)
        \leq\E_s\ip{\nabla v(X_s)}{X_s-\mu_s}
        \explan{=}{a}\tr(V_s\Sigma_s)
        \explan{\leq}{b}\eps\tr(\bar H_s\Sigma_s).
        \ineqlabel{fb:penalty-mean}
    \end{align}
    Here \reason{a} is Gaussian integration by parts, interpreted
    through the weak derivatives of $v$, and \reason{b} uses
    \cref{fb:penalty-curvature} and $\Sigma_s\succeq0$.
    Combining \cref{fb:response-penalty,fb:penalty-second-moment,fb:penalty-mean}
    gives the roundwise bound
    \begin{align}
        m_s
        \leq4+\frac{20}{\eps}\tr(\bar H_s\Sigma_s)
        +\frac{40}{\eps^2}e^{-L^4/128}.
        \ineqlabel{fb:response-second-moment}
    \end{align}

    \proofstepinline{3} We sum over the epoch prefix.
    We first check that the last term is negligible.
    The parameters give $\eps\geq n^{-1/2}$ because
    $A^4d^2ML^5\geq1$. Also $\log n\leq L/100$ and
    $L\geq100\log200$, so
    \begin{align}
        \eps^{-2}e^{-L^4/128}
        &\leq n e^{-L^4/128}
        \leq\exp\!\left(\frac L{100}-\frac{L^4}{128}\right)
        \leq e^{-L}\leq\frac1{160}.
        \ineqlabel{fb:tail-remainder}
    \end{align}
    The third inequality follows from $L^4/128\geq101L/100$.
    Summing \cref{fb:response-second-moment}, using $t\leq n$, and
    applying \cref{fb:trace-sum}, we obtain
    \begin{align}
        \sum_{s=1}^t m_s
        &\leq4n+\frac{20}{\eps}\sum_{s=1}^t\tr(\bar H_s\Sigma_s)
        +\frac{40n}{\eps^2}e^{-L^4/128}\nonumber\\
        &\leq4n+\frac{320dL}{\eps\lambda\eta}
        +\frac{40n}{\eps^2}e^{-L^4/128}\nonumber\\
        &\explan{=}{a}4n+\frac{320n}{A\sqrt d\,M}
        +\frac{40n}{\eps^2}e^{-L^4/128}\nonumber\\
        &\explan{\leq}{b}4n+\frac n4+\frac n4
        \leq5n.
    \end{align}
    In \reason{a}, substitute
    $\lambda\eta\eps=A d^{3/2}ML/n$ from \cref{eq:parameters}.
    For \reason{b}, use $\sqrt d\,M\geq1$, $A\geq1280$ and
    \cref{fb:tail-remainder}.

    Every step used only the covariance comparison before each sample
    in the prefix. The result therefore includes round $t$ even if
    its sampled covariance update is the first to violate that
    comparison. This is the form needed in the stopped concentration
    argument of \cref{conc:section}.
\end{proof}

\section{Concentration of the derivative estimates}
\label{conc:section}

The preceding appendices bound conditional expectations along the
sampled trajectory. We now compare these quantities with the estimates
used by the algorithm. There are three tasks: control the observed
feedback and gradient energy, show that the sampled covariance remains
close to the covariance formed from conditional mean Hessians, and
control the quadratic errors throughout the focus set. The last task
needs uniformity because both the focus set and the points used by the
restart rule depend on the observations.

Within an epoch, write
\begin{align}
    Q_t(x)&=\sum_{s=1}^t q_s(x),&
    \bar Q_t(x)&=\sum_{s=1}^t\bar q_s(x),\notag\\
    \Gamma_t(x)&=\frac{\lambda}{4}
    \sum_{s=1}^t\|x-\mu_s\|_{H_s}^2,&
    \bar\Gamma_t(x)&=\frac{\lambda}{4}
    \sum_{s=1}^t\|x-\mu_s\|_{\bar H_s}^2.
    \eqlabel{conc:quadratic-definitions}
\end{align}
Here $q_s$ is the sampled quadratic and
$\bar q_s(x)=\E_s[q_s(x)]$ for each fixed $x$. Thus the barred
quantities use conditional means along the same trajectory as the
algorithm. Since $H_s$ need not be positive semidefinite,
$\|z\|_{H_s}^2$ denotes the quadratic form $z^\top H_sz$.

\begin{lemma}[Uniform estimation]
    \label{conc:uniform}
    Suppose the parameters are as in \cref{eq:parameters}, with
    $\eps\leq1/2$ and $A$ sufficiently large. There is an absolute
    constant $C$ such that, with probability at least $1-\delta/4$, the
    following hold at every round $t$.
    \begin{enumerate}
        \item The sampled inverse covariance is close to the one formed
            from conditional mean Hessians:
            \begin{align}
                \tfrac34\bar\Sigma_{t+1}^{-1}
                \preceq\Sigma_{t+1}^{-1}
                \preceq\tfrac54\bar\Sigma_{t+1}^{-1}.
                \ineqlabel{conc:covariance}
            \end{align}
        \item The observed responses and the gradient estimates satisfy
            \begin{align}
                \sum_{s=1}^tY_s^2\leq Cn,\qquad
                \sum_{s=1}^t\|g_s\|_{\Sigma_{s+1}}^2\leq CdnL.
                \ineqlabel{conc:energy}
            \end{align}
        \item The sampled quadratics are close to their conditional means
            on the focus set: with $\Delta=CL^2\sqrt{dn}/\lambda$,
            \begin{align}
                \sup_{x\in K_t}|Q_t(x)-\bar Q_t(x)|\leq\Delta,\qquad
                \sup_{x\in K_t}|\Gamma_t(x)-\bar\Gamma_t(x)|\leq\Delta.
                \ineqlabel{conc:quadratics}
            \end{align}
    \end{enumerate}
\end{lemma}

Throughout the proof, $C$ denotes an absolute constant that may increase
from one display to the next; it is independent of $A$. We choose $A$
after these constants. We will use the following consequences of
\cref{eq:parameters}:
\begin{align}
    &\eps^{-2},d,L\leq n,\qquad \lambda\leq1,\qquad
    \sigma\leq L^{-2},\qquad R\geq1,\notag\\
    &n\eps^2\geq L^3,\qquad
    \eps\geq\frac{d^{3/2}L}{\sqrt n},\qquad
    \eta\lambda=\frac{1}{A^3\sqrt{dn}L^4}.
    \ineqlabel{conc:parameters}
\end{align}

The following two-sided form of Freedman's inequality follows from
\citet[Theorem~3.10]{bercu2015concentration} applied to a martingale and its
negative.
\begin{lemma}[Freedman's inequality]
    \label{lem:freedman}
    Let $(D_s)_{s=1}^t$ be a square-integrable martingale-difference
    sequence such that, for some $b>0$, $|D_s|\leq b$ almost surely for
    every $s$. Then, for every $r,z>0$,
    \begin{align}
        \Pp\left(
            \left|\sum_{s=1}^tD_s\right|\geq rz+\frac{bz^2}{3},
            \quad
            \sum_{s=1}^t\E_s[D_s^2]\leq r^2
        \right)
        \leq2\exp\left(-\frac{z^2}{2}\right).
    \end{align}
\end{lemma}

\begin{proof}[Proof of \cref{conc:uniform}]
    \proofstepinline{1} First, we fix a possible starting round of an epoch.
    The epochs are random: the round at which an epoch begins and the
    number of rounds it lasts both depend on the restart test.
    We fix $s_{\mathrm{epoch}}\in\{1,\dots,n\}$ and set
    $n_{\mathrm{epoch}}=n-s_{\mathrm{epoch}}+1$, the deterministic number of
    calendar rounds remaining. If an epoch starts at $s_{\mathrm{epoch}}$,
    let $\tau$ be its number of rounds, including the round that triggers
    a restart or the fallback; otherwise let $\tau=0$. We do not condition
    on $\tau$. Local round $s$ corresponds to calendar round
    $s_{\mathrm{epoch}}+s-1$, so in this proof $\E_s$ and $\Pp_s$
    condition on $\F_{s_{\mathrm{epoch}}+s-2}$.
    The variables in the concentration sums are extended by zero for
    $s>\tau$, before taking conditional expectations. Expressions involving
    the samples are evaluated only during the epoch, and their retained
    terms are zero otherwise. The proposed covariance update after round
    $\tau$ is still included. At the very end, we will take a union bound
    over the at most $n$ possible starting rounds to get a global bound.

    \proofstepinline{2} We keep only the rounds at which the epoch is still
    active and the covariance comparison has held so far.
    \Cref{fb:spectrum,fb:budget} assume the comparison \cref{fb:comparison},
    which is part of what we are proving. For
    $s=1,\dots,n_{\mathrm{epoch}}$, let $\chi_s=1$ if $s\leq\tau$ and
    \begin{align}
        \tfrac12\bar\Sigma_j^{-1}\preceq\Sigma_j^{-1}
        \preceq2\bar\Sigma_j^{-1}\qquad\text{for every }j\leq s,
    \end{align}
    and $\chi_s=0$ otherwise. Whether the epoch has started and is still
    active is known before the sample in round $s$, as is $\Sigma_s$, which
    is computed at the end of round $s-1$. Thus $\chi_s$ is determined
    before the sample in round $s$. In particular, it retains the round
    that triggers a restart or whose proposed update first violates the
    covariance comparison.
    If an epoch starts at $s_{\mathrm{epoch}}$, then $\chi_1=1$ because
    $\Sigma_1=\bar\Sigma_1$. If $\chi_s=1$, then
    $\bar\Sigma_s\preceq\sigma^2I_d$ and \cref{fb:spectrum} give
    \begin{align}
        \Sigma_s\preceq2\sigma^2I_d,\qquad
        \|\Sigma_s^{-1}\|_{\op}\leq2n^2,\qquad
        I_d\preceq\bar\Sigma_{s+1}^{-1}\preceq n^2I_d.
        \ineqlabel{conc:stopped}
    \end{align}
    The rounds with $\chi_s=1$ are the first few rounds of the epoch, so
    \cref{fb:budget} gives
    \begin{align}
        \sum_{s=1}^{n_{\mathrm{epoch}}}\chi_s\E_s[Y_s^2]\leq5n.
        \ineqlabel{conc:budget}
    \end{align}

    \proofstepinline{3}
    We now introduce a \textbf{global} (across all epochs) high probability event $\mathcal G$ that captures
    the typical size of the Gaussian samples and the noise.
    Define
    \begin{align}
        W_s=X_s-\mu_s,\qquad Z_s=\Sigma_s^{-1/2}W_s.
    \end{align}
    During the epoch, conditionally on the past, $Z_s\sim\N(0,I_d)$, and
    the estimators are
    \begin{align}
        g_s&=Y_s\Sigma_s^{-1/2}Z_s,\notag\\
        H_s&=Y_s\Sigma_s^{-1/2}(Z_sZ_s^\top-I_d)\Sigma_s^{-1/2}.
        \eqlabel{conc:scores}
    \end{align}
    Let
    \begin{align}
        \mathcal T_s=\bigl\{\|Z_s\|^2\leq8(d+L),\
        |\xi_s|\leq\sqrt{8L},\ \pi(W_s)\leq1/2\bigr\}.
    \end{align}
    The chi-square inequality of \citet[Lemma~1]{laurent2000adaptive}, the
    noise assumption \cref{eq:noise}, and the gauge tail bound
    \cref{fb:gauge} imply, respectively,
    \begin{align}
        \Pp_s(\mathcal T_s^c)
        &\leq e^{-2L}+2e^{-8L}+e^{-L^4/128}
        \leq4e^{-2L}.
        \ineqlabel{conc:truncation-tail}
    \end{align}
    For the last tail, use $1/2-2/L^2\geq1/4$ and
    $\sigma\leq L^{-2}$. The gauge tail bound \cref{fb:gauge} needs
    $\Sigma_s\preceq2\sigma^2I_d$, which \cref{conc:stopped} gives when
    $\chi_s=1$, so \cref{conc:truncation-tail} holds at every round with
    $\chi_s=1$. Let $\mathcal G$ be the event that, in every epoch of the
    algorithm, $\mathcal T_s$ holds at every round $s$ with $\chi_s=1$. The
    algorithm runs for $n$ rounds in total, so the conditional tail bounds
    and a union bound give
    \begin{align}
        \Pp(\mathcal G^c)\leq4ne^{-2L}.
        \ineqlabel{conc:truncation-event}
    \end{align}
    Since $\mu_s\in\K_\eps$, the response formula and
    $v(X_s)\leq2\pi(W_s)$ show that, on $\mathcal T_s$,
    \begin{align}
        0\leq v(X_s)\leq1,\qquad
        |Y_s|\leq2(1+\sqrt{8L})+\frac2\eps\leq\frac{CL}{\eps}.
        \ineqlabel{conc:response-range}
    \end{align}

    \proofstepinline{4}
    We state one concentration bound that we will apply
    to several sums. In each application, $V_s$ is one of the following,
    for $s=1,\dots,n_{\mathrm{epoch}}$:
    \begin{itemize}
        \item $V_s=Y_s^2$;
        \item $V_s=u^\top H_su$ for a fixed unit vector $u$, whose sum
            controls the sampled inverse covariance;
        \item $V_s=q_s(x)$ for a fixed $x\in\mathbb B_{2d+1}$, whose sum
            is $Q_t(x)$;
        \item $V_s=\frac\lambda4\|x-\mu_s\|_{H_s}^2$ for a fixed
            $x\in\mathbb B_{2d+1}$, whose sum is $\Gamma_t(x)$.
    \end{itemize}
    In each case $V_s$ is determined by the history up to the end of
    round $s$.
    For the last three choices, the bound $\|\Sigma_s^{-1}\|_{\op}\leq2n^2$
    from \cref{conc:stopped}, together with $\|x-\mu_s\|\leq4d+1$ (since
    $\mu_s\in\K_\eps\subseteq\mathbb B_{2d}$) and $\lambda\leq1$, gives
    \begin{align}
        \chi_s|V_s|\leq Cn^4|Y_s|(1+\|Z_s\|^2).
        \ineqlabel{conc:crude}
    \end{align}
    For example, $|u^\top H_su|=|Y_s|\,|(u^\top\Sigma_s^{-1/2}Z_s)^2
    -u^\top\Sigma_s^{-1}u|\leq2n^2|Y_s|(1+\|Z_s\|^2)$, and the other two
    cases are similar, with the extra factor coming from $\|x-\mu_s\|^2$.

    \proofstepinline{4.1} We bound the fourth moment of $V_s$.
    Since $v(X_s)\leq2\|W_s\|\leq4\|Z_s\|$ and
    $Y_s=e_s(X_s)+(1+v(X_s))\xi_s$ with $0\leq f_s\leq1$, we have
    \begin{align}
        |Y_s|\leq\frac C\eps(1+|\xi_s|)(1+\|Z_s\|).
    \end{align}
    Conditionally on $X_s$ and the past, \cref{eq:noise} bounds every
    fixed moment of $\xi_s$, so
    $\E_s[|Y_s|^4\mid X_s]\leq C\eps^{-4}(1+\|Z_s\|)^4$. Combining this
    with \cref{conc:crude} gives
    \begin{align}
        \chi_s\E_s|V_s|^4
        \leq Cn^{16}\eps^{-4}\,\E(1+\|Z\|)^{12}
        \leq Cn^{24},\qquad Z\sim\N(0,I_d),
        \ineqlabel{conc:fourth-moment}
    \end{align}
    where the last inequality uses $\eps^{-4}\leq n^2$ and
    $\E\|Z\|^{12}=\prod_{j=0}^5(d+2j)\leq Cd^6\leq Cn^6$.

    \proofstepinline{4.2} We bound the effect of truncation on the
    conditional means. Let $O_s$ be either $\mathcal T_s^c$ or
    $\{|w_s^\top Z_s|>\sqrt{8L}\}$, where $w_s$ is a unit vector determined
    before the sample in round $s$.
    By \cref{conc:truncation-tail} and the Gaussian
    tail bound, $\chi_s\Pp_s(O_s)\leq4e^{-2L}$.
    H\"older's inequality and
    \cref{conc:fourth-moment} then give, for $j=1,2$,
    \begin{align}
        \chi_s\E_s[|V_s|^j\1_{O_s}]
        \leq\bigl(\chi_s\E_s|V_s|^4\bigr)^{j/4}
        \bigl(\chi_s\Pp_s(O_s)\bigr)^{1-j/4}
        \leq Cn^{12}e^{-L}.
        \ineqlabel{conc:bias}
    \end{align}

    \proofstepinline{4.3} We reduce to a martingale with bounded
    increments. Put
    \begin{align}
        D_s=\chi_s\bigl(V_s\1_{\mathcal T_s}-\E_s[V_s\1_{\mathcal T_s}]\bigr).
    \end{align}
    Since $\chi_s$ is determined before the sample in round $s$, we have
    $\E_sD_s=0$. On $\mathcal G$ we have $\chi_s\1_{\mathcal T_s}=\chi_s$, so for
    every $t\leq n_{\mathrm{epoch}}$,
    \begin{align}
        \sum_{s=1}^t\chi_s(V_s-\E_sV_s)
        =\sum_{s=1}^tD_s
        -\sum_{s=1}^t\chi_s\E_s[V_s\1_{\mathcal T_s^c}].
        \eqlabel{conc:centering}
    \end{align}
    By \cref{conc:bias} with $j=1$ and $O_s=\mathcal T_s^c$, the last sum is
    at most $Cn^{13}e^{-L}\leq1$ in absolute value, since
    $L=100\log(100nd/\delta)$.

    For the first sum, let $b>0$ be a number
    such that $\chi_s|V_s|\1_{\mathcal T_s}\leq b$ for every $s$. Then
    $|D_s|\leq2b$ and $\E_s[D_s^2]\leq \chi_s\E_s[V_s^2\1_{\mathcal T_s}]$.
    Applying \cref{lem:freedman} for each $t\leq n_{\mathrm{epoch}}$ shows
    that, for every $r,z>0$, with probability at least
    $1-2ne^{-z^2/2}$,
    \begin{align}
        \left|\sum_{s=1}^tD_s\right|<rz+\frac{2bz^2}{3}
        \qquad\text{for every }t\leq n_{\mathrm{epoch}}\text{ with }
        \sum_{s=1}^t\chi_s\E_s[V_s^2\1_{\mathcal T_s}]\leq r^2.
        \ineqlabel{conc:freedman}
    \end{align}
    Each of the following steps chooses $V_s$, finds $b$ and $r$, and
    applies \cref{conc:freedman}.

    \proofstepinline{5} We bound $\sum_{s=1}^tY_s^2$ and
    $\sum_{s=1}^t\|g_s\|_{\Sigma_s}^2$ first.
    To bound $\sum_{s=1}^tY_s^2$, take $V_s=Y_s^2$.
    By \cref{conc:response-range},
    $\chi_sY_s^2\1_{\mathcal T_s}\leq b$ with $b=CL^2/\eps^2$, and
    \cref{conc:budget} gives
    \begin{align}
        \sum_{s=1}^{n_{\mathrm{epoch}}}\chi_s\E_s[Y_s^4\1_{\mathcal T_s}]
        \leq b\sum_{s=1}^{n_{\mathrm{epoch}}}\chi_s\E_s[Y_s^2]\leq5bn.
    \end{align}
    Since $Y_s^2\geq0$, the last sum in \cref{conc:centering} is
    nonnegative, so we may drop it and do not need \cref{conc:bias}.
    Applying \cref{conc:freedman} with $r=\sqrt{5bn}$ and $z=2\sqrt L$
    shows that, with probability at least $1-2ne^{-2L}$, on $\mathcal G$,
    \begin{align}
        \sum_{s=1}^t\chi_sY_s^2
        \leq5n+2\sqrt{5bnL}+\frac{8bL}{3}
        \leq Cn
        \qquad\text{for every }t\leq n_{\mathrm{epoch}}.
        \ineqlabel{conc:feedback-energy}
    \end{align}
    The last inequality uses $bL\leq CL^3/\eps^2\leq Cn$, which follows from
    \cref{conc:parameters}.

    \proofstepinline{5.1} We bound $\sum_{s=1}^t\|g_s\|_{\Sigma_s}^2$.
    By \cref{conc:scores}, $\|g_s\|_{\Sigma_s}^2=Y_s^2\|Z_s\|^2$. On
    $\mathcal G$ we have $\chi_s\|Z_s\|^2\leq8(d+L)\chi_s$, so
    \cref{conc:feedback-energy} and $d+L\leq2dL$ give
    \begin{align}
        \sum_{s=1}^t\chi_s\|g_s\|_{\Sigma_s}^2
        \leq8(d+L)\sum_{s=1}^t\chi_sY_s^2
        \leq CdnL
        \qquad\text{for every }t\leq n_{\mathrm{epoch}}.
        \ineqlabel{conc:current}
    \end{align}
    The second bound in \cref{conc:energy} uses $\Sigma_{s+1}$ in place
    of $\Sigma_s$. We pass between the two after proving the covariance
    comparison.

    \proofstepinline{6} We prove the covariance comparison
    \cref{conc:covariance}.
    Both $\Sigma_s^{-1}$ and $\bar\Sigma_s^{-1}$ start at $\sigma^{-2}I_d$.
    At each round the first adds $\frac{\eta\lambda}2H_s$ and the second
    adds $\frac{\eta\lambda}2\bar H_s$. Hence
    \begin{align}
        \Sigma_{t+1}^{-1}-\bar\Sigma_{t+1}^{-1}
        =\frac{\eta\lambda}{2}\sum_{s=1}^t(H_s-\bar H_s).
        \eqlabel{conc:telescoping}
    \end{align}
    The comparison \cref{conc:covariance} therefore follows once we show
    that, for every unit vector $u$,
    \begin{align}
        \left|u^\top\sum_{s=1}^t(H_s-\bar H_s)u\right|
        \leq\frac{1}{2\eta\lambda}\,u^\top\bar\Sigma_{t+1}^{-1}u.
        \ineqlabel{conc:goal}
    \end{align}

    \proofstepinline{6.1} First we truncate the sum to the event $\mathcal T_s$.
    Fix a unit vector $u$, take $V_s=u^\top H_su$ in Step~4, and let
    \begin{align}
        D_s=\chi_s\bigl(u^\top H_su\,\1_{\mathcal T_s}
        -\E_s[u^\top H_su\,\1_{\mathcal T_s}]\bigr).
    \end{align}
    Let $t\leq n_{\mathrm{epoch}}$ with $\chi_t=1$, so that $\chi_s=1$ for every
    $s\leq t$. Since $\E_s[u^\top H_su]=u^\top\bar H_su$,
    \cref{conc:centering} and the bound on its last sum show that, on
    $\mathcal G$,
    \begin{align}
        \left|u^\top\sum_{s=1}^t(H_s-\bar H_s)u-\sum_{s=1}^tD_s\right|\leq1.
    \end{align}
    Since $\eta\lambda\leq1/4$ and
    $u^\top\bar\Sigma_{t+1}^{-1}u\geq\sigma^{-2}\geq1$, this error is at most
    $\frac1{4\eta\lambda}u^\top\bar\Sigma_{t+1}^{-1}u$. Hence
    \cref{conc:goal} follows once we show that
    \begin{align}
        \left|\sum_{s=1}^tD_s\right|
        \leq\frac{1}{4\eta\lambda}\,u^\top\bar\Sigma_{t+1}^{-1}u.
        \ineqlabel{conc:truncated-goal}
    \end{align}

    \proofstepinline{6.2}
    Note that the right side of \cref{conc:truncated-goal} is random, so no single choice of $r$ and $b$ in
    Freedman's inequality will give the bound we need.
    However, we do have a range for the right side:
    since $\chi_t=1$ by the choice of $t$, from \cref{fb:spectrum} we have
    \begin{align}
        \sigma^{-2}\leq u^\top\bar\Sigma_{t+1}^{-1} u \leq n^2\,.
    \end{align}
    We therefore cover this range by the levels
    \begin{align}
        a_k \triangleq 2^k\sigma^{-2},\qquad k=0,1,\dots,\lceil\log_2(n^2\sigma^2)\rceil,
    \end{align}
    of which there are at most $2\log_2n+2$. If $k$ is the smallest index
    with $a_k \geq u^\top\bar\Sigma_{t+1}^{-1}u$, then
    \begin{align}
        u^\top\bar\Sigma_{t+1}^{-1}u\leq
        a_k
        \leq2\,u^\top\bar\Sigma_{t+1}^{-1}u.
    \end{align}
    We apply Freedman's inequality once for each level, with $r$ and $b$
    chosen for that level, and then use the level that matches the right
    side of \cref{conc:truncated-goal}.

    \proofstepinline{6.3}
    Applying \cref{conc:freedman} also needs a bound on the absolute value of the increments $D_s$.
    To this end, observe that
    \begin{align}
        u^\top H_su
        =Y_s\bigl((u^\top\Sigma_s^{-1/2}Z_s)^2-u^\top\Sigma_s^{-1}u\bigr)
        =(u^\top\Sigma_s^{-1}u)\,Y_s\,(G_s^2-1),
        \eqlabel{conc:direction-form}
    \end{align}
    where $G_s=\frac{u^\top\Sigma_s^{-1/2}Z_s}{\sqrt{u^\top\Sigma_s^{-1}u}}$.
    Note that $\frac{u^\top\Sigma_s^{-1/2}}{\sqrt{u^\top\Sigma_s^{-1}u}}$
    has unit length and is determined before the sample in round $s$, and $Z_s$ is
    standard normal given the past, so $G_s\sim\mathcal N(0,1)$ is standard normal given the
    past.
    Now, on $\mathcal T_s$, \cref{conc:direction-form,conc:response-range} and
    $G_s^2\leq\|Z_s\|^2\leq8(d+L)$ give
    \begin{align}
        1_{\mathcal{T}_s}|u^\top H_su|\leq\bigl(u^\top\Sigma_s^{-1}u\bigr)\frac{CdL^2}{\eps}\,.
        \label{conc:direction-range}
    \end{align}
    Now, since $\chi_s=1$, we have that
    \begin{align}
        u^\top\Sigma_s^{-1}u\leq2\,u^\top\bar\Sigma_s^{-1}u
        \leq2\,u^\top\bar\Sigma_{t+1}^{-1}u,
    \end{align}
    where the last inequality uses that $\bar\Sigma_s^{-1}$ is increasing in $s$.
    Let
    \begin{align}
        \iota_{k,s}\triangleq 1\{u^\top \bar\Sigma_s^{-1} u \leq a_k\}
        \qquad(s\leq\tau),
    \end{align}
    and set $\iota_{k,s}=0$ for $s>\tau$. Note that $\iota_{k,s}$ is known
    before the sample in round $s$, so
    \begin{align}
        D_{k,s} \triangleq \iota_{k,s}D_s
    \end{align}
    is a martingale difference sequence.
    To apply Freedman's inequality to $D_{k,s}$, we need to bound the increments and their conditional second moments.

    \proofstepinline{6.3.1}
    First we bound the increments size $|D_{k,s}|$.
    From the definition of $D_{k,s}$, we have
    \begin{align}
        |D_{k,s}| = |\iota_{k,s}D_s|
        &= \iota_{k,s}\chi_s\left|
        u^\top H_su\,\1_{\mathcal T_s}
        -\E_s[u^\top H_su\,\1_{\mathcal T_s}]\right|\\
        &\leq\iota_{k,s}\chi_s\Bigl(|u^\top H_su|\,\1_{\mathcal T_s}
        +\E_s\bigl[|u^\top H_su|\,\1_{\mathcal T_s}\bigr]\Bigr).
    \end{align}
    When $\iota_{k,s}\chi_s=1$, we have
    $u^\top\Sigma_s^{-1}u\leq2\,u^\top\bar\Sigma_s^{-1}u\leq2a_k$, so \cref{conc:direction-range} gives
    $\iota_{k,s}\chi_s|u^\top H_su|\,\1_{\mathcal T_s}\leq2a_kCdL^2/\eps$.
    The conditional mean satisfies the same bound, and hence
    \begin{align}
        |D_{k,s}|\leq\frac{Ca_kdL^2}{\eps}.
        \ineqlabel{conc:level-range}
    \end{align}

    \proofstepinline{6.3.2}
    Next we bound the conditional second moments of $D_{k,s}$.
    Since $D_{k,s}=X-\E_s[X]$ with
    $X=\iota_{k,s}\chi_s\,u^\top H_su\,\1_{\mathcal T_s}$, we have
    $\E_s[D_{k,s}^2]=\E_s[X^2]-(\E_s[X])^2\leq\E_s[X^2]$, that is,
    \begin{align}
        \E_s[D_{k,s}^2]
        \leq\iota_{k,s}\chi_s\E_s\bigl[(u^\top H_su)^2\1_{\mathcal T_s}\bigr].
    \end{align}
    We split according to whether $|G_s|\leq\sqrt{8L}$. On this event
    $(G_s^2-1)^2\leq64L^2$, so when $\iota_{k,s}\chi_s=1$,
    \cref{conc:direction-form} and $u^\top\Sigma_s^{-1}u\leq2a_k$ give
    $(u^\top H_su)^2\leq256a_k^2L^2Y_s^2$. On its complement we use
    \cref{conc:bias} with $j=2$ and $O_s=\{|G_s|>\sqrt{8L}\}$. Summing over
    $s$ and using \cref{conc:budget}, we get
    \begin{align}
        \sum_{s=1}^{n_{\mathrm{epoch}}}\E_s[D_{k,s}^2]
        \leq256a_k^2L^2\sum_{s=1}^{n_{\mathrm{epoch}}}\chi_s\E_s[Y_s^2]
        +Cn^{13}e^{-L}
        \leq Ca_k^2L^2n.
        \ineqlabel{conc:level-variance}
    \end{align}
    The last inequality uses $Cn^{13}e^{-L}\leq1\leq a_k^2$.

    \proofstepinline{6.4}
    We apply Freedman's inequality at a fixed level. Fix $u$ and $k$, and
    put
    \begin{align}
        b=\frac{Ca_kdL^2}{\eps},\qquad r=Ca_kL\sqrt n,\qquad z=\sqrt{8dL},
    \end{align}
    so that $|D_{k,s}|\leq b$ by \cref{conc:level-range} and
    $\sum_{s=1}^t\E_s[D_{k,s}^2]\leq r^2$ for every $t$ by
    \cref{conc:level-variance}. Applying \cref{lem:freedman} for each
    $t\leq n_{\mathrm{epoch}}$ shows that, with probability at least
    $1-2ne^{-4dL}$,
    \begin{align}
        \left|\sum_{s=1}^tD_{k,s}\right|
        \leq rz+\frac{bz^2}{3}
        \leq Ca_k\left(L^{3/2}\sqrt{dn}+\frac{d^2L^3}{\eps}\right)
        \leq Ca_kL^2\sqrt{dn}
        \qquad\text{for every }t\leq n_{\mathrm{epoch}}.
        \ineqlabel{conc:level-freedman}
    \end{align}
    The last inequality uses $\eps\geq d^{3/2}L/\sqrt n$.

    \proofstepinline{6.5}
    We choose the level that matches the right side of
    \cref{conc:truncated-goal}. By a union bound over the at most
    $2\log_2n+2$ levels, with probability at least
    $1-2n(2\log_2n+2)e^{-4dL}$, \cref{conc:level-freedman} holds for every
    $k$ at once. Let $t\leq n_{\mathrm{epoch}}$ with $\chi_t=1$, and let $k$ be
    the smallest index with $a_k\geq u^\top\bar\Sigma_{t+1}^{-1}u$, so that
    $a_k\leq2\,u^\top\bar\Sigma_{t+1}^{-1}u$. Since $\bar\Sigma_s^{-1}$ is
    nondecreasing in $s$, we have
    $u^\top\bar\Sigma_s^{-1}u\leq u^\top\bar\Sigma_{t+1}^{-1}u\leq a_k$ for
    every $s\leq t$. Hence $\iota_{k,s}=1$ and $D_{k,s}=D_s$ for every
    $s\leq t$, and
    \begin{align}
        \left|\sum_{s=1}^tD_s\right|
        =\left|\sum_{s=1}^tD_{k,s}\right|
        \leq Ca_kL^2\sqrt{dn}
        \leq2CL^2\sqrt{dn}\,u^\top\bar\Sigma_{t+1}^{-1}u
        \leq\frac{1}{4\eta\lambda}\,u^\top\bar\Sigma_{t+1}^{-1}u.
    \end{align}
    The last inequality uses $\eta\lambda=1/(A^3\sqrt{dn}L^4)$ and that $A$
    is large. This proves \cref{conc:truncated-goal}, and hence, on
    $\mathcal G$, \cref{conc:goal}, for the fixed unit vector $u$.

    \proofstepinline{6.6}
    We extend \cref{conc:goal} from a fixed unit vector to all unit vectors.
    Let $h=n^{-40}$, and fix a set $\mathcal N$ of at most $(3/h)^d$ unit
    vectors such that every unit vector is within distance $h$ of some
    element of $\mathcal N$. A union bound over $\mathcal N$ shows that,
    with probability at least $1-(3/h)^d\,2n(2\log_2n+2)e^{-4dL}$, the
    conclusion of Step~6.5 holds at every $u_0\in\mathcal N$. Together with
    Step~6.1 this gives, on $\mathcal G$, for every $u_0\in\mathcal N$ and
    every $t\leq n_{\mathrm{epoch}}$ with $\chi_t=1$,
    \begin{align}
        \left|u_0^\top\sum_{s=1}^t(H_s-\bar H_s)u_0\right|
        \leq\frac{1}{4\eta\lambda}\,u_0^\top\bar\Sigma_{t+1}^{-1}u_0+1.
    \end{align}
    To pass to an arbitrary unit vector we use crude bounds on the
    matrices. On $\mathcal G$, for $s\leq t$,
    \cref{conc:scores,conc:stopped,conc:response-range} give
    $\|H_s\|_{\op}\leq|Y_s|\,\|\Sigma_s^{-1}\|_{\op}(\|Z_s\|^2+1)\leq Cn^5$,
    and \cref{fb:hessian-norm} gives
    $\|\bar H_s\|_{\op}\leq(5/\eps)\|\Sigma_s^{-1}\|_{\op}^{1/2}\leq Cn^2$.
    Hence
    \begin{align}
        \left\|\sum_{s=1}^t(H_s-\bar H_s)\right\|_{\op}\leq Cn^6.
    \end{align}
    Now let $u$ be a unit vector and choose $u_0\in\mathcal N$ with
    $\|u-u_0\|\leq h$. For any matrix $M$, the identity
    $u^\top Mu-u_0^\top Mu_0=(u-u_0)^\top Mu+u_0^\top M(u-u_0)$ gives
    $|u^\top Mu-u_0^\top Mu_0|\leq2h\|M\|_{\op}$. Applying this to the sum
    above and to $\bar\Sigma_{t+1}^{-1}$, whose operator norm is at most
    $n^2$, and using $u^\top\bar\Sigma_{t+1}^{-1}u\geq1$, we get
    \begin{align}
        \left|u^\top\sum_{s=1}^t(H_s-\bar H_s)u\right|
        \leq\frac{1+2hn^2}{4\eta\lambda}\,u^\top\bar\Sigma_{t+1}^{-1}u
        +1+Cn^6h
        \leq\frac{1}{2\eta\lambda}\,u^\top\bar\Sigma_{t+1}^{-1}u.
    \end{align}
    The last inequality uses $h=n^{-40}$, $\eta\lambda\leq1/16$ and
    $u^\top\bar\Sigma_{t+1}^{-1}u\geq1$. This proves \cref{conc:goal} for
    every unit vector $u$.

    \proofstepinline{6.7}
    We conclude the covariance comparison. Let $t\leq n_{\mathrm{epoch}}$
    with $\chi_t=1$. By \cref{conc:telescoping} and \cref{conc:goal} for every
    unit vector $u$,
    \begin{align}
        \left|u^\top\bigl(\Sigma_{t+1}^{-1}-\bar\Sigma_{t+1}^{-1}\bigr)u\right|
        \leq\tfrac14\,u^\top\bar\Sigma_{t+1}^{-1}u
        \qquad\text{for every }u,
    \end{align}
    that is,
    \begin{align}
        \tfrac34\bar\Sigma_{t+1}^{-1}\preceq\Sigma_{t+1}^{-1}
        \preceq\tfrac54\bar\Sigma_{t+1}^{-1}.
        \ineqlabel{conc:comparison-proved}
    \end{align}
    Here $\Sigma_{t+1}^{-1}=\Sigma_t^{-1}+\frac{\eta\lambda}2H_t$ is the
    matrix the algorithm forms in round $t$. By
    \cref{conc:comparison-proved} it is positive definite, so the algorithm
    does not take the fallback in round $t$. Since $\frac12\leq\frac34$ and
    $\frac54\leq2$, the comparison in the definition of $\chi_s$ also holds at
    round $t+1$, so $\chi_{t+1}=1$ whenever $t+1\leq\tau$.
    If an epoch starts at $s_{\mathrm{epoch}}$, then $\chi_1=1$, so induction
    gives $\chi_s=1$ for every $s\leq\tau$, and
    \cref{conc:comparison-proved} holds at every round of the epoch,
    including its final update. This proves \cref{conc:covariance}.

    \proofstepinline{6.8}
    We prove \cref{conc:energy}. Since $\chi_s=1$ for every
    $s\leq\tau$, the first bound in \cref{conc:energy} is
    \cref{conc:feedback-energy}. For the second, let
    $s\leq\tau$. By \cref{conc:comparison-proved} at rounds
    $s$ and $s+1$ (for $s=1$ use $\Sigma_1=\bar\Sigma_1$), and since
    $\bar\Sigma_s^{-1}$ is nondecreasing in $s$,
    \begin{align}
        \Sigma_{s+1}^{-1}\succeq\tfrac34\bar\Sigma_{s+1}^{-1}
        \succeq\tfrac34\bar\Sigma_s^{-1}
        \succeq\tfrac35\Sigma_s^{-1}.
    \end{align}
    Inverting gives $\Sigma_{s+1}\preceq\frac53\Sigma_s$, so
    $\|g_s\|_{\Sigma_{s+1}}^2\leq\frac53\|g_s\|_{\Sigma_s}^2$. Together with
    \cref{conc:current}, this gives
    \begin{align}
        \sum_{s=1}^t\|g_s\|_{\Sigma_{s+1}}^2\leq CdnL
        \qquad\text{for every }t\leq\tau,
    \end{align}
    which is the second bound in \cref{conc:energy}.

    \proofstepinline{7} We prove the bounds on the quadratic errors
    \cref{conc:quadratics}. By the definitions
    \cref{conc:quadratic-definitions},
    \begin{align}
        Q_t(x)-\bar Q_t(x)=\sum_{s=1}^t\bigl(q_s(x)-\bar q_s(x)\bigr),
        \qquad
        \Gamma_t(x)-\bar\Gamma_t(x)
        =\frac\lambda4\sum_{s=1}^t\|x-\mu_s\|_{H_s-\bar H_s}^2,
    \end{align}
    and since $\bar q_s(x)=\E_s[q_s(x)]$ and $\bar H_s=\E_s[H_s]$, both are
    martingales for each fixed $x$. As in Step~6, we first treat a fixed
    point and then pass to all points of the focus set with a net.

    \proofstepinline{7.1}
    We write the terms at a fixed point in closed form. Fix
    $x\in\mathbb B_{2d+1}$ and let
    $\theta_s=\|x-\mu_s\|_{\Sigma_s^{-1}}$. If $\theta_s>0$, let
    \begin{align}
        G_s=\frac{(x-\mu_s)^\top\Sigma_s^{-1/2}Z_s}{\theta_s}.
    \end{align}
    The vector $\Sigma_s^{-1/2}(x-\mu_s)/\theta_s$ has unit length and is
    determined before the sample in round $s$, so $G_s$ is standard normal
    given the past. By \cref{conc:scores},
    \begin{align}
        q_s(x)=Y_s\Bigl(\theta_sG_s+\frac\lambda2\theta_s^2(G_s^2-1)\Bigr),
        \qquad
        \frac\lambda4\|x-\mu_s\|_{H_s}^2
        =\frac\lambda4\theta_s^2\,Y_s(G_s^2-1).
        \eqlabel{conc:point-form}
    \end{align}
    Both terms vanish if $\theta_s=0$. For a point of the focus set we
    will have $\theta_s\leq R$, and the net in Step~7.5 needs a little
    room, so we keep only the rounds with $\theta_s\leq2R$. Let
    \begin{align}
        \iota_s=\1\{\theta_s\leq2R\}\qquad(s\leq\tau),
    \end{align}
    and set $\iota_s=0$ for $s>\tau$. This is determined before the sample
    in round $s$. Let $V_s$ be either
    of the two terms in \cref{conc:point-form}, and let $D_s$ be as in
    Step~4.3 for this $V_s$. Then $\iota_sD_s$ is a martingale difference
    sequence, and by \cref{conc:centering} and the bound on its last sum,
    on $\mathcal G$,
    \begin{align}
        \left|\sum_{s=1}^t\chi_s\iota_s(V_s-\E_sV_s)
        -\sum_{s=1}^t\iota_sD_s\right|\leq1
        \qquad\text{for every }t\leq n_{\mathrm{epoch}}.
        \ineqlabel{conc:point-centering}
    \end{align}

    \proofstepinline{7.2}
    We bound the increments $\iota_sD_s$. Since $R^2=1/(32\lambda^2L)$,
    when $\iota_s=1$ and $|G_s|\leq c$ the coefficient of $Y_s$ in either
    term of \cref{conc:point-form} is at most
    \begin{align}
        2Rc+2\lambda R^2c^2
        =\frac{1}{\lambda}\left(\frac{c}{\sqrt{8L}}+\frac{c^2}{16L}\right).
        \ineqlabel{conc:point-coefficient}
    \end{align}
    On $\mathcal T_s$ we have $|G_s|\leq\|Z_s\|\leq\sqrt{8(d+L)}$, so with
    $c=\sqrt{8(d+L)}$ and $d+L\leq2dL$ this coefficient is at most
    $Cd/\lambda$. Together with \cref{conc:response-range}, this gives
    $\chi_s\iota_s|V_s|\1_{\mathcal T_s}\leq CdL/(\lambda\eps)$, and the
    conditional mean satisfies the same bound. Hence
    \begin{align}
        |\iota_sD_s|\leq\frac{CdL}{\lambda\eps}.
        \ineqlabel{conc:point-range}
    \end{align}

    \proofstepinline{7.3}
    We bound the conditional second moments. As in Step~6.3.2,
    $\E_s[(\iota_sD_s)^2]\leq \chi_s\iota_s\E_s[V_s^2\1_{\mathcal T_s}]$. We
    split according to whether $|G_s|\leq\sqrt{8L}$. On this event,
    \cref{conc:point-coefficient} with $c=\sqrt{8L}$ shows that the
    coefficient of $Y_s$ is at most $2/\lambda$, so
    $V_s^2\leq4Y_s^2/\lambda^2$. On its complement we use \cref{conc:bias}
    with $j=2$ and $O_s=\{|G_s|>\sqrt{8L}\}$. Summing over $s$ and using
    \cref{conc:budget}, we get
    \begin{align}
        \sum_{s=1}^{n_{\mathrm{epoch}}}\E_s[(\iota_sD_s)^2]
        \leq\frac4{\lambda^2}\sum_{s=1}^{n_{\mathrm{epoch}}}\chi_s\E_s[Y_s^2]
        +Cn^{13}e^{-L}
        \leq\frac{Cn}{\lambda^2}.
        \ineqlabel{conc:point-variance}
    \end{align}

    \proofstepinline{7.4}
    We apply Freedman's inequality at the fixed point. Put
    \begin{align}
        b=\frac{CdL}{\lambda\eps},\qquad r=\frac{C\sqrt n}{\lambda},
        \qquad z=\sqrt{8dL}.
    \end{align}
    By \cref{conc:point-range,conc:point-variance}, \cref{lem:freedman}
    applies to $\iota_sD_s$. Applying it for each
    $t\leq n_{\mathrm{epoch}}$ and for each of the two choices of $V_s$,
    and using \cref{conc:point-centering}, shows that, with probability at
    least $1-4ne^{-4dL}$, on $\mathcal G$,
    \begin{align}
        \left|\sum_{s=1}^t\chi_s\iota_s(V_s-\E_sV_s)\right|
        \leq rz+\frac{bz^2}{3}+1
        \leq\frac C\lambda\left(\sqrt{dnL}+\frac{d^2L^2}{\eps}\right)+1
        \leq\frac\Delta2
        \ineqlabel{conc:point}
    \end{align}
    for both choices of $V_s$ and every $t\leq n_{\mathrm{epoch}}$. The last
    inequality uses $\eps\geq d^{3/2}L/\sqrt n$, $\lambda\leq1$, and the
    definition $\Delta=CL^2\sqrt{dn}/\lambda$ with $C$ large enough.

    \proofstepinline{7.5}
    We pass to every point of the focus set. Let $h=n^{-40}$, and fix a set
    $\mathcal N'$ of at most $(7dn^{40})^d$ points of $\mathbb B_{2d+1}$
    such that every point of $\mathbb B_{2d+1}$ is within distance $h$ of
    some element of $\mathcal N'$. A union bound over $\mathcal N'$ shows
    that, with probability at least $1-(7dn^{40})^d\,4ne^{-4dL}$,
    \cref{conc:point} holds at every $x_0\in\mathcal N'$.

    Let $t\leq\tau$ and $x\in K_t$, and choose
    $x_0\in\mathcal N'$ with $\|x-x_0\|\leq h$. The focus sets are nested
    and $K_s\subseteq E_s$, so $\|x-\mu_s\|_{\Sigma_s^{-1}}\leq R$ for every
    $s\leq t$. Since $\|\Sigma_s^{-1}\|_{\op}\leq2n^2$ and $R\geq1$, this
    gives
    \begin{align}
        \|x_0-\mu_s\|_{\Sigma_s^{-1}}
        \leq R+\sqrt2\,n\,h\leq2R
        \qquad\text{for every }s\leq t.
    \end{align}
    Hence $\iota_s=1$ at $x_0$ for every $s\leq t$, and $\chi_s=1$ by
    Step~6.7, so the two sums in \cref{conc:point} at $x_0$ are
    $Q_t(x_0)-\bar Q_t(x_0)$ and $\Gamma_t(x_0)-\bar\Gamma_t(x_0)$.

    It remains to pass from $x_0$ to $x$. For $z\in\mathbb B_{2d+1}$,
    \begin{align}
        \nabla(Q_t-\bar Q_t)(z)
        &=\sum_{s=1}^t\bigl(g_s-\bar g_s
        +\lambda(H_s-\bar H_s)(z-\mu_s)\bigr),\notag\\
        \nabla(\Gamma_t-\bar\Gamma_t)(z)
        &=\frac\lambda2\sum_{s=1}^t(H_s-\bar H_s)(z-\mu_s).
    \end{align}
    On $\mathcal G$, \cref{conc:scores,conc:stopped,conc:response-range}
    give $\|g_s\|\leq|Y_s|\,\|\Sigma_s^{-1}\|_{\op}^{1/2}\|Z_s\|\leq Cn^3$,
    and \cref{lem:extension} gives $\|\bar g_s\|\leq5/\eps$. Together with
    the bound $\|H_s-\bar H_s\|_{\op}\leq Cn^5$ from Step~6.6,
    $\lambda\leq1$ and $\|z-\mu_s\|\leq4d+1$, both gradients have norm at
    most $Cn^7$. The segment joining $x$ to $x_0$ lies in
    $\mathbb B_{2d+1}$, so each error changes by at most
    $Cn^7h\leq1\leq\Delta/2$ between $x_0$ and $x$. Therefore
    \begin{align}
        |Q_t(x)-\bar Q_t(x)|\leq\Delta,\qquad
        |\Gamma_t(x)-\bar\Gamma_t(x)|\leq\Delta
        \qquad\text{for every }x\in K_t,
    \end{align}
    which proves \cref{conc:quadratics}.

    \proofstepinline{8}
    We collect the failure probabilities. For a fixed starting round,
    Steps~5, 6 and~7 fail with probability at most
    \begin{align}
        2ne^{-2L}+(3n^{40})^d\,2n(2\log_2n+2)e^{-4dL}
        +(7dn^{40})^d\,4ne^{-4dL}
        \leq2ne^{-2L}+e^{-3dL},
    \end{align}
    where the inequality uses $L=100\log(100nd/\delta)$. Taking a union
    bound over the at most $n$ possible starting rounds,
    and adding the bound \cref{conc:truncation-event} on
    $\Pp(\mathcal G^c)$, the total failure probability is at most
    \begin{align}
        n\bigl(2ne^{-2L}+e^{-3dL}\bigr)+4ne^{-2L}
        \leq7n^2e^{-2L}\leq\frac\delta4.
    \end{align}
    Every epoch of the algorithm begins at some round, so on the
    complement of these events Steps~5--7 apply to
    every epoch. This proves \cref{conc:uniform}.
\end{proof}

\section{Regret of the sampled algorithm}
\label{transfer:section}

We now prove \cref{thm:main} for the trajectory generated by the sampled
updates. The quantities $\bar q_t$ and $\bar Q_t$ below are evaluated
along this trajectory. They are not the quadratics produced by running
the population algorithm separately.

The concentration result supplies two kinds of control. The covariance
comparison makes the updates valid and allows us to use regret switching.
The uniform errors let us replace the sampled quadratics by their
conditional means after summing the online Newton step inequality.
We then prove directly that a round without a restart has a minimizer
of $\bar Q_t$ over $\K_\eps$ inside the focus region. This is the premise
needed to compare against every loss comparator, including those
outside the focus region.

\begin{proof}[Proof of \cref{thm:main}]
    \begingroup\allowdisplaybreaks[1]
    If $\eps>1/2$, the fixed-action branch of the algorithm satisfies
    $\Reg_n\leq n<2n\eps=2\rho/\eta$, so the claimed bound holds
    without a probability argument. Henceforth suppose $\eps\leq1/2$.
    Work on the event of \cref{conc:uniform}, which has probability at least
    $1-\delta/4$.

    Within each epoch, the index starts at one. Write
    \begin{align}
        F_s=F_{s,\Sigma_s}(\mu_s),\qquad
        c_s=2\tr(\bar H_s\Sigma_s)+n^{-3}.
        \eqlabel{transfer:round-cost}
    \end{align}
    The estimates of \cref{conc:uniform} hold simultaneously over all epoch
    prefixes, including the update after a restart round. In particular,
    every proposed inverse covariance is positive definite, so the
    algorithm never takes its fallback branch. The focus sets are nonempty:
    $\mu_1=0\in K_0$, and inductively $\mu_s\in K_{s-1}\cap E_s=K_s$
    because the preceding projection puts $\mu_s$ in $K_{s-1}$ and
    $\mu_s$ is the center of $E_s$. The objective used by the restart test
    also has a minimizer, since $K_s$ is compact. In fact,
    \begin{align}
        \nabla^2J_s
        =\eta\lambda\sum_{j=1}^sH_j+2\Sigma_1^{-1}
        =2\Sigma_{s+1}^{-1}\succ0.
    \end{align}
    Thus both optimization problems in the algorithm are well-defined.

    \proofstepinline{1}
    Let $C$ dominate the absolute constants in \cref{conc:uniform}, and
    set $\Delta=CL^2\sqrt{dn}/\lambda$. The constant $C$ is fixed before
    choosing $A$. We will use the following consequences of the parameters:
    \begin{align}
        \frac{2d^2}{\sigma^2}
        +\frac{C\eta^2dnL}{2}+2\eta\Delta&\leq\frac\rho4,
        &\eta\Delta&\leq\frac\rho{16},\nonumber\\
        \frac{4d^2}{\sigma^2}&\leq\frac\rho8,
        &R^2&\geq256\rho,\nonumber\\
        \eta\left(1+\sum_{s=1}^tc_s\right)&\leq\frac\rho4
        &&\text{for every epoch prefix }t.
        \ineqlabel{transfer:parameter-bounds}
    \end{align}
    Here is why one sufficiently large universal $A$ gives all these
    inequalities. Substituting \cref{eq:parameters} yields
    \begin{align}
        \frac{2d^2}{\sigma^2\rho}&=\frac2{A^3},
        &\frac{C\eta^2dnL}{2\rho}&=\frac{C}{2A^5L^5},
        &\frac{\eta\Delta}{\rho}&=\frac C{A^2},\nonumber\\
        \frac{4d^2}{\sigma^2\rho}&=\frac4{A^3},
        &\frac{R^2}{\rho}&=\frac{AL}{32}.
    \end{align}
    The first three terms therefore have sum at most
    $2/A^3+C/(2A^5)+2C/A^2$, which can be made at most $1/4$.
    The separate error and initial-distance bounds follow by increasing
    $A$ further.
    The radius bound follows, for example, from $A\geq8192$
    and $L\geq1$.
    For the final inequality in \cref{transfer:parameter-bounds},
    \begin{align}
        \frac\eta\rho\left(1+\sum_{s=1}^tc_s\right)
        &\explan{\leq}{a}\frac\eta\rho(1+n^{-2})
        +\frac{32dL}{\lambda\rho}\nonumber\\
        &\explan{\leq}{b}\frac{2}{A^4d^2ML^5\sqrt n}
        +\frac{32}{A\sqrt d\,M}\nonumber\\
        &\explan{\leq}{c}\frac2{A^4}+\frac{32}{A}.
    \end{align}
    Here \reason{a} uses \cref{fb:trace} and $t\leq n$; \reason{b}
    substitutes \cref{eq:parameters} and uses $n^{-2}\leq1$; and
    \reason{c} uses $M\geq d^{-1/2}$ and $d,n,L\geq1$.
    The last expression is at most $1/4$ for a universal choice of $A$.
    Since $C$ does not depend on $A$, we can also take $A$ large enough for \cref{conc:uniform} to apply

    \proofstepinline{2}
    Recall from \cref{conc:quadratic-definitions} that $\Gamma_t$ uses
    $H_s$ and $\bar\Gamma_t$ uses $\bar H_s$. The latter is nonnegative,
    whereas the former need not be. We first derive the inequality with
    the sampled matrices, keeping both terms until we can apply concentration.

    Fix $x\in K_t$ and $s\leq t$. Since $K_t\subseteq K_s$, the point
    $x$ is admissible for the projection defining $\mu_{s+1}$. Therefore
    \begin{align}
        \frac12\normm{\mu_{s+1}-x}{\Sigma_{s+1}^{-1}}^2
        &\explan{\leq}{a}
        \frac12\normm{\mu_s-\eta\Sigma_{s+1}g_s-x}
        {\Sigma_{s+1}^{-1}}^2\nonumber\\
        &=\frac12\normm{\mu_s-x}{\Sigma_{s+1}^{-1}}^2
        +\eta\ip{g_s}{x-\mu_s}
        +\frac{\eta^2}{2}\normm{g_s}{\Sigma_{s+1}}^2\nonumber\\
        &\explan{=}{b}\frac12\normm{\mu_s-x}{\Sigma_s^{-1}}^2
        +\eta q_s(x)-\frac{\eta\lambda}{4}
        \normm{x-\mu_s}{H_s}^2
        +\frac{\eta^2}{2}\normm{g_s}{\Sigma_{s+1}}^2.
    \end{align}
    Here \reason{a} is the projection inequality in the positive definite
    metric $\Sigma_{s+1}^{-1}$; and \reason{b} substitutes the covariance
    update and $q_s(x)=\ip{g_s}{x-\mu_s}
    +(\lambda/2)\normm{x-\mu_s}{H_s}^2$, which is an algebraic identity
    even when $H_s$ is indefinite.
    Summing over $s$ telescopes the squared distances and gives
    \begin{align}
        \frac12\normm{\mu_{t+1}-x}{\Sigma_{t+1}^{-1}}^2+\eta\Gamma_t(x)
        &\leq\eta Q_t(x)+\frac12\normm{x}{\Sigma_1^{-1}}^2
        +\frac{\eta^2}{2}\sum_{s=1}^t\normm{g_s}{\Sigma_{s+1}}^2.
        \ineqlabel{transfer:sampled-ons}
    \end{align}
    We used $\mu_1=0$ here. Since $x\in K_t\subseteq\mathbb B_{2d}$,
    the initial term is at most $2d^2/\sigma^2$.
    Hence
    \begin{align}
        &\frac12\normm{\mu_{t+1}-x}{\Sigma_{t+1}^{-1}}^2
        +\eta\bar\Gamma_t(x)\nonumber\\
        &\quad\explan{\leq}{a}\eta\bar Q_t(x)
        +\frac{2d^2}{\sigma^2}+\frac{C\eta^2dnL}{2}
        +\eta|Q_t(x)-\bar Q_t(x)|
        +\eta|\Gamma_t(x)-\bar\Gamma_t(x)|\nonumber\\
        &\quad\explan{\leq}{b}\eta\bar Q_t(x)
        +\frac{2d^2}{\sigma^2}+\frac{C\eta^2dnL}{2}
        +2\eta\Delta\nonumber\\
        &\quad\explan{\leq}{c}\eta\bar Q_t(x)+\frac\rho4.
        \ineqlabel{transfer:population-ons}
    \end{align}
    Here \reason{a} uses \cref{transfer:sampled-ons}, the bound on the
    initial term and the second bound in \cref{conc:energy}; \reason{b}
    uses both uniform errors in \cref{conc:quadratics}; and \reason{c}
    uses \cref{transfer:parameter-bounds}.
    We have obtained a nonnegative curvature term without assuming that
    any sampled Hessian is positive semidefinite. Both this term and the
    terminal squared distance will be needed below.

    \proofstepinline{3} We show that if round $t$ does not restart, then $\bar Q_t$ has a minimizer over $\K_\eps$ that lies in $K_t$.
    For each prefix choose
    \begin{align}
        x_t^q\in\argmin_{x\in K_t}\bar Q_t(x).
    \end{align}
    This minimizer exists because $K_t$ is compact and nonempty.
    We will show by induction that, whenever round $t$ does not restart,
    \begin{align}
        x_t^q&\in\argmin_{x\in\K_\eps}\bar Q_t(x),
        &x_t^q&\in K_{t+1},
        &\eta\bar\Gamma_t(\mu_{t+1})&\leq18\rho.
        \ineqlabel{transfer:focus-induction}
    \end{align}
    At time zero take $x_0^q=0$. Then $\bar Q_0=\bar\Gamma_0=0$ and
    $0=\mu_1\in K_1$.

    We first record two inequalities for the population curvature. Its
    Hessian satisfies
    \begin{align}
        \eta\nabla^2\bar\Gamma_j
        =\frac{\eta\lambda}{2}\sum_{s=1}^j\bar H_s
        =\bar\Sigma_{j+1}^{-1}-\Sigma_1^{-1}.
    \end{align}
    For any $u,z\in\R^d$, expand $u-\mu_s=(z-\mu_s)+(u-z)$ and
    use $\|a+b\|_B^2\leq2\|a\|_B^2+2\|b\|_B^2$ for $B\succeq0$.
    After summing, this gives
    \begin{align}
        \eta\bar\Gamma_j(u)
        &\leq2\eta\bar\Gamma_j(z)
        +\normm{u-z}{\bar\Sigma_{j+1}^{-1}-\Sigma_1^{-1}}^2.
        \ineqlabel{transfer:curvature-upper}
    \end{align}
    Applying the same quadratic inequality to
    $u-z=(u-\mu_s)-(z-\mu_s)$ gives
    \begin{align}
        \normm{u-z}{\bar\Sigma_{j+1}^{-1}-\Sigma_1^{-1}}^2
        &\leq\eta\lambda\sum_{s=1}^j
        \left(\normm{u-\mu_s}{\bar H_s}^2
        +\normm{z-\mu_s}{\bar H_s}^2\right)
        =4\eta\bar\Gamma_j(u)+4\eta\bar\Gamma_j(z).
        \ineqlabel{transfer:curvature-lower}
    \end{align}
    These statements allow singular $\bar H_s$; no inverse of a Hessian
    or of $\bar\Sigma_{j+1}^{-1}-\Sigma_1^{-1}$ is used.

    Suppose round $t$ is active and the induction holds for every preceding
    round in the epoch. Thus
    $\eta\bar\Gamma_{r-1}(\mu_r)\leq18\rho$ for $1\leq r\leq t$.
    Take any $x\in K_t\cap\partial E_r$, where $r\leq t$.
    We show that the curvature at such a point is large.
    We obtain
    \begin{align}
        \eta\bar\Gamma_t(x)
        &\explan{\geq}{a}\eta\bar\Gamma_{r-1}(x)\nonumber\\
        &\explan{\geq}{b}\frac14\normm{x-\mu_r}
        {\bar\Sigma_r^{-1}-\Sigma_1^{-1}}^2
        -\eta\bar\Gamma_{r-1}(\mu_r)\nonumber\\
        &\explan{\geq}{c}\frac18\normm{x-\mu_r}{\Sigma_r^{-1}}^2
        -\frac14\normm{x-\mu_r}{\Sigma_1^{-1}}^2-18\rho\nonumber\\
        &\explan{\geq}{d}\frac{R^2}{8}-\frac{4d^2}{\sigma^2}-18\rho
        \explan{\geq}{e}4\rho.
        \ineqlabel{transfer:boundary-curvature}
    \end{align}
    Here \reason{a} holds because $\bar H_s\succeq0$, so
    $\bar\Gamma_j(x)$ is nondecreasing in $j$; \reason{b} is
    \cref{transfer:curvature-lower} at $j=r-1$; \reason{c} uses
    $\Sigma_r^{-1}\preceq2\bar\Sigma_r^{-1}$ and the bound on
    $\eta\bar\Gamma_{r-1}(\mu_r)$; \reason{d} uses $x\in\partial E_r$ and
    $\|x-\mu_r\|\leq4d$, since both points lie in $\K_\eps$; and
    \reason{e} uses \cref{transfer:parameter-bounds}, which gives
    $R^2/8-4d^2/\sigma^2-18\rho\geq32\rho-\rho/8-18\rho>4\rho$.
    Only the bounds on the preceding centers were used here.

    If $x_t^q$ lies on the boundary of any of $E_1,\ldots,E_t$, then
    \begin{align}
        \eta Q_t(y_t)
        &\explan{\geq}{a}\eta\bar Q_t(y_t)-\eta\Delta\nonumber\\
        &\explan{\geq}{b}\eta\bar Q_t(x_t^q)-\eta\Delta\nonumber\\
        &\explan{\geq}{c}\frac12\normm{\mu_{t+1}-x_t^q}{\Sigma_{t+1}^{-1}}^2
        +\eta\bar\Gamma_t(x_t^q)-\frac\rho4-\eta\Delta\nonumber\\
        &\explan{\geq}{d}4\rho-\frac\rho4-\frac\rho{16}>2\rho.
    \end{align}
    Here \reason{a} is uniform concentration; \reason{b} is minimality of
    $x_t^q$ over $K_t$; \reason{c} rearranges
    \cref{transfer:population-ons}; and \reason{d} drops a nonnegative
    squared distance and uses \cref{transfer:boundary-curvature} and
    $\eta\Delta\leq\rho/16$.
    Thus touching a focus ellipsoid forces the actual sampled restart test
    to fire.

    Consequently, when round $t$ does not restart, $x_t^q$ lies in the
    interior of every $E_r$, $r\leq t$. To see that it then minimizes
    $\bar Q_t$ over all of $\K_\eps$, fix $y\in\K_\eps$. For all sufficiently
    small $h>0$, the point $x_t^q+h(y-x_t^q)$ belongs to every $E_r$;
    it belongs to $\K_\eps$ by convexity. Hence
    \begin{align}
        \ip{\nabla\bar Q_t(x_t^q)}{y-x_t^q}
        =\lim_{h\downarrow0}
        \frac{\bar Q_t(x_t^q+h(y-x_t^q))-\bar Q_t(x_t^q)}h
        \geq0.
        \ineqlabel{transfer:global-first-order}
    \end{align}
    Convexity of $\bar Q_t$ now gives
    $\bar Q_t(y)\geq\bar Q_t(x_t^q)$ for every $y\in\K_\eps$.

    It remains to prove the other two assertions in
    \cref{transfer:focus-induction}, so that the argument can continue.
    On a round without a restart,
    \begin{align}
        \eta\bar Q_t(x_t^q)
        \leq\eta\bar Q_t(y_t)
        \leq\eta Q_t(y_t)+\eta\Delta
        <2\rho+\frac\rho{16}.
    \end{align}
    Substituting this into \cref{transfer:population-ons} yields
    \begin{align}
        \frac12\normm{\mu_{t+1}-x_t^q}{\Sigma_{t+1}^{-1}}^2
        +\eta\bar\Gamma_t(x_t^q)
        <2\rho+\frac\rho{16}+\frac\rho4<3\rho.
        \ineqlabel{transfer:distance-curvature}
    \end{align}
    Both terms are nonnegative. Thus the squared distance is at most
    $6\rho<R^2$, and $x_t^q\in E_{t+1}\cap K_t=K_{t+1}$.
    For the new center,
    \begin{align}
        \eta\bar\Gamma_t(\mu_{t+1})
        &\explan{\leq}{a}2\eta\bar\Gamma_t(x_t^q)
        +\normm{\mu_{t+1}-x_t^q}
        {\bar\Sigma_{t+1}^{-1}-\Sigma_1^{-1}}^2\nonumber\\
        &\explan{\leq}{b}2\eta\bar\Gamma_t(x_t^q)
        +2\normm{\mu_{t+1}-x_t^q}{\Sigma_{t+1}^{-1}}^2\nonumber\\
        &\explan{\leq}{c}6\rho+12\rho=18\rho.
    \end{align}
    Here \reason{a} is \cref{transfer:curvature-upper}; \reason{b} uses
    $\bar\Sigma_{t+1}^{-1}\preceq2\Sigma_{t+1}^{-1}$, which the covariance
    comparison gives including after the current update; and \reason{c}
    uses \cref{transfer:distance-curvature}. This closes the induction.

    \proofstepinline{4} We apply the regret-switching bound at a global quadratic minimizer.
    The covariance comparison implies $\Sigma_s\preceq2\sigma^2I_d$, so
    \cref{sw:tangent} applies. For $x\in K_t$ and $y\in\K_\eps$,
    \begin{align}
        \sum_{s=1}^t(F_s-e_s(y))
        &\explan{\leq}{a}-\bar Q_t(y)
        +\frac\lambda2\sum_{s=1}^t\normm{x-y}{\bar H_s}^2
        +\sum_{s=1}^tc_s\nonumber\\
        &\explan{=}{b}-\bar Q_t(x)-\ip{\nabla\bar Q_t(x)}{y-x}
        +\sum_{s=1}^tc_s.
        \ineqlabel{transfer:switch-sum}
    \end{align}
    Here \reason{a} sums \cref{sw:tangent} over $s\leq t$; and \reason{b}
    is the exact quadratic expansion of $\bar Q_t$ at $x$.
    The same $x$ is admissible in every round because
    $K_t\subseteq E_s$ for $s\leq t$.
    If $x$ is a global minimizer of $\bar Q_t$ over $\K_\eps$, its
    first-order condition makes the inner product nonnegative. Hence
    \begin{align}
        \sum_{s=1}^t(F_s-e_s(y))
        \leq-\bar Q_t(x)+\sum_{s=1}^tc_s
        \qquad(y\in\K_\eps).
        \ineqlabel{transfer:global-switch}
    \end{align}
    This is a simultaneous statement for every loss comparator $y$.

    \proofstepinline{5} Every completed epoch has nonpositive extension regret.
    Suppose round $t$ restarts. Let $z=x_{t-1}^q$, with $z=0$ when $t=1$.
    The preceding rounds did not restart, so Step 3 gives
    $z\in K_t$ and global minimality of $\bar Q_{t-1}$ at $z$.
    Apply \cref{transfer:global-switch} to the first $t-1$ rounds:
    \begin{align}
        \sum_{s=1}^{t-1}(F_s-e_s(y))
        \leq-\bar Q_{t-1}(z)+\sum_{s=1}^{t-1}c_s.
    \end{align}
    For $t=1$ both sides are zero. In the last round, apply regret switching
    with both points equal to $z$, which removes the curvature-distance
    term. Since $z,y\in\K_\eps$ and $e_t=f_t\in[0,1]$ there,
    \begin{align}
        F_t-e_t(y)
        &=F_t-e_t(z)+e_t(z)-e_t(y)
        \leq-\bar q_t(z)+c_t+1.
    \end{align}
    Adding the two inequalities gives
    \begin{align}
        \eta\sum_{s=1}^t(F_s-e_s(y))
        \leq-\eta\bar Q_t(z)+\eta\left(1+\sum_{s=1}^tc_s\right).
        \ineqlabel{transfer:restart-regret}
    \end{align}

    The restart test makes the first term sufficiently negative. Indeed,
    \begin{align}
        \eta Q_t(z)
        &\explan{\geq}{a}\eta Q_t(y_t)+\normm{y_t}{\Sigma_1^{-1}}^2
        -\normm{z}{\Sigma_1^{-1}}^2-\frac\rho8
        \explan{\geq}{b}2\rho-\frac{4d^2}{\sigma^2}-\frac\rho8
        \explan{\geq}{c}\frac{7\rho}{4}.
    \end{align}
    Here \reason{a} uses $J_t(y_t)\leq J_t(z)+\rho/8$; \reason{b} uses
    $\eta Q_t(y_t)\geq2\rho$ and $z\in\K_\eps\subseteq\mathbb B_{2d}$; and
    \reason{c} uses \cref{transfer:parameter-bounds}.
    Since $z\in K_t$, uniform concentration then gives
    \begin{align}
        \eta\bar Q_t(z)\geq\eta Q_t(z)-\eta\Delta
        \geq\frac{7\rho}{4}-\frac\rho{16}
        =\frac{27\rho}{16}.
    \end{align}
    Substitution into \cref{transfer:restart-regret} yields
    \begin{align}
        \eta\sum_{s=1}^t(F_s-e_s(y))
        \leq-\frac{27\rho}{16}+\frac\rho4
        =-\frac{23\rho}{16}<0
        \qquad(y\in\K_\eps).
        \ineqlabel{transfer:completed-epoch}
    \end{align}
    In particular, this conclusion includes the round that triggers the
    restart; it does not charge a positive error for each completed epoch.

    \proofstepinline{6} Only the final epoch contributes positive regret.
    If the last epoch does not end in a restart, let $t$ be its length.
    By Step 3, $x_t^q$ is a global minimizer in $K_t$. Hence
    \begin{align}
        \eta\sum_{s=1}^t(F_s-e_s(y))
        &\explan{\leq}{a}\frac\rho4
        -\frac12\normm{\mu_{t+1}-x_t^q}{\Sigma_{t+1}^{-1}}^2
        -\eta\bar\Gamma_t(x_t^q)+\eta\sum_{s=1}^tc_s\nonumber\\
        &\explan{\leq}{b}\frac\rho2
        -\frac12\normm{\mu_{t+1}-x_t^q}{\Sigma_{t+1}^{-1}}^2
        -\eta\bar\Gamma_t(x_t^q)\nonumber\\
        &\explan{\leq}{c}\rho.
        \ineqlabel{transfer:last-epoch}
    \end{align}
    Here \reason{a} combines \cref{transfer:global-switch,transfer:population-ons};
    \reason{b} uses \cref{transfer:parameter-bounds}; and \reason{c}
    uses nonnegativity of the two retained terms.
    If the last calendar round itself triggers a restart, there is no
    uncompleted epoch and every epoch satisfies \cref{transfer:completed-epoch}.

    Return to calendar time, using one comparator $y\in\K_\eps$ in every
    epoch. Sum \cref{transfer:completed-epoch} over the completed epochs
    and add \cref{transfer:last-epoch} if an uncompleted epoch remains.
    Taking the supremum over $y$ only after this summation gives
    \begin{align}
        \sup_{y\in\K_\eps}\sum_{t=1}^n(F_t-e_t(y))\leq\frac\rho\eta.
        \ineqlabel{transfer:regret}
    \end{align}
    The bounds used above hold simultaneously for all $y$, so this also
    allows the best comparator to depend on the realized loss sequence.

    \proofstepinline{7} We return to realized losses.
    Define, for the actual actions of the algorithm,
    \begin{align}
        \mathcal M_n
        =\sum_{t=1}^n\bigl(f_t(A_t)-\E_t[f_t(A_t)]\bigr).
    \end{align}
    Each summand is centered given the past and lies in $[-1,1]$, since
    $f_t$ takes values in $[0,1]$. Azuma's inequality therefore gives
    \begin{align}
        \mathcal M_n\leq\sqrt{2n\log(2/\delta)}
        \quad\text{with probability at least }1-\delta/2.
        \ineqlabel{transfer:martingale}
    \end{align}
    This event does not involve a comparator.

    On the concentration event the algorithm samples from its Gaussian in
    every round. By \cref{lem:extension},
    $\E_t[f_t(A_t)]\leq\E_t[e_t(X_t)]=F_t$.
    For any $x\in\K$, let $y=(1-\eps)x\in\K_\eps$. Convexity and
    $f_t\in[0,1]$ give
    \begin{align}
        e_t(y)=f_t((1-\eps)x)
        \leq(1-\eps)f_t(x)+\eps f_t(0)
        \leq f_t(x)+\eps.
    \end{align}
    Consequently, for every $x\in\K$,
    \begin{align}
        \sum_{t=1}^n(f_t(A_t)-f_t(x))
        &=\mathcal M_n+\sum_{t=1}^n(\E_t[f_t(A_t)]-f_t(x))\nonumber\\
        &\leq\mathcal M_n+n\eps+\sum_{t=1}^n(F_t-e_t((1-\eps)x)).
    \end{align}
    The map $x\mapsto(1-\eps)x$ takes $\K$ onto $\K_\eps$.
    Taking the supremum gives the precise reduction
    \begin{align}
        \Reg_n\leq\mathcal M_n+n\eps
        +\sup_{y\in\K_\eps}\sum_{t=1}^n(F_t-e_t(y)).
        \ineqlabel{eq:reduction}
    \end{align}
    Thus no concentration bound is being applied at a random hindsight
    comparator: only the learner's scalar loss fluctuations are handled by
    \cref{transfer:martingale}.

    The intersection of the concentration event and
    \cref{transfer:martingale} has probability at least
    $1-\delta/4-\delta/2\geq1-\delta$. On this intersection,
    \begin{align}
        \Reg_n
        &\explan{\leq}{a}\sqrt{2n\log(2/\delta)}+n\eps+\frac\rho\eta
        \explan{\leq}{b}3A^4d^2ML^5\sqrt n=\frac{3\rho}{\eta}.
    \end{align}
    Here \reason{a} combines
    \cref{eq:reduction,transfer:martingale,transfer:regret}; and \reason{b}
    uses $n\eps=\rho/\eta=A^4d^2ML^5\sqrt n$ and
    $2\log(2/\delta)\leq L$, together with $A\geq1$, $M\geq d^{-1/2}$
    and $d,L\geq1$. This proves \cref{thm:main}.
    \endgroup
\end{proof}

\section{Affine preprocessing}
\label{app:positioning}

The bound on $M$ needed for the regret guarantee can be obtained by putting
the scaled polar body in approximately isotropic position. The following
proposition states that the required geometric bounds can be achieved for
any convex body by randomized polynomial-time preprocessing.

\begin{proposition}
    \label{prop:positioning}
    Let $\K\subset\R^d$ be a convex body given by a separation oracle,
    together with known $x_0\in\R^d$ and $0<r\leq R$ such that
    $x_0+\mathbb B_r\subseteq\K\subseteq x_0+\mathbb B_R$.
    For every $\delta\in(0,1)$, there is a randomized algorithm that
    returns an invertible affine map $T$ such that, with probability at
    least $1-\delta$, the body $J=T(\K)$ satisfies
    \begin{align}
        \mathbb B_1\subseteq J\subseteq\mathbb B_{2d},
        \qquad
        \max\left\{\frac1{\sqrt d},\E[\pi_J(U)]\right\}
        \leq C\sqrt{\frac{\log(ed)}d}.
        \ineqlabel{eq:positioning}
    \end{align}
    Here $U$ is uniform on the Euclidean unit sphere, $\pi_J$ is the
    Minkowski functional of $J$, and $C\geq1$ is a universal constant.
    The number of oracle calls and arithmetic operations is polynomial
    in $d$, $\log(R/r)$, and $\log(1/\delta)$.
\end{proposition}

For a convex body $J$ with the origin in its interior, the polar body is
\begin{align}
    J^\circ=\left\{y\in\R^d:\ip{x}{y}\leq1\text{ for all }x\in J\right\}.
\end{align}
For $s$ in the interior of $\K$, the volume of $(\K-s)^\circ$ is a strictly
convex function of $s$ that tends to infinity as $s$ approaches the boundary
of $\K$. The Santal\'o point $s(\K)$ is the unique point at which this volume
is minimal \citep{santalo1949affine}. Equivalently, it is the unique point
$s$ for which the centroid of $(\K-s)^\circ$ is the origin, since the
gradient of the volume of $(\K-s)^\circ$ with respect to $s$ equals
$(d+1)\int_{(\K-s)^\circ}y\,dy$.

We seek a position $J$ for which $L=(d+1)J^\circ$ is isotropic, meaning
that its uniform distribution has mean zero and covariance $I$.
For $d\geq2$, \citet[Theorem~1.1]{bizeul2026optimal} proves that
\begin{align}
    (d+1)\E[\pi_J(U)]=\E[h_L(U)]\leq C\sqrt{d\log d},
    \qquad h_L(u)=\sup_{y\in L}\ip{u}{y},
\end{align}
without any symmetry assumption. Since
$\pi_J=h_{J^\circ}=h_L/(d+1)$, this gives
\begin{align}
    M(J)
    =\max\left\{\frac1{\sqrt d},\frac{\E[h_L(U)]}{d+1}\right\}
    \leq C\sqrt{\frac{\log(ed)}d}.
\end{align}
It's worth noting that the same $d^{-1/2}$ dependence up to logarithmic factors already follows
from \citet[Theorem~1.2]{milman2015mean}, with $\log^2(1+d)$ in place
of $\sqrt{\log d}$ in the mean-width bound. The rest of the appendix
constructs an approximate version of this position in polynomial time
and controls the effect of the approximation errors on $M$ and the ball
inclusions in \cref{prop:positioning}.

\begin{theorem}
    \label{thm:approximate-positioning}
    Let $\K\subset\R^d$ be a convex body containing
    $x_0+\mathbb B_r$ for some $x_0\in\R^d$ and $r>0$, and let
    $s=s(\K)$ be its Santal\'o point. Suppose that
    \begin{align}
        \|\hat s-s\|_2\leq\zeta,
        \qquad 0\leq\zeta\leq\frac{r}{16(d+1)^4}.
    \end{align}
    Then $\hat s$ lies in the interior of $\K$.
    Set
    \begin{align}
        P=(\K-\hat s)^\circ,
        \qquad \Sigma=\operatorname{Cov}(\operatorname{Unif}(P)).
    \end{align}
    If a positive definite matrix $A$ satisfies
    \begin{align}
        (1-\nu)\Sigma\preceq A\preceq(1+\nu)\Sigma,
        \qquad 0\leq\nu\leq\frac1{16(d+1)^2},
    \end{align}
    then the body $J=(d+1)A^{1/2}(\K-\hat s)$ satisfies
    \begin{align}
        \mathbb B_1\subseteq J\subseteq\mathbb B_{d+1}
        \subseteq\mathbb B_{2d},
        \qquad
        M(J)=\max\left\{\frac1{\sqrt d},\E[\pi_J(U)]\right\}
        \leq C\sqrt{\frac{\log(ed)}d},
    \end{align}
    where $U$ is uniform on the Euclidean unit sphere and $C\geq1$
    is a universal constant.
\end{theorem}

\begin{proof}
    Write $B=\mathbb B_1$.

    \proofstepinline{1} We first establish the bounds for the exact
    Santal\'o point and covariance. Define
    \begin{align}
        P_0=(\K-s)^\circ,
        \qquad \Sigma_0=\operatorname{Cov}(\operatorname{Unif}(P_0)),
        \qquad H=\Sigma_0^{-1/2}P_0.
    \end{align}
    Since $s$ is the Santal\'o point, $P_0$ has centroid zero. Thus $H$
    is isotropic. By \citet[Theorem~4.1]{kannan1995isoperimetric},
    \begin{align}
        \sqrt{\frac{d+2}{d}}B\subseteq H\subseteq\sqrt{d(d+2)}B.
        \ineqlabel{eq:positioning-isotropic-balls}
    \end{align}
    For $d\geq2$, \citet[Theorem~1.1]{bizeul2026optimal} gives
    \begin{align}
        \E[h_H(U)]\leq C\sqrt{d\log d}.
        \ineqlabel{eq:positioning-isotropic-width}
    \end{align}
    The exact construction $J_0=(d+1)\Sigma_0^{1/2}(\K-s)$ satisfies
    $(d+1)J_0^\circ=H$. Taking polars in
    \cref{eq:positioning-isotropic-balls} therefore gives
    \begin{align}
        \frac{d+1}{\sqrt{d(d+2)}}B
        \subseteq J_0\subseteq(d+1)\sqrt{\frac d{d+2}}B.
    \end{align}
    Since $d(d+2)<(d+1)^2$ and $d/(d+2)<1$, this implies
    $B\subseteq J_0\subseteq(d+1)B$. Also, for $d\geq2$,
    \begin{align}
        \E[\pi_{J_0}(U)]
        =\frac1{d+1}\E[h_H(U)]
        \leq\frac C{d+1}\sqrt{d\log d}.
    \end{align}
    We next control the changes caused by replacing $s$ and $\Sigma_0$
    with $\hat s$ and $A$.

    \proofstepinline{2} We compare the polars after changing the center.
    First, \cref{eq:positioning-isotropic-balls} gives
    $-H\subseteq\sqrt{d(d+2)}B\subseteq dH$. Undoing the linear transformation
    yields $-P_0\subseteq dP_0$, and taking polars gives
    $-(\K-s)\subseteq d(\K-s)$. Since
    $x_0+rB\subseteq\K$, we have
    \begin{align}
        rB\subseteq\K-\K
        =(\K-s)-(\K-s)
        \subseteq(\K-s)+d(\K-s)
        =(d+1)(\K-s).
    \end{align}
    In particular, $\frac r{d+1}B\subseteq\K-s$. Set
    \begin{align}
        e=\hat s-s,
        \qquad \beta=\frac{(d+1)\zeta}{r}
        \leq\frac1{16(d+1)^3}<1.
    \end{align}
    The bound $\|e\|_2\leq\zeta$ now gives
    $e,-e\in\beta(\K-s)$. By convexity,
    \begin{align}
        (\K-s)-e&\subseteq(1+\beta)(\K-s),\\
        (1-\beta)(\K-s)+e&\subseteq\K-s.
    \end{align}
    Therefore
    \begin{align}
        (1-\beta)(\K-s)
        \subseteq\K-\hat s
        \subseteq(1+\beta)(\K-s).
    \end{align}
    The first inclusion shows that $\hat s$ lies in the interior of
    $\K$. Taking polars gives
    \begin{align}
        \frac1{1+\beta}P_0\subseteq P
        \subseteq\frac1{1-\beta}P_0.
        \ineqlabel{eq:positioning-polar-comparison}
    \end{align}

    \proofstepinline{3} We compare the covariance matrices.
    For a convex body $Q$, let $\Sigma_Q$ be the covariance of its
    uniform distribution and let $|Q|$ denote its volume.
    If $Q\subseteq R$, then for every
    $v\in\R^d$,
    \begin{align}
        v^\top\Sigma_Rv
        &=\min_{t\in\R}\frac1{|R|}
        \int_R(\ip{v}{y}-t)^2\,dy
        \geq\frac1{|R|}\min_{t\in\R}
        \int_Q(\ip{v}{y}-t)^2\,dy
        =\frac{|Q|}{|R|}v^\top\Sigma_Qv.
    \end{align}
    The inequality follows from nonnegativity of the integrand before
    minimizing over $t$. Consequently,
    \begin{align}
        \Sigma_R\succeq\frac{|Q|}{|R|}\Sigma_Q,
        \ineqlabel{eq:positioning-covariance-inclusion}
    \end{align}
    without requiring the two centroids to agree.

    Applying \cref{eq:positioning-covariance-inclusion} to
    $(1+\beta)^{-1}P_0\subseteq P$ from
    \cref{eq:positioning-polar-comparison},
    \begin{align}
        \Sigma
        &\succeq\frac{|(1+\beta)^{-1}P_0|}{|P|}
        \frac{\Sigma_0}{(1+\beta)^2}
        =\frac{|P_0|}{(1+\beta)^{d+2}|P|}\Sigma_0
        \succeq\frac{(1-\beta)^d}{(1+\beta)^{d+2}}\Sigma_0,
    \end{align}
    where the last inequality uses $|P|\leq(1-\beta)^{-d}|P_0|$.
    Applying the same inequality to $P\subseteq(1-\beta)^{-1}P_0$
    and rearranging,
    \begin{align}
        \Sigma
        &\preceq\frac{|(1-\beta)^{-1}P_0|}{|P|}
        \frac{\Sigma_0}{(1-\beta)^2}
        =\frac{|P_0|}{(1-\beta)^{d+2}|P|}\Sigma_0
        \preceq\frac{(1+\beta)^d}{(1-\beta)^{d+2}}\Sigma_0,
    \end{align}
    where the last inequality uses $|P|\geq(1+\beta)^{-d}|P_0|$.
    Combining these bounds with the assumed accuracy of $A$ gives
    \begin{align}
        (1-\nu)\frac{(1-\beta)^d}{(1+\beta)^{d+2}}\Sigma_0
        \preceq A\preceq
        (1+\nu)\frac{(1+\beta)^d}{(1-\beta)^{d+2}}\Sigma_0.
        \ineqlabel{eq:positioning-matrix-comparison}
    \end{align}

    \proofstepinline{4} We prove the ball inclusions for the approximate
    construction. Set
    \begin{align}
        T=A^{-1/2}\Sigma_0^{1/2},
        \qquad T^\top T=\Sigma_0^{1/2}A^{-1}\Sigma_0^{1/2}.
    \end{align}
    Then $T=I$ when $A=\Sigma_0$, and
    $(d+1)J^\circ=A^{-1/2}P$.
    Since $P_0=\Sigma_0^{1/2}H$,
    \cref{eq:positioning-polar-comparison} implies
    \begin{align}
        \frac1{1+\beta}TH
        \subseteq A^{-1/2}P
        \subseteq\frac1{1-\beta}TH.
        \ineqlabel{eq:positioning-transformed-polar}
    \end{align}
    To bound the eigenvalues of $T^\top T$, set
    \begin{align}
        \kappa=\frac1{\sqrt{1-\nu}}
        \left(\frac{1+\beta}{1-\beta}\right)^{(d+2)/2}.
    \end{align}
    Inverting \cref{eq:positioning-matrix-comparison} and multiplying
    on both sides by $\Sigma_0^{1/2}$ gives
    \begin{align}
        \frac{\lambda_{\min}(T^\top T)}
        {(1+\beta)^2}
        \geq\frac1{1+\nu}
        \left(\frac{1-\beta}{1+\beta}\right)^{d+2}
        \explan{\geq}{a}\frac1{\kappa^2},\qquad\text{and}\qquad
        \frac{\lambda_{\max}(T^\top T)}
        {(1-\beta)^2}
        \leq\kappa^2.
        \ineqlabel{eq:positioning-eigenvalues}
    \end{align}
    Here $\lambda_{\min}(\cdot)$ and $\lambda_{\max}(\cdot)$ denote
    the smallest and largest eigenvalues, respectively and \reason{a} uses
    $(1+\nu)^{-1}\geq1-\nu$.
    For every $x\in\R^d$,
    \begin{align}
        \|Tx\|^2=x^\top T^\top Tx,
    \end{align}
    so the square roots of these eigenvalues bound the change in
    Euclidean lengths. Combining
    \cref{eq:positioning-isotropic-balls,eq:positioning-transformed-polar,eq:positioning-eigenvalues},
    we obtain
    \begin{align}
        \frac1\kappa\sqrt{\frac{d+2}{d}}B
        \subseteq A^{-1/2}P\subseteq\kappa\sqrt{d(d+2)}B.
        \ineqlabel{eq:positioning-approximate-balls}
    \end{align}

    It remains to check that $\kappa$ is small enough. For
    $0\leq t\leq1/2$, we have
    $\log((1+t)/(1-t))\leq3t$ and $-\log(1-t)\leq2t$.
    Since $\beta\leq1/(16(d+1)^3)$ and $\nu\leq1/(16(d+1)^2)$,
    \begin{align}
        \log\kappa^2
        &=(d+2)\log\frac{1+\beta}{1-\beta}-\log(1-\nu)\\
        &\leq3(d+2)\beta+2\nu
        \leq\frac7{16(d+1)^2}
        \leq-\log\left(1-\frac1{(d+1)^2}\right).
    \end{align}
    Thus $\kappa\leq(d+1)/\sqrt{d(d+2)}\leq2/\sqrt3$, and
    \begin{align}
        \kappa\sqrt{d(d+2)}\leq d+1,
        \qquad \frac1\kappa\sqrt{\frac{d+2}{d}}
        \geq\frac{d+2}{d+1}>1.
    \end{align}
    It follows from \cref{eq:positioning-approximate-balls} that
    $B\subseteq A^{-1/2}P\subseteq(d+1)B$.
    Since $A^{-1/2}P=(d+1)J^\circ$, taking polars gives
    \begin{align}
        B\subseteq J\subseteq(d+1)B\subseteq2dB.
    \end{align}

    \proofstepinline{5} We bound $M(J)$. First, we control the effect of
    $T$ on the average support function.
    Let $G\sim\N(0,I_d)$, and independently let $Z$ be centered
    Gaussian with covariance
    \begin{align}
        \lambda_{\max}(T^\top T)I_d-T^\top T,
    \end{align}
    which is positive semidefinite. Then
    \begin{align}
        \E[h_{TH}(G)]
        &=\E[h_H(T^\top G)]
        \explan{\leq}{a}\E[h_H(T^\top G+Z)]
        =\sqrt{\lambda_{\max}(T^\top T)}
        \,\E[h_H(G)].
    \end{align}
    For \reason{a}, apply conditional Jensen to the convex support
    function, using $\E[Z\mid G]=0$. The last equality uses
    positive homogeneity and
    \begin{align}
        T^\top G+Z
        \sim\N\!\left(0,
        \lambda_{\max}(T^\top T)I_d\right).
    \end{align}
    The Gaussian radius and direction are independent,
    so dividing by $\E\|G\|$ gives
    \begin{align}
        \E[h_{TH}(U)]
        \leq\sqrt{\lambda_{\max}(T^\top T)}
        \,\E[h_H(U)].
        \ineqlabel{eq:positioning-width-transformation}
    \end{align}
    For $d\geq2$, we can therefore bound
    \begin{align}
        \E[\pi_J(U)]
        &=\frac1{d+1}\E[h_{A^{-1/2}P}(U)]\\
        &\explan{\leq}{a}\frac1{(d+1)(1-\beta)}
        \E[h_{TH}(U)]\\
        &\explan{\leq}{b}
        \frac{\sqrt{\lambda_{\max}(T^\top T)}}
        {(d+1)(1-\beta)}
        \E[h_H(U)]\\
        &\explan{\leq}{c}\frac\kappa{d+1}\E[h_H(U)]\\
        &\explan{\leq}{d}\frac{C\kappa}{d+1}\sqrt{d\log d}
        \leq C'\sqrt{\frac{\log(ed)}d}.
    \end{align}
    Here \reason{a} uses \cref{eq:positioning-transformed-polar};
    \reason{b} uses \cref{eq:positioning-width-transformation};
    \reason{c} uses \cref{eq:positioning-eigenvalues}; and
    \reason{d} uses \cref{eq:positioning-isotropic-width}. The last
    inequality uses $\kappa\leq2/\sqrt3$ and
    $\sqrt d/(d+1)\leq d^{-1/2}$.
    For $d=1$, the inclusion $B\subseteq J$ gives
    $\pi_J(U)\leq1$ and hence $M(J)=1$. Finally, for every $d\geq1$,
    the term $d^{-1/2}$ in the definition of $M(J)$ is also at most
    $\sqrt{\log(ed)/d}$. This proves the stated bound on $M(J)$.
\end{proof}

\subsection{Computation}
\label{app:positioning-computation}

We first use the following consequence of
\citet[Theorem~4.6.1]{grotschel1988geometric} to obtain the normalization
required for computing the Santal\'o point.

\begin{lemma}
    \label{lem:initial-rounding}
    Let $\K\subset\R^d$ be a convex body given by a separation oracle,
    together with known $x_0\in\R^d$ and $0<r\leq R$ such that
    $x_0+\mathbb B_r\subseteq\K\subseteq x_0+\mathbb B_R$.
    A deterministic algorithm computes an invertible affine map $\Phi$
    such that
    \begin{align}
        \mathbb B_1\subseteq\Phi(\K)
        \subseteq\sqrt d(d+1)\mathbb B_1
        \subseteq(d+1)^{3/2}\mathbb B_1,
    \end{align}
    using a number of oracle calls and arithmetic operations polynomial
    in $d$ and $\log(R/r)$.
\end{lemma}

Apply the rounding theorem of
\citet*[Theorem~4.6.1]{grotschel1988geometric} to
$(\K-x_0)/r$, whose known inner and outer radii are $1$ and $R/r$.
This theorem finds two ellipsoids with the same center and axes: an inner
ellipsoid contained in the normalized body, and an outer ellipsoid
containing it. Every radius of the outer ellipsoid is $\sqrt d(d+1)$
times the corresponding inner radius. An affine map taking the inner
ellipsoid to $\mathbb B_1$ therefore takes the outer ellipsoid to
$\sqrt d(d+1)\mathbb B_1$, proving the stated ball inclusions.
Separately, the running-time guarantee in GLS Theorem~4.6.1 gives a
number of oracle calls and arithmetic operations polynomial in $d$ and
$\log(R/r)$ for this normalized input.

We use the following result of
\citet[Proposition~31]{reis2023subspace}.\footnote{The confidence parameter
    is not explicit in their statement, but the formulation below follows from
    their proof. The proof uses at most $Q=\operatorname{poly}(d,1/\delta)$
    volume estimates. By \citet[Theorem~2.1]{kannan1997random}, each estimate
    can have failure probability at most $\alpha/Q$ at a cost logarithmic in
    $Q/\alpha$. A union bound gives the stated success probability and
polynomial dependence on $\log(1/\alpha)$.}

\begin{theorem}
    \label{thm:santalo}
    Let $\K\subset\R^d$ be a convex body given by a separation oracle,
    satisfying
    \begin{align}
        \mathbb B_1\subseteq\K\subseteq(d+1)^{3/2}\mathbb B_1,
    \end{align}
    and let $s(\K)$ be its Santal\'o point. For every distance accuracy
    $\delta>0$ and failure probability $\alpha\in(0,1)$, there is a
    randomized algorithm that returns $\hat s\in\R^d$ such that
    \begin{align}
        \Pp\left(\|\hat s-s(\K)\|_2\leq\delta\right)\geq1-\alpha.
    \end{align}
    The number of oracle calls and arithmetic operations is polynomial
    in $d$, $1/\delta$, and $\log(1/\alpha)$.
\end{theorem}

The next step is to sample from $(\K-\hat s)^\circ$ to estimate its
covariance. Given known inner and outer ball bounds centered at the origin
for $\K-\hat s$, separation-oracle access to $\K$ gives an
$\epsilon'$-weak membership oracle for $(\K-\hat s)^\circ$, for any
$\epsilon'>0$, in deterministic polynomial time, with polynomial
dependence on $\log(1/\epsilon')$ for $\epsilon'<1$
\citep[Theorem~4.2.2 and Lemma~4.4.1]{grotschel1988geometric}.

We use the following uniform-sampling special case of
\citet[Theorem~B.6(1)]{dadush2011enumerative}, obtained by taking the
density in their theorem to be constant.

\begin{lemma}
    \label{lem:uniform-sampling}
    Let $P\subset\R^d$ be a convex body with known bounds
    $\mathbb B_r\subseteq P\subseteq\mathbb B_R$, where $0<r\leq R$,
    and access to an $\epsilon'$-weak membership oracle for every
    $\epsilon'>0$. For every $\eta\in(0,1)$, there is a randomized
    algorithm that produces $Y\in P$ such that
    \begin{align}
        d_{\mathrm{TV}}\!\left(\mathcal L(Y),\operatorname{Unif}(P)\right)
        \leq\eta.
    \end{align}
    Here $d_{\mathrm{TV}}$ denotes total variation distance and
    $\mathcal L(Y)$ is the distribution of $Y$. The number of oracle
    calls and arithmetic operations is polynomial in $d$, $\log(R/r)$,
    and $\log(1/\eta)$.
\end{lemma}

\begin{lemma}[{\citet[Lemma~4.2]{vempala2010learning}}]
    \label{lem:covariance-estimation}
    Let $X_1,\ldots,X_m$ be independent samples from an isotropic
    log-concave distribution on $\R^d$, and define the sample mean and
    covariance by
    \begin{align}
        \hat\mu=\frac1m\sum_{i=1}^m X_i,
        \qquad
        \widehat\Sigma=\frac1m\sum_{i=1}^m
        (X_i-\hat\mu)(X_i-\hat\mu)^\top.
    \end{align}
    There is a universal constant $C>0$ such that, for every accuracy
    $\eps\in(0,1)$ and failure probability $\alpha\in(0,1/2)$, if
    \begin{align}
        m>C\frac{d}{\eps^2}\log^2\!\left(\frac d\alpha\right),
    \end{align}
    then, with probability at least $1-\alpha$,
    \begin{align}
        \|\hat\mu\|_2\leq\eps,
        \qquad
        \|\widehat\Sigma-I\|_{\op}\leq\eps.
    \end{align}
\end{lemma}

\begin{proof}[Proof of \cref{prop:positioning}]
    Write $B=\mathbb B_1$.

    \proofstepinline{1} We first round the original body.
    Apply \cref{lem:initial-rounding} and write
    $\widetilde\K=\Phi(\K)$, so that
    \begin{align}
        B\subseteq\widetilde\K\subseteq(d+1)^{3/2}B.
    \end{align}
    The known affine map $\Phi$ also gives separation-oracle access to
    $\widetilde\K$. We perform the remaining steps in these coordinates.

    \proofstepinline{2} We approximate the Santal\'o point and obtain
    ball bounds for the shifted polar.
    Apply \cref{thm:santalo} with distance accuracy
    \begin{align}
        \zeta=\frac1{16(d+1)^4}
    \end{align}
    and failure probability $\delta/3$. With probability at least
    $1-\delta/3$,
    \mbox{$\|\hat s-s(\widetilde\K)\|_2\leq\zeta$}.
    On this event, Step~2 of the proof of
    \cref{thm:approximate-positioning}, with $r=1$, gives
    \begin{align}
        \frac1{2(d+1)}B
        \subseteq\widetilde\K-\hat s,
    \end{align}
    since $(d+1)^{-1}B\subseteq\widetilde\K-s(\widetilde\K)$
    and $\zeta\leq1/(2(d+1))$.

    We also ensure that the sampling procedure has valid ball bounds
    when the center estimate fails. Using the separation oracle, test
    whether
    \begin{align}
        \hat s\pm\frac{e_i}{2(d+1)}\in\widetilde\K,
        \qquad i=1,\ldots,d,
    \end{align}
    where $e_1,\ldots,e_d$ are the coordinate unit vectors. If any test
    fails, return $\Phi$. All tests pass whenever the center estimate
    succeeds. Whenever they pass, convexity gives
    \begin{align}
        \hat s+\frac1{2\sqrt d(d+1)}B\subseteq\widetilde\K.
    \end{align}
    Also, $\hat s\in\widetilde\K$ implies
    $\widetilde\K-\hat s\subseteq2(d+1)^{3/2}B$.
    Thus the polar $P=(\widetilde\K-\hat s)^\circ$ has the known bounds
    \begin{align}
        \frac1{2(d+1)^{3/2}}B
        \subseteq P\subseteq2(d+1)^{3/2}B.
        \ineqlabel{eq:positioning-sampling-balls}
    \end{align}

    \proofstepinline{3} We sample from the shifted polar.
    Set
    \begin{align}
        \nu=\frac1{16(d+1)^2},
        \qquad
        m=\left\lceil C\frac d{\nu^2}
        \log^2\!\left(\frac{3d}{\delta}\right)\right\rceil,
        \qquad \eta=\frac{\delta}{3m},
    \end{align}
    with a sufficiently large universal constant $C$.
    The separation oracle for $\widetilde\K$ gives weak membership
    access to $P$ as described above. Apply
    \cref{lem:uniform-sampling} independently $m$ times, each with
    total variation accuracy $\eta$, to obtain $Y_1,\ldots,Y_m$.
    Conditional on any center that passes the tests in Step~2,
    the joint law of these samples is within $m\eta=\delta/3$
    in total variation of $m$ independent uniform samples from $P$.
    This follows by coupling each sampler output with a uniform sample
    and taking a union bound over the $m$ couplings.

    \proofstepinline{4} We estimate the covariance.
    Compute
    \begin{align}
        \bar Y=\frac1m\sum_{i=1}^m Y_i,
        \qquad
        A=\frac1m\sum_{i=1}^m(Y_i-\bar Y)(Y_i-\bar Y)^\top.
    \end{align}
    To analyze this estimate, first suppose that the $Y_i$ are exact
    independent uniform samples from $P$. Let $\mu$ and $\Sigma$
    be the mean and covariance of $\operatorname{Unif}(P)$.
    The random vectors
    $X_i=\Sigma^{-1/2}(Y_i-\mu)$ are isotropic and log-concave, and
    their centered empirical covariance is
    $\Sigma^{-1/2}A\Sigma^{-1/2}$.
    Applying \cref{lem:covariance-estimation} with accuracy $\nu$
    and failure probability $\delta/3$ therefore gives
    \begin{align}
        (1-\nu)\Sigma\preceq A\preceq(1+\nu)\Sigma
        \ineqlabel{eq:positioning-empirical-covariance}
    \end{align}
    with probability at least $1-\delta/3$.
    The total variation bound in Step~3 shows that
    \cref{eq:positioning-empirical-covariance} holds for the actual
    samples with conditional probability at least $1-2\delta/3$.
    The unknown $\mu$ and $\Sigma$ are used only in this analysis;
    the algorithm computes $A$ by subtracting the sample mean.

    \proofstepinline{5} We construct the affine map.
    If $A$ is not positive definite, return $\Phi$.
    Otherwise, compute an upper triangular Cholesky factor $F$ with
    $F^\top F=A$ and return
    \begin{align}
        T(x)=(d+1)F\bigl(\Phi(x)-\hat s\bigr).
    \end{align}
    This is an invertible affine map. Since $F=OA^{1/2}$ for an
    orthogonal matrix $O$, its image of $\K$ is a rotation of
    $(d+1)A^{1/2}(\widetilde\K-\hat s)$.
    On the event that the center estimate and
    \cref{eq:positioning-empirical-covariance} both hold,
    \cref{thm:approximate-positioning} applies with $r=1$.
    Euclidean balls and $M$ are invariant under orthogonal
    transformations, so it gives precisely \cref{eq:positioning}
    for $J=T(\K)$. The probability that either estimate fails is at
    most $\delta/3+2\delta/3=\delta$.

    \proofstepinline{6} We check the running time.
    Initial rounding uses a number of oracle calls and arithmetic
    operations polynomial in $d$ and $\log(R/r)$.
    The Santal\'o computation is polynomial in $d$ and
    $\log(1/\delta)$ because $1/\zeta=16(d+1)^4$.
    The $2d$ membership tests add only polynomial work.
    The ratio of the radii in
    \cref{eq:positioning-sampling-balls} is $4(d+1)^3$, and
    \begin{align}
        m=O\!\left(d(d+1)^4\log^2\!\left(\frac{3d}{\delta}\right)\right),
        \qquad
        \log(1/\eta)=O\!\left(\log(d+1)+\log(1/\delta)\right).
    \end{align}
    Thus sampling is polynomial in $d$ and $\log(1/\delta)$,
    including the implementation of the weak membership oracle by
    separation. This composition is also used in the proof of
    \citet[Theorem~3.7]{dadush2011enumerative}.
    The sample covariance and Cholesky factor require $O(md^2+d^3)$
    arithmetic operations, including scalar square roots.
    The stated complexity follows.
\end{proof}

\end{document}